\documentclass{article}

\PassOptionsToPackage{numbers}{natbib}
\usepackage[main, final]{neurips_2026}

\input{preamble}
\usepackage{amsthm}
\usepackage{amsfonts}
\usepackage{amssymb}
\usepackage{color}
\usepackage{bm}
\usepackage{multirow}
\usepackage{algorithm}
\usepackage{algpseudocode}
\algtext*{EndIf}
\algtext*{EndFor}
\algtext*{EndWhile}
\usepackage{float}
\usepackage{wrapfig}
\usepackage{subcaption}
\usepackage{thmtools}
\usepackage{tikz}
\usetikzlibrary{arrows.meta}
\usepackage[breaklinks=true,bookmarks=false]{hyperref}
\usepackage[capitalize]{cleveref}

\newcommand{\cC}{\ensuremath{\mathcal{C}}}

\newcommand{\cE}{\ensuremath{\mathcal{E}}}
\newcommand{\cF}{\ensuremath{\mathcal{F}}}

\newcommand{\cK}{\ensuremath{\mathcal{K}}}
\newcommand{\cL}{\ensuremath{\mathcal{L}}}

\newcommand{\cN}{\ensuremath{\mathcal{N}}}

\newcommand{\cP}{\ensuremath{\mathcal{P}}}

\newcommand{\cT}{\ensuremath{\mathcal{T}}}

\newcommand{\cV}{\ensuremath{\mathcal{V}}}

\DeclareMathOperator{\E}{\textbf{E}}

\newcommand{\bzero}{\ensuremath{\bm{0}}}
\newcommand{\bone}{\ensuremath{\bm{1}}}
\newcommand{\bA}{\ensuremath{\bm{A}}}

\newcommand{\bI}{\ensuremath{\bm{I}}}

\newcommand{\bU}{\ensuremath{\bm{U}}}

\newcommand{\bZ}{\ensuremath{\bm{Z}}}
\newcommand{\ba}{\ensuremath{\bm{a}}}

\newcommand{\bd}{\ensuremath{\bm{d}}}
\newcommand{\be}{\ensuremath{\bm{e}}}

\newcommand{\bg}{\ensuremath{\bm{g}}}
\newcommand{\bh}{\ensuremath{\bm{h}}}

\newcommand{\br}{\ensuremath{\bm{r}}}
\newcommand{\bs}{\ensuremath{\bm{s}}}

\newcommand{\bu}{\ensuremath{\bm{u}}}

\newcommand{\bw}{\ensuremath{\bm{w}}}
\newcommand{\bx}{\ensuremath{\bm{x}}}
\newcommand{\by}{\ensuremath{\bm{y}}}
\newcommand{\bz}{\ensuremath{\bm{z}}}

\def\st/{\textsuperscript{st}}
\def\nd/{\textsuperscript{nd}}
\def\rd/{\textsuperscript{rd}}
\def\th/{\textsuperscript{th}}

\newcommand{\abs}[1]{\ensuremath{{\left\vert #1 \right\vert}}}

\newcommand{\ind}{\mathbf{1}}

\newcommand{\R}{\mathbb{R}}

\RequirePackage{xspace}
\newcommand{\HAS}{\ensuremath{\mathcal{HAS}}\xspace}
\newcommand{\HASViol}{\ensuremath{\mathcal{HAS}\text{-}\mathrm{Viol}}\xspace}

\makeatletter
\@ifundefined{theorem}{\newtheorem{theorem}{Theorem}[section]}{}
\@ifundefined{proposition}{\newtheorem{proposition}[theorem]{Proposition}}{}
\@ifundefined{lemma}{\newtheorem{lemma}[theorem]{Lemma}}{}
\@ifundefined{corollary}{\newtheorem{corollary}[theorem]{Corollary}}{}
\@ifundefined{conjecture}{}{}
\theoremstyle{definition}
\@ifundefined{definition}{\newtheorem{definition}[theorem]{Definition}}{}
\@ifundefined{assumption}{\newtheorem{assumption}[theorem]{Assumption}}{}
\theoremstyle{remark}
\@ifundefined{remark}{\newtheorem{remark}[theorem]{Remark}}{}
\makeatother
\crefname{assumption}{assumption}{assumptions}
\Crefname{assumption}{Assumption}{Assumptions}
\Crefname{equation}{Eq.}{Eqs.}

\newcommand{\myparagraph}[1]{\noindent\textbf{#1.}}

\title{Provable Cross-Modal Transfer of Hierarchical Concept Structure in CLIP Embeddings}

\author{%
  Nghia Nguyen \quad Tianjiao Ding \quad Lachlan Ewen MacDonald \quad Ren\'e Vidal \\
  University of Pennsylvania \\
  \texttt{nghianhh@seas.upenn.edu}
}

\begin{document}
\maketitle
\begin{abstract}
Contrastive vision-language models such as CLIP are trained to align images and text, not to represent taxonomies. Yet empirical studies show that CLIP embeddings exhibit hierarchical structure aligned with the hierarchy of semantic concepts. A theoretical characterization of why such hierarchy emerges from a non-hierarchical objective remains open. In this paper, we formalize this phenomenon as \emph{Hierarchical Angular Separation} (\HAS), a geometric property whereby each non-leaf concept embedding is more similar to its descendants than to off-branch concepts. Empirically, we show that \HAS holds approximately in raw images and randomly initialized CLIP image embeddings, while in text embeddings it holds approximately only after contrastive learning. To theoretically explain this asymmetry, we analyze an \emph{unconstrained features model} in which the image embeddings are fixed and only the text embeddings are trained. At the global minimizer, the text-side \HAS violation percentage at each node is at most the image-side \HAS violation and near-violation percentage. The key mechanism is the hierarchical co-occurrence structure of image-text pairs, which biases text embeddings toward root-centered image prototypes. For a linear contrastive loss, we derive a closed-form solution that makes this mechanism explicit. We extend the analysis to the InfoNCE loss and establish analogous guarantees under sufficient regularization. Synthetic experiments validate the theory and show that \HAS is transferred even beyond the regimes covered by our analysis.

\end{abstract}

\section{Introduction}

Given a collection of image-text pairs, Contrastive Language-Image Pretraining (CLIP) learns text and image encoders by aligning paired embeddings and separating unpaired ones. The learned representations enable strong transfer across tasks such as image classification, visual question answering, and multimodal retrieval~\citep{radford2021learning,chattopadhyayBOOTSTRAPPINGVARIATIONALINFORMATION2024,lin2023fine}. 

One possible explanation for this empirical success is that CLIP embeddings implicitly encode rich semantic and geometric structure. In particular, embeddings of semantic concepts have been shown to organize into linear subspaces~\citep{bhallaInterpretingCLIPSparse2024,gandelsmanInterpretingCLIPsImage2024} and to reflect hierarchical relationships aligned with noun hierarchies in WordNet~\citep{miller_wordnet_1995,alperEmergentVisualSemanticHierarchies2024,zaigrajewInterpretingCLIPHierarchical2025a,nguyenHierarchicalConceptEmbedding2026}. 
For example, a \texttt{cat} and a \texttt{dog} are both \texttt{mammals}, and a \texttt{mammal} and a \texttt{bird} are both \texttt{animals}. 
Recent work has attempted to characterize such hierarchical structure through geometric conditions, for example via orthogonality relations between embeddings of semantic concepts~\citep{parkGeometryCategoricalHierarchical2025,jiangUncoveringMeaningsEmbeddings2023}. However, these conditions fail to capture a fundamental property of semantic hierarchies: a concept should be more similar to its descendants than to concepts from unrelated branches~\citep{ganeaHyperbolicEntailmentCones2018,desaiHyperbolicImageTextRepresentations2024,palCompositionalEntailmentLearning2025,alperEmergentVisualSemanticHierarchies2024}. Empirical studies further suggest that CLIP image embeddings encode hierarchy through separated subtrees~\citep{nguyenHierarchicalConceptEmbedding2026}, but this evidence does not quantify the extent of such structure across models and scales, nor does it provide a principled explanation of its origin.

A natural hypothesis is that hierarchical structure is learned through contrastive learning: if one modality encodes semantic relationships, alignment should transfer them to the other. Our experiments show that the two modalities differ. Approximate hierarchical structure is already present in raw-pixel image data and 
this structure is preserved by a randomly initialized CLIP encoder, implying that the CLIP embedding space natively exhibits hierarchy prior to any training.
In contrast, raw text consists of discrete symbols whose semantic relationships are not reflected in their randomly initialized embeddings; any geometric structure must therefore be learned through training.

These observations raise a central question: \emph{how does contrastive learning transform and propagate hierarchical structure across modalities?} In particular, why does alignment amplify and transfer structure from image to text embeddings, rather than suppress it? While the origin of hierarchical structure in image data remains open, our focus in this paper is on understanding how contrastive learning transfers such structure to other modalities once it is present in one modality.

\begin{figure}[t]
\centering
\includegraphics[width=\linewidth]{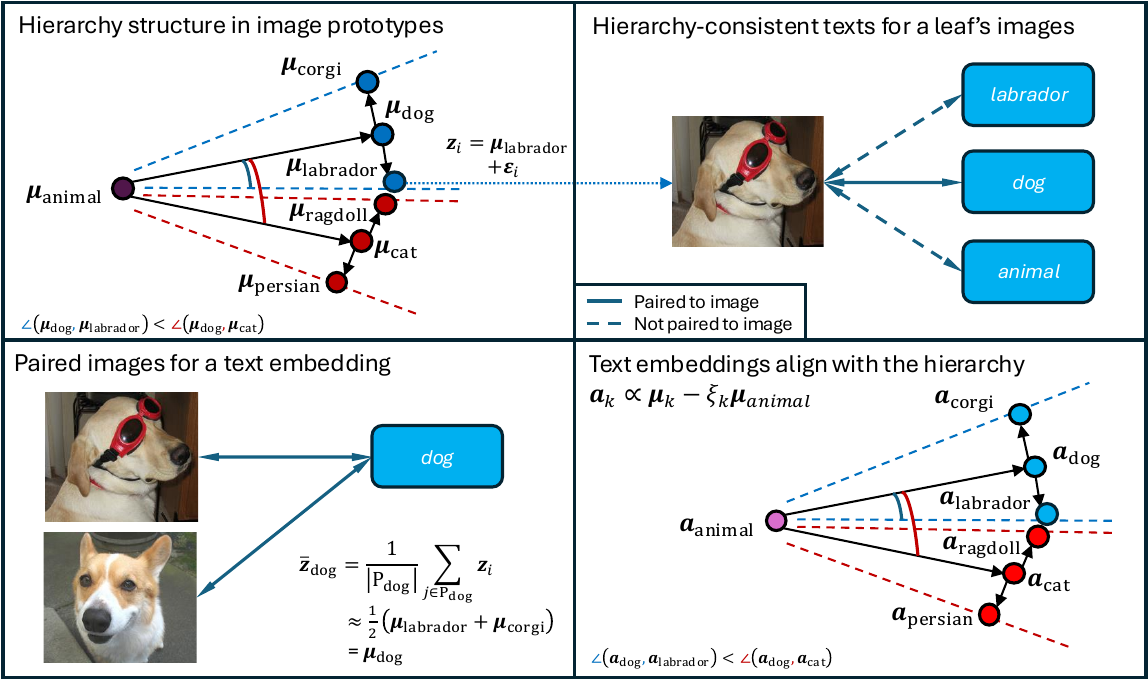}
\caption[How contrastive learning transfers hierarchy from image to text embeddings]{How contrastive learning transfers hierarchy from image to text embeddings. 
\emph{Top-left:} the image prototypes are arranged so that each concept is closer to its descendants than to concepts in other subtrees. The blue and red dotted lines indicate the boundary of the \texttt{dog} and \texttt{cat} subtrees, respectively.
\emph{Top-right:} each image of a \texttt{labrador} is a variation of the \texttt{labrador} prototype, and its caption may pair with any concept on the path from the root to that leaf, so an image can be captioned \texttt{labrador}, \texttt{dog} or \texttt{animal}. 
\emph{Bottom-left:} the pairing also means that the caption \texttt{dog} pairs with images of \texttt{dog} or any of its descendants, so the average of these images approaches the mean of the \texttt{dog} subtree.
\emph{Bottom-right:} contrastive learning pulls each text embedding toward the corresponding subtree mean, so the text embeddings inherit the arrangement of the image prototypes.}
\label{fig:path-consistent-caption}
\end{figure}

To study this phenomenon, we formalize hierarchical structure as a geometric property called \emph{Hierarchical Angular Separation} (\HAS): given a taxonomy of concepts, an ancestor has higher cosine similarity to each of its descendants concepts than to any of its off-branch concepts. We quantify violations from this ordering using the \HAS violation percentage (\HASViol): the percentage of off-branch concepts that violate the \HAS condition. Here, off-branch concepts exclude both the concept's subtree and its ancestors. 
Our experiments across CLIP models of varying scales show that image embeddings natively exhibit low \HASViol at random initialization, whereas text embeddings begin with high \HASViol and achieve low values only after contrastive training.

These empirical observations motivate the need for a theoretical analysis. However, analyzing contrastive learning theoretically is challenging due to the non-convexity of the optimization problem. To address this, we adopt the \emph{Unconstrained Features Model} (UFM)~\citep{zhu2021geometric,mixon2022neural,ji2021unconstrained}, which abstracts away the deep network by optimizing directly over last-layer features. Within this framework, we show that hierarchical structure in image embeddings transfers to text embeddings under contrastive learning. The key mechanism, illustrated in \Cref{fig:path-consistent-caption}, is the hierarchical co-occurrence structure of image-text pairs: although each image corresponds to a specific concept (e.g., \texttt{dog}), its paired text may correspond to that concept or to one of its hypernyms (e.g., \texttt{mammal}, \texttt{animal}). This structure induces a bias that propagates hierarchical relationships from image to text embeddings.

\myparagraph{Paper Contributions} In this paper, we empirically characterize the hierarchical structure of CLIP embeddings and theoretically explain how contrastive learning transfers this structure from image to text embeddings:
\begin{enumerate}
    \item \textbf{Empirical characterization of the hierarchical structure.}
    Using WordNet, we show that text embeddings have high \HASViol at random initialization and low \HASViol after pretraining, while image embeddings already have low \HASViol at initialization. This reveals a systematic asymmetry: hierarchy is learned in text but partly present in image embeddings prior to training.

    \item \textbf{Hierarchical co-occurrence transfers image geometry to text at the linear-loss optimum.}
    Under the UFM with hierarchical co-occurrence, we derive a closed-form minimizer of a linear contrastive loss~\citep{jiPowerContrastFeature2023}, showing that optimal text embeddings align with root-centered image prototypes. The text-side \HASViol at each anchor is at most the image-side percentage of off-branch nodes that violate or nearly violate \HAS.

    \item \textbf{Extension to InfoNCE and robustness beyond theory.}
    We establish an analogous bound on text \HASViol for the InfoNCE loss used in CLIP under sufficient regularization. Furthermore, we empirically show that transfer occurs even when the regularization condition does not hold.
\end{enumerate}

\section{Related Work}

\myparagraph{Unconstrained Features Model}
The UFM was originally used to explain Neural Collapse~\citep{papyan2020prevalence}. Subsequent work showed that the Simplex Equiangular Tight Frame is the unique optimal geometry for cross-entropy with weight decay~\citep{zhu2021geometric,mixon2022neural,ji2021unconstrained} and for supervised contrastive losses~\citep{grafDissectingSupervisedContrastive2021,behniaSupervisedContrastiveRepresentation2024}. However, these analyses focus on classification with mutually exclusive classes. We extend the UFM to \emph{contrastive learning with hierarchically structured} image-text pairings and characterize the geometry that emerges at global minimizers.

\myparagraph{Geometry of Representations in Foundation Models}
Recent work has documented geometric structure in pretrained embedding spaces. One line of work shows that semantic concepts form linear or compositional subspaces~\citep{parkLinearRepresentationHypothesis2024,tragerLinearSpacesMeanings2023}, exhibit hierarchical organization~\citep{parkGeometryCategoricalHierarchical2025,nguyenHierarchicalConceptEmbedding2026,alperEmergentVisualSemanticHierarchies2024}, or give rise to phenomena such as the modality gap~\citep{leviDoubleEllipsoidGeometryCLIP2025,liangMindGapUnderstanding2022}. Relatedly, \citet{zhaoGeometrySemanticsNextToken2025} infer a semantic hierarchy from sign patterns in a learned concept basis for next-token prediction. \citet{navaHierarchicalConceptGeometry2026} derive coarse-to-fine spectral organization from hierarchy-dependent word co-occurrence in word2vec-style embeddings. We instead study a given tree of object categories and explain how \HAS transfers from image to text embeddings during contrastive pretraining.

\myparagraph{Learning Theory with Hierarchy Models} 
Another line of work studies how hierarchical latent structure shapes learning and generation, including hierarchical concept learning in language models~\citep{allen-zhuPhysicsLanguageModels2023,kongLearningDiscreteConcepts2025}, emergent properties under hierarchical contrastive learning~\citep{guBioCLIP2Emergent2025}, and phase transitions in diffusion models signaling an underlying latent hierarchy~\citep{sclocchiPhaseTransitionDiffusion2025,sclocchiProbingLatentHierarchical2025}. Closest to our work, several papers~\citep{sclocchiPhaseTransitionDiffusion2025,cagnettaHowDeepNeural2024} analyze the Random Hierarchy Model, a \emph{supervised} setting in which the sample for a class is generated by a hierarchical composition of latent features. However, these works study compositional hierarchies of features, while we instead study hierarchies over object categories.

\myparagraph{Hierarchical Representation Learning}
Traditionally, hierarchical information is explicitly encoded into models through specialized architectures or objectives. A popular approach is to embed hierarchies in hyperbolic space, exploiting its exponential volume growth~\citep{nickelPoincareEmbeddingsLearning2017, ganeaHyperbolicEntailmentCones2018, desaiHyperbolicImageTextRepresentations2024, palCompositionalEntailmentLearning2025, wangLearningVisualHierarchies2025}. In contrast, we analyze when implicit hierarchical structure transfers in self-supervised learning.

\myparagraph{Contrastive Learning Theory}
Theoretical analyses have connected contrastive objectives to spectral decompositions of augmentation graphs~\citep{haochenProvableGuaranteesSelfSupervised2021,nakadaUnderstandingMultimodalContrastive2023}, and have shown that contrastive objectives can invert the data-generating process~\citep{zimmermannContrastiveLearningInverts2021}. A separate line of work provides theoretical and empirical evidence that self-supervised representations cluster at multiple hierarchical levels~\citep{tian2021understandingselfsupervisedlearningdual,benshaul2023reverseengineeringselfsurvised}. By contrast, we analyze how the contrastive loss interacts with hierarchical co-occurrence structures.

\section{Hierarchical Angular Separation and Its Modality Asymmetry in CLIP}\label{sec:hierarchical-angular-separation}
We first define a hierarchical similarity ordering for embeddings,  Hierarchical Angular Separation (\HAS). In this definition, each concept embedding is closer to its descendant concept embeddings than to any of its off-branch concepts.
Pretraining creates a modality asymmetry in CLIP's hierarchical structure: text embeddings acquire \HAS after training, whereas image embeddings exhibit \HAS approximately at initialization and strengthened \HAS after training. We formalize \HAS in \Cref{sec:hierarchical-angular-separation-definition} and measure it across CLIP variants in \Cref{sec:hierarchical-angular-separation-empirical}.

\subsection{Definition of Hierarchical Angular Separation}\label{sec:hierarchical-angular-separation-definition}

Datasets of natural images and language often contain semantic concepts that can be organized into hierarchies. 
For example, a \texttt{cat} and a \texttt{dog} are both \texttt{mammals}, and a \texttt{mammal} and a \texttt{bird} are both \texttt{animals}. Below, we introduce the notation used to describe such hierarchies.

\myparagraph{Hierarchy}
Let $\mathcal{T}=(V,E)$ be a rooted tree with $|V|=M$ nodes, leaf set $V_{\mathrm{leaf}}\subseteq V$, and root node $\mathrm{root}\in V$. 
Fix an enumeration of the nodes and identify $V$ with $[M]:=\{1,\ldots,M\}$.
Each $u\in V$ corresponds to a \emph{synset} (e.g., \texttt{mammal}). For $u\in V$, let $\mathrm{Anc}(u)$ denote its ancestors, $\mathrm{Desc}(u)$ its descendants, and $\mathrm{Sub}(u):=\{u\}\cup\mathrm{Desc}(u)$ the subtree rooted at $u$.

\myparagraph{Concept Embeddings}
For each node $k\in V$, we associate a concept embedding $\bh_k\in\mathbb{R}^d$, and write the embedding of the hierarchy $\cT$ as the collection $\{\bh_k:k\in V\}$. In practice, concept embeddings are obtained using text and image encoders in two common ways:
(1) \textit{instance aggregation}~\citep{parkLinearRepresentationHypothesis2024,wollschlagerGeometryRefusalLarge2025,luoPaCEParsimoniousConcept2024}, which averages all image embeddings $\bz_j \in \mathbb{R}^d$ corresponding to a node $u\in V$ (e.g., averaging over all \texttt{dog} images to estimate the concept embedding for \texttt{dog});
(2) \textit{direct concept encoding}~\citep{radford2021learning,tragerLinearSpacesMeanings2023}, which encodes the concept string (e.g., \texttt{image of a dog}) to obtain a text embedding.

\myparagraph{Hierarchical Angular Separation}
We define \HAS as a Euclidean similarity formalization of hierarchy based on angular separation, applicable to either image or text embeddings.
To make this precise, for non-zero $\bx,\by\in\R^d$, we define $\cos(\bx,\by):=\bx^\top\by/(\|\bx\|\|\by\|)$.
\begin{definition}[Hierarchical Angular Separation (\HAS)]\label{def:has}
The concept embeddings $\{\bh_k\}_{k\in V}$ of hierarchy $\mathcal{T}$ (obtained by either instance aggregation or direct concept encoding) satisfy \HAS if for every non-leaf node $u\in V$, every descendant $v\in\mathrm{Desc}(u)$, and every node outside the subtree and not an ancestor, $w\notin \mathrm{Sub}(u)\cup\mathrm{Anc}(u)$, we have
\begin{equation}\label{eq:has}
  \cos(\bh_u,\bh_v) > \cos(\bh_u,\bh_w).
\end{equation}
\end{definition}
\citet{nguyenHierarchicalConceptEmbedding2026} use instance aggregation to estimate concept embeddings, and claim to have observed subtree separation in pretrained CLIP image embeddings without specifying how.

\HAS differs from prior hierarchy-learning objectives that prescribe hierarchical structure during training. Hyperbolic entailment cones~\citep{ganeaHyperbolicEntailmentCones2018, desaiHyperbolicImageTextRepresentations2024, palCompositionalEntailmentLearning2025} and Poincar\'e embeddings~\citep{nickelPoincareEmbeddingsLearning2017} optimize non-Euclidean embeddings with losses designed to enforce hierarchical relations. In contrast, and as we compare in detail in \Cref{app:formalizations-hierarchy}, we define \HAS descriptively rather than prescriptively: \Cref{eq:has} is a Euclidean cosine inequality measured in the learned CLIP space, where the observed angles determine the descendant cone around $\bh_u$ rather than the training objective imposing it.
To quantify how far a set of embeddings is from satisfying \Cref{def:has}, we use the following per-node metric.

\begin{definition}[\HAS Violation Percentage]\label{def:violation-percentage}
For a non-root non-leaf node $u$, we first define the \emph{descendant threshold}, the minimum cosine similarity between $u$ and its least similar descendant, as $s_u:=\min_{v\in\mathrm{Desc}(u)}\cos(\bh_u,\bh_v)$.
The \HAS \emph{violation percentage} at $u$ is then the percentage of the set of off-branch (not in the subtree and not an ancestor) nodes $\{w\}$ that has a larger cosine similarity with $u$ than $s_u$
\begin{equation}\label{eq:violation-percentage}
  \HASViol(u) := \frac{\left|\{w\notin \mathrm{Sub}(u)\cup\mathrm{Anc}(u) : \cos(\bh_u, \bh_w) \geq s_u\}\right|}{\left|\{w\notin \mathrm{Sub}(u)\cup\mathrm{Anc}(u)\}\right|}\times 100\%.
\end{equation}
\end{definition}
$\HASViol(u)=0$ for every non-root non-leaf $u\in V$ if and only if \HAS (\Cref{def:has}) holds. We are now ready to measure \HAS in CLIP embeddings.

\begin{figure}[t]
\centering
\begin{minipage}[c]{0.025\linewidth}
  \centering
  \raisebox{2.7\baselineskip}{\rotatebox{90}{\small Text}}
\end{minipage}%
\hfill
\begin{subfigure}[c]{0.31\linewidth}
  \centering
  \includegraphics[width=\linewidth]{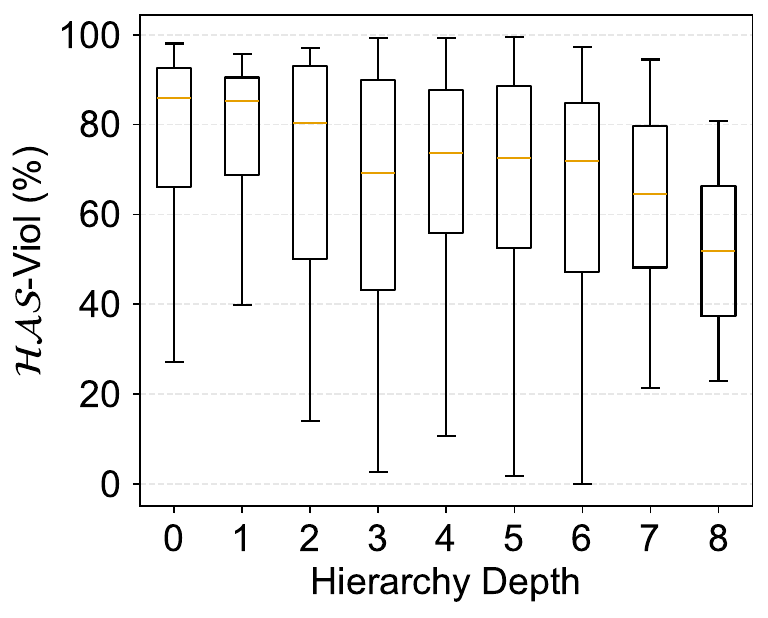}
  \caption{Randomly initialized.}
  \label{fig:has-text-violation-random}
\end{subfigure}
\hfill
\begin{subfigure}[c]{0.31\linewidth}
  \centering
  \includegraphics[width=\linewidth]{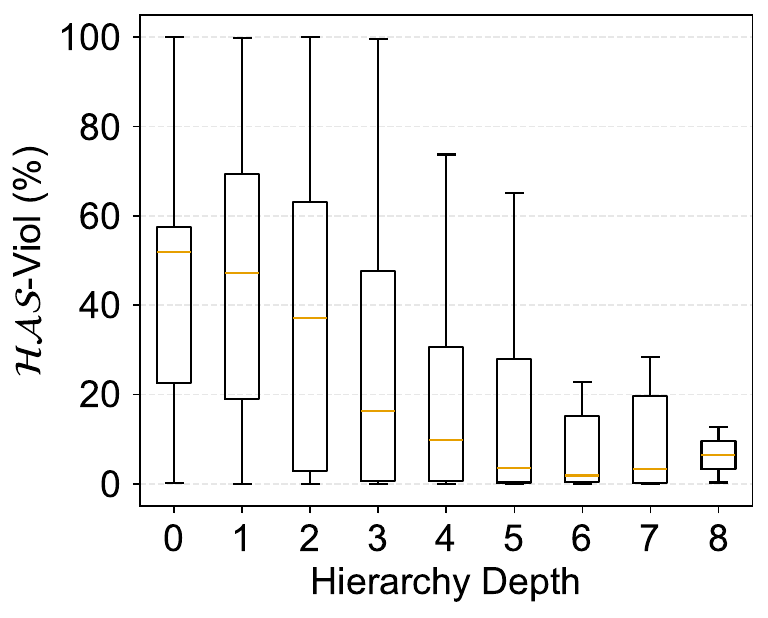}
  \caption{OpenAI CLIP.}
  \label{fig:has-text-violation-openai}
\end{subfigure}
\hfill
\begin{subfigure}[c]{0.31\linewidth}
  \centering
  \includegraphics[width=\linewidth]{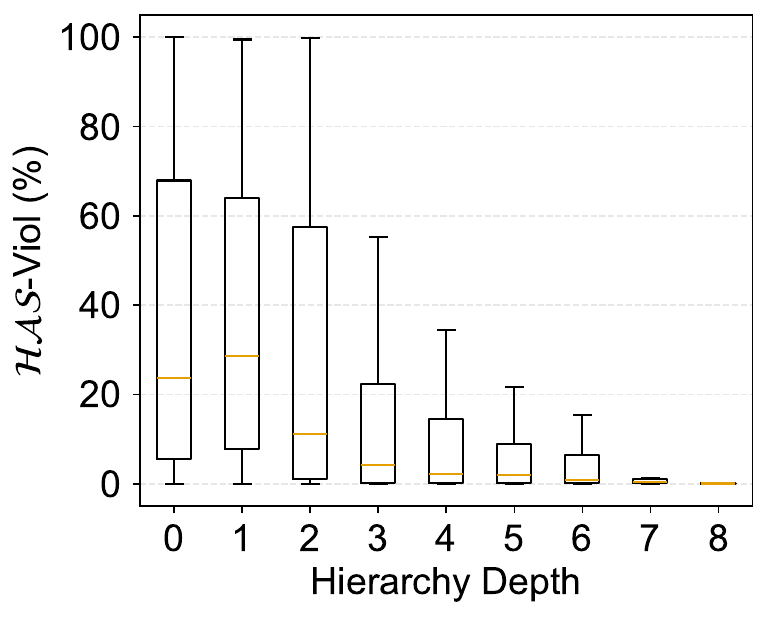}
  \caption{OpenCLIP.}
  \label{fig:has-text-violation-datacomp}
\end{subfigure}

\begin{minipage}[c]{0.025\linewidth}
  \centering
  \raisebox{2.7\baselineskip}{\rotatebox{90}{\small Image}}
\end{minipage}%
\hfill
\begin{subfigure}[c]{0.31\linewidth}
  \centering
  \includegraphics[width=\linewidth]{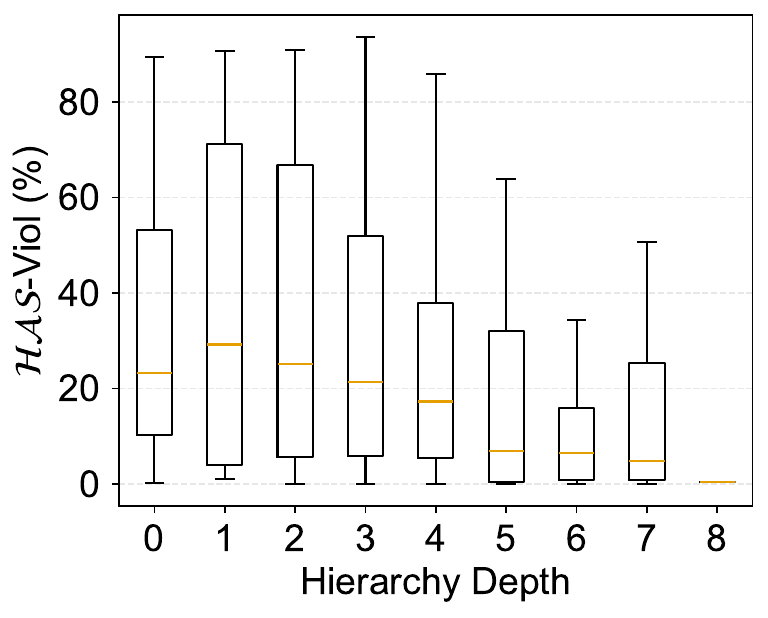}
  \caption{Randomly initialized.}
  \label{fig:has-image-violation-random}
\end{subfigure}
\hfill
\begin{subfigure}[c]{0.31\linewidth}
  \centering
  \includegraphics[width=\linewidth]{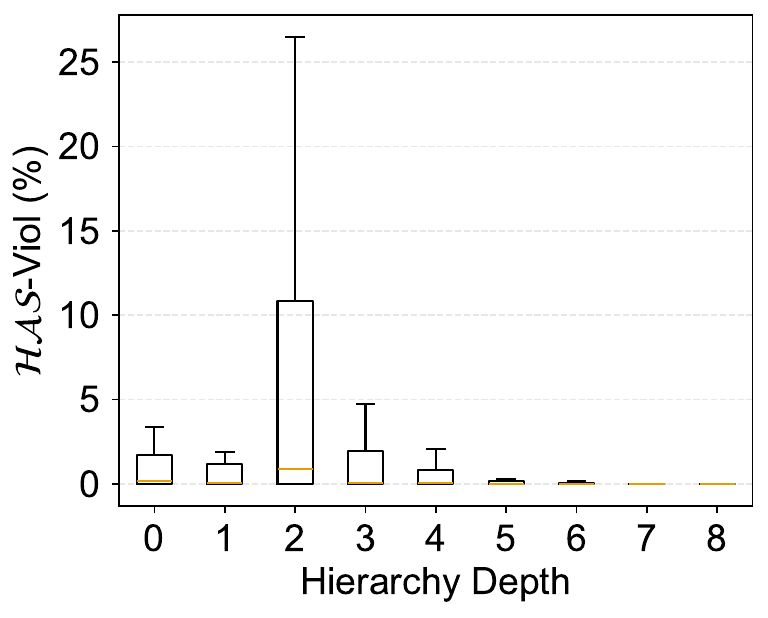}
  \caption{OpenAI CLIP.}
  \label{fig:has-image-violation-openai}
\end{subfigure}
\hfill
\begin{subfigure}[c]{0.31\linewidth}
  \centering
  \includegraphics[width=\linewidth]{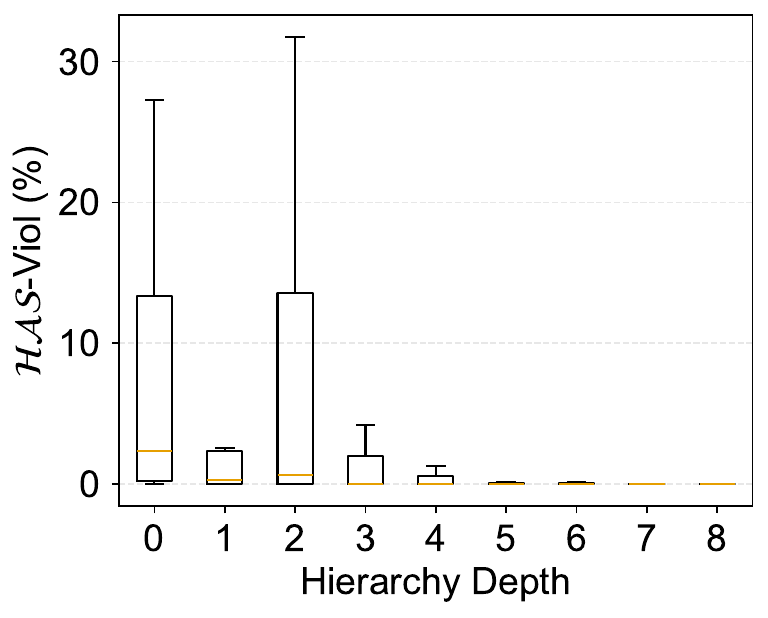}
  \caption{OpenCLIP.}
  \label{fig:has-image-violation-datacomp}
\end{subfigure}
\caption{Distribution of $\HASViol$ vs depth on ViT-L/14 CLIP embeddings over non-root and non-leaf nodes. 
\emph{Top:} \emph{text} embeddings show that \HAS is acquired through pretraining.
\emph{Bottom:} \emph{image} embeddings show approximate \HAS at random initialization and stronger \HAS after pretraining.}
\vspace{-1.00em}
\label{fig:has-violation}
\end{figure}

\subsection{Hierarchical Angular Separation in CLIP embeddings}\label{sec:hierarchical-angular-separation-empirical}

We measure \HASViol on the ImageNet-1K training split~\citep{deng2009imagenet} and the WordNet noun hierarchy~\citep{miller_wordnet_1995}.\footnote{Code is available at \url{https://github.com/nghiahhnguyen/hierarchical-angular-separation}.}
We evaluate both the \emph{text} and \emph{image} embeddings of two pretrained CLIP ViT-L/14 checkpoints (OpenAI CLIP~\citep{radford2021learning} and OpenCLIP~\citep{cherti2023reproducible} trained on DataComp-1B~\citep{gadre2023datacomp}) and a randomly initialized ViT-L/14.
To reduce the effect of outliers, we replace the descendant threshold $s_u=\min_{v\in\mathrm{Desc}(u)} \cos(\bh_u, \bh_v)$ by its $95$th percentile over $\mathrm{Desc}(u)$. Corresponding results on ViT-B/32, ViT-B/16, ViT-L/14@336px, SigLIP2~\citep{tschannen2025siglip2}, MetaCLIP2~\citep{chuang2025metaclip2}, and on the CUB-200 taxonomy~\citep{wah2011caltech}, are deferred to \Cref{app:additional-experiments-has}.

\myparagraph{Text embeddings: \HAS is acquired through pretraining}
In \Cref{fig:has-violation} (top), both pretrained text models have low violation percentages across depths, whereas the randomly initialized text model has substantially larger violations across the tree. Hence, for text concept embeddings, \HAS is a \emph{learned} property of contrastive pretraining. Additional results estimating concept embeddings by instance aggregation reach the same conclusion (\Cref{app:additional-experiments-has}).

\myparagraph{Image embeddings: \HAS is present in raw pixels and strengthened by pretraining}
In \Cref{fig:has-violation} (bottom), pretrained image embeddings have violation percentages concentrated near zero across depths. The randomly initialized ViT-L/14 has substantially lower violations than randomly initialized text embeddings, although its violations remain non-negligible.

\begin{table}[H]
\centering
\setlength{\tabcolsep}{5pt}
\caption{Mean image-side \HASViol. Each representation is paired with its own hierarchy-shuffled null. Null values report mean $\pm$ standard deviation over five hierarchy shuffles.}
\label{tab:raw-pixel-has}
\begin{tabular}{lcc}
\toprule
Representation & \HASViol (\%) $\downarrow$ & \HASViol shuffled (\%) $\downarrow$ \\
\midrule
Raw pixels & $26.51$ & $50.95 \pm 0.29$ \\
Randomly initialized ViT-L/14 & $27.88 \pm 0.24$ & $56.80 \pm 0.42$ \\
Pretrained OpenAI CLIP ViT-L/14 & $2.95$ & $32.93 \pm 0.48$ \\
i.i.d. Gaussian vectors ($d=768$) & $76.02 \pm 0.97$ & $76.33 \pm 0.93$ \\
\bottomrule
\end{tabular}
\end{table}

In \Cref{tab:raw-pixel-has}, raw pixels and randomly initialized image embeddings separate from their own hierarchy-shuffled nulls, whereas i.i.d. Gaussian vectors do not. Approximate image-side \HAS therefore comes from raw image data, remains after a randomly initialized encoder, and strengthens with contrastive pretraining, while text-side \HAS is acquired only through pretraining.

\section{Provable Transfer of Hierarchical Angular Separation from Image to Text Embeddings}\label{sec:hierarchical-angular-transfer}

\begin{figure}[t]
\centering
\definecolor{theoryblue}{HTML}{0072B2}
\definecolor{theoryorange}{HTML}{E69F00}
\begin{tikzpicture}[
  box/.style={rounded corners=5pt, line width=0.7pt, align=center, text width=0.42\linewidth, minimum height=1.2cm, inner sep=5pt},
  linear/.style={box, draw=theoryblue, fill=theoryblue!8},
  nce/.style={box, draw=theoryorange!80!black, fill=theoryorange!12},
  arrow/.style={-{Latex[length=3mm,width=2mm]}, line width=1.2pt}
]
\node[linear] (linear) at (-3.5,0) {\textbf{Linear contrastive loss}\\Exact prototype decomposition\\\Cref{cor:linear-sampled-path-noisy-centroid-form}};
\node[nce] (nce) at (3.5,0) {\textbf{InfoNCE loss}\\Regularized prototype decomposition\\\Cref{prop:large-lambda,cor:prototype-form}};
\node[box, draw=black!55, fill=black!3, text width=0.68\linewidth] (shared) at (0,-2) {\textbf{Shared margin argument}\\Small perturbations preserve image-side triplet orderings with sufficient margin\\\Cref{lem:per-triplet-margin-preservation}};
\node[linear, fill=theoryblue!15] (linearresult) at (-3.5,-4) {\textbf{Linear loss transfer}\\\Cref{thm:linear-sampled-path-noisy-has}};
\node[nce, fill=theoryorange!22] (nceresult) at (3.5,-4) {\textbf{Regularized InfoNCE loss transfer}\\\Cref{thm:large-lambda-has-noisy}};
\draw[arrow, theoryblue] (linear) -- (shared);
\draw[arrow, theoryorange!80!black] (nce) -- (shared);
\draw[arrow, theoryblue] (shared) -- (linearresult);
\draw[arrow, theoryorange!80!black] (shared) -- (nceresult);
\end{tikzpicture}
\caption{Overview of the \HAS transfer results. Under assumptions on the data model, both losses decompose the text embeddings into prototypes plus perturbations. Bounding the perturbations relative to the prototype norms gives a bound on text-side violations in terms of image-side margins.}
\label{fig:theory-flow}
\end{figure}
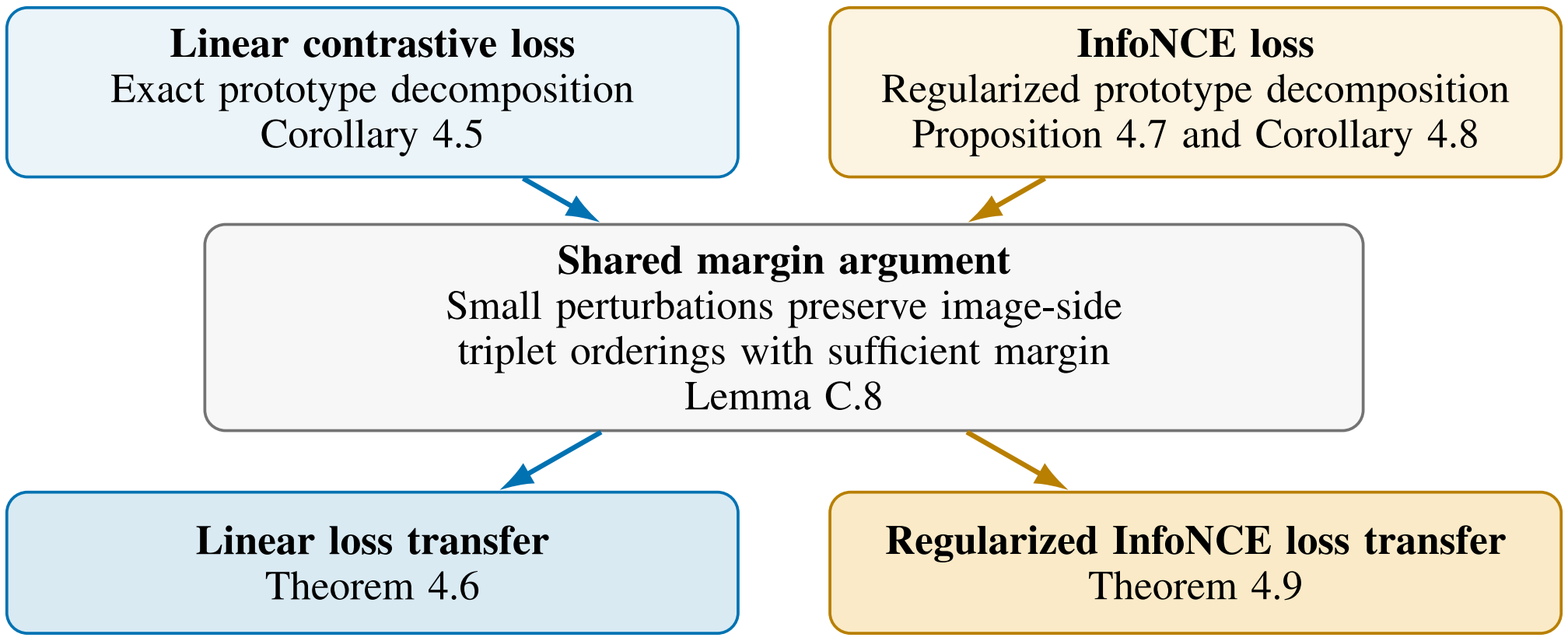

Motivated by the observed empirical asymmetry, we study theoretically how contrastive learning transfers pre-existing image geometry to text. We use an Unconstrained Features Model (UFM) in which the image embeddings are fixed and only the text embeddings are learned. Under hierarchical co-occurrence, each text concept aggregates images from its descendant leaves, and contrastive learning therefore aligns the text embedding with the corresponding root-centered image prototype. We first derive this mechanism in closed form for a linear contrastive loss (\Cref{sec:linear-sampled-path}), then show that regularized symmetric InfoNCE inherits it under common assumptions (\Cref{sec:has-infonce}). We summarize the relation between these theoretical results in \Cref{fig:theory-flow}. We also show that the transfer is robust to weaker regularization than required by theory (\Cref{sec:synthetic-experiments}).

\subsection{Setup and Notation}\label{sec:has-setup}

\myparagraph{Hierarchical Image Prototypes and Samples}
Continuing the notation of \Cref{sec:hierarchical-angular-separation-definition}, we consider a tree $\mathcal{T}=(V,E)$ together with an embedding $\{\bm{\mu}_k\}_{k\in V}$ of its nodes, as illustrated in \Cref{fig:path-consistent-caption} (top-left). Each vector $\bm{\mu}_k\in\R^d$, for $k\in V$, is called an \emph{image prototype}.

Our transfer analysis compares the image-side margin of each prototype triplet with the perturbations introduced by sampling and optimization. Because image embedding instances deviate from their prototypes, we model the resulting finite-sample error as follows.
We sample exactly $n$ images per leaf, for a total of $N=n|V_{\mathrm{leaf}}|$ images.

\begin{assumption}[Prototype variation model]\label{ass:noisy-prototype}
The image embeddings $\{\bz_j\}_{j=1}^N$ are sampled as $\bz_j=\bm{\mu}_{y_j} + \bm{\varepsilon}_j$, where $y_j\in V_{\mathrm{leaf}}$ is the leaf node associated with image $j$ and $\{\bm{\varepsilon}_j\}_{j=1}^N$ are independent, zero-mean, $\sigma$-sub-Gaussian random vectors in $\R^d$.
\end{assumption}

\myparagraph{Hierarchical captions determine each concept's positive images}
For each node $u\in V$, let $\mathrm{Path}(u)$ denote the root-to-$u$ path.
Each image $j\in[N]$ is paired with a fixed observed caption node $c_j\in\mathrm{Path}(y_j)$ (\Cref{fig:path-consistent-caption}, top-right). For example, an image from the leaf \texttt{labrador} may be captioned as \texttt{labrador}, \texttt{dog}, \texttt{mammal}, or \texttt{animal}.

For each node $k\in V$, define its empirical positive set and caption count by
\begin{equation}\label{eq:caption-positive-sets}
  \cP_k := \{j\in[N] : c_j=k\},
  \qquad
  N_k := |\cP_k|.
\end{equation}
As shown in \Cref{fig:path-consistent-caption} (bottom-left), the positive set of a non-leaf concept such as \texttt{dog} can pool images across descendant leaves such as \texttt{labrador} and \texttt{corgi}.

Hierarchical co-occurrence distinguishes our UFM from non-hierarchical UFM analyses. Prior UFM work for cross-entropy or supervised contrastive learning assumes mutually exclusive class labels and analyzes the resulting Simplex ETF geometry~\citep{zhu2021geometric,mixon2022neural,grafDissectingSupervisedContrastive2021,behniaSupervisedContrastiveRepresentation2024}. Here, each image still has one caption, but that caption may be any concept along $\mathrm{Path}(y_j)$ rather than only the mutually exclusive leaf $y_j$. The resulting dependencies across nodes allow each optimum to depend on subtree prototype averages, making \HAS, rather than a flat simplex, the relevant geometry.

\myparagraph{The UFM optimizes only the text embeddings}
Following the Unconstrained Features Model (UFM) approach~\citep{mixon2022neural}, we fix the observed image embeddings $\bZ=[\bz_1,\ldots,\bz_N]\in\R^{d\times N}$ and caption labels $\{c_j\}_{j=1}^N$, and optimize the text concept embeddings $\bA=[\ba_1,\ldots,\ba_M]\in\R^{d\times M}$. For a regularization coefficient $\lambda>0$, the objective is

\begin{equation}\label{eq:objective-generic}
  \min_{\bA}\; \cL_{\mathrm{contr}}(\bA;\bZ)
  \;+\; \frac{\lambda}{2}\|\bA\|_F^2.
\end{equation}

\begin{remark}[Relation to CLIP's normalized formulation]\label{rem:normalization}
In practice, CLIP normalizes embeddings to the unit sphere, which bounds logits and guarantees finite minimizers.
We instead use unnormalized embeddings with $L_2$ regularization, following the standard UFM approach~\citep{mixon2022neural, grafDissectingSupervisedContrastive2021, behniaSupervisedContrastiveRepresentation2024}.
This substitution is motivated by tractability: normalization introduces non-convexity, which makes analysis more complex. The regularization coefficient $\lambda$ prevents unbounded growth of embedding norms as a normalization operator does.
\end{remark}

Next, we study the geometry of $\bA^*$ at the global minimizers of the contrastive objective \eqref{eq:objective-generic}. In \Cref{sec:linear-sampled-path}, we first analyze when the contrastive loss is a linear contrastive loss, which admits an exact minimizer in closed form. We use the closed-form minimizer to identify which image-side triplet orderings are transferred and thereby bound the text-side \HAS violation percentage.

\subsection{Linear Contrastive Loss Analysis}\label{sec:linear-sampled-path}
To get a closed-form analysis of \HAS transfer in a contrastive learning setting, we first consider the linear contrastive loss of \citet{jiPowerContrastFeature2023}.
The symmetric linear contrastive loss $\cL^{\mathrm{lin}}(\bA)$ is
\begin{align}\label{eq:objective-linear-sampled-path}
  -\frac{1}{2}\sum_{j=1}^{N}\left[
    \ba_{c_j}^\top \bz_j
    - \frac{1}{M-1}\sum_{\substack{k\in[M]\\ k\neq c_j}}\ba_k^\top \bz_j
    + \ba_{c_j}^\top \bz_j
    - \frac{1}{N-1}\sum_{\substack{j'\in[N]\\ j'\neq j}}\ba_{c_j}^\top \bz_{j'}
  \right]
  + \frac{\lambda}{2}\|\bA\|_F^2.
\end{align}

Solving the first-order condition for \Cref{eq:objective-linear-sampled-path} gives a unique minimizer. For each node $k\in V$ with $N_k>0$, define the empirical mean of the images captioned $k$ as $\bar{\bz}_k:=\frac{1}{N_k}\sum_{j\in\cP_k}\bz_j$, and define the global empirical image mean $\bar{\bz}:=\frac{1}{N}\sum_{j=1}^N\bz_j$. The optimal text embedding for $k$ is a \textit{weighted combination} of these two means (\Cref{prop:linear-sampled-path-minimizer,cor:linear-sampled-path-centroid-form} in \Cref{app:proofs-linear-sampled-path}). The transfer proof therefore reduces to identifying the prototype component of each $\bar{\bz}_k$. We first make this identification exact with two simplifying assumptions: leaf-uniform coverage makes each mean uniform over descendant leaves, and prototype consistency identifies that average with the corresponding node prototype.
\begin{assumption}[Leaf-uniform caption coverage]\label{ass:leaf-uniform-caption}
Every non-root node $k\in V\setminus\{\mathrm{root}\}$ has $N_k>0$, and its positive set $\cP_k$ contains the same number of images from each leaf in $\mathrm{Sub}(k)\cap V_{\mathrm{leaf}}$.
\end{assumption}
\begin{assumption}[Prototype consistency]\label{ass:prototype-centroid-consistency}
For every node $k\in V$, the average of the prototypes over leaves in $\mathrm{Sub}(k)\cap V_{\mathrm{leaf}}$ is exactly $\bm{\mu}_k$.
\end{assumption}

Using \Cref{ass:leaf-uniform-caption,ass:prototype-centroid-consistency,ass:noisy-prototype}, we can write $\bar{\bz}_k=\bm{\mu}_k+\bar{\bm{\varepsilon}}_k$ and $\bar{\bz}=\bm{\mu}_{\mathrm{root}}+\bar{\bm{\varepsilon}}$, where $\bar{\bm{\varepsilon}}_k:=\frac{1}{N_k}\sum_{j\in\cP_k}\bm{\varepsilon}_j$ and the global noise average is $\bar{\bm{\varepsilon}}:=\frac{1}{N}\sum_{j=1}^N\bm{\varepsilon}_j$, yielding the following prototype decomposition for the minimizer.
The proof is deferred to \Cref{app:proofs-linear-sampled-path}.

\begin{restatable}[Prototype Decomposition for the Linear Contrastive Loss]{corollary}{LinearSampledPathNoisyCentroidFormCor}\label{cor:linear-sampled-path-noisy-centroid-form}
Suppose \Cref{ass:leaf-uniform-caption,ass:prototype-centroid-consistency,ass:noisy-prototype} hold. For each non-root $k$, define $\xi_k^{\mathrm{lin}}:=\frac{N\!\left(\frac{1}{2(M-1)}+\frac{N_k}{2(N-1)}\right)}{N_k\!\left(1+\frac{1}{2(M-1)}+\frac{1}{2(N-1)}\right)}$. Then every such $k$ satisfies
\begin{align}\label{eq:linear-sampled-path-noisy-centroid-form}
  \ba_k^{\mathrm{lin},*}
  &=
  \frac{N_k}{\lambda}\left(
    1
    +
    \frac{1}{2(M-1)}
    +
    \frac{1}{2(N-1)}
  \right)\left(
    \bm{\mu}_k
    +
    \bar{\bm{\varepsilon}}_k
    -
    \xi_k^{\mathrm{lin}}\left(
      \bm{\mu}_{\mathrm{root}}
      +
      \bar{\bm{\varepsilon}}
    \right)
  \right).
\end{align}
\end{restatable}

\begin{figure}[t]
\centering
\begin{subfigure}{0.49\linewidth}
  \includegraphics[width=\linewidth]{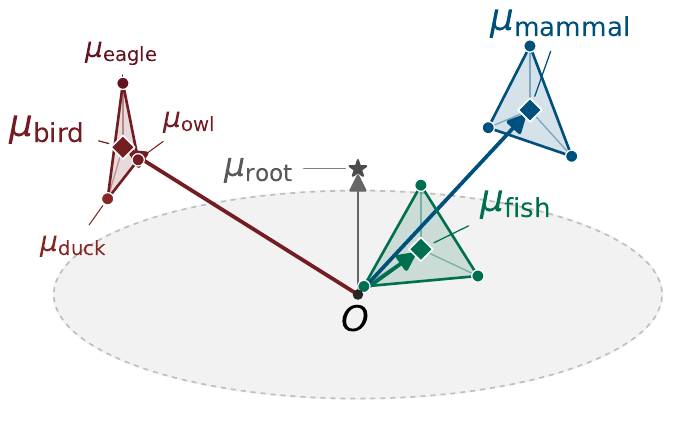}
  \caption{Fixed image prototypes.}
\end{subfigure}
\hfill
\begin{subfigure}{0.49\linewidth}
  \includegraphics[width=\linewidth]{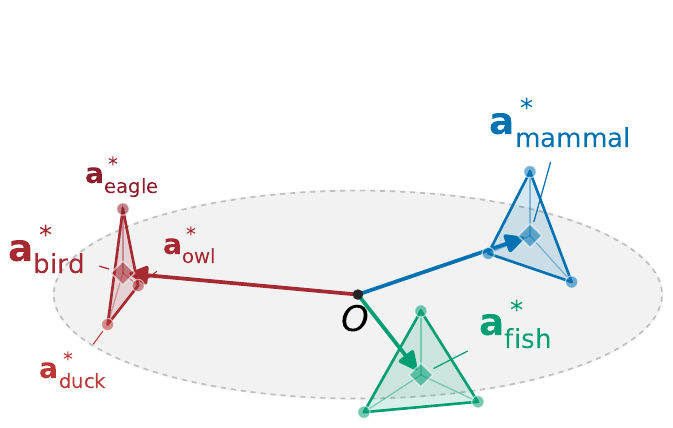}
  \caption{Optimal text geometry.}
\end{subfigure}
\caption{Optimal text geometry for a stylized hierarchy. Each colored triangle is a regular simplex of child concepts, with its parent (diamond) at the centroid. The optimal text directions in (b) satisfy $\ba_k^{\mathrm{lin},*}\propto\bm{\mu}_k-\xi_k^{\mathrm{lin}}\bm{\mu}_{\mathrm{root}}$, where $\xi_k^{\mathrm{lin}}$ is the root-centering coefficient. Each child simplex remains centered at its parent, and \HAS is preserved.}
\label{fig:toy-optimal-geometry}
\end{figure}

In \Cref{cor:linear-sampled-path-noisy-centroid-form}, $\ba_k^{\mathrm{lin},*}$ is a positive rescaling of $\bm{\mu}_k$ plus a perturbation (\Cref{fig:path-consistent-caption}, bottom-right). The perturbation consists of the deterministic root-centering term $-\xi_k^{\mathrm{lin}}\bm{\mu}_{\mathrm{root}}$, which shifts each prototype along the root direction (\Cref{fig:toy-optimal-geometry}), and the stochastic variation term $\bar{\bm{\varepsilon}}_k-\xi_k^{\mathrm{lin}}\bar{\bm{\varepsilon}}$, which is bounded with high probability by \Cref{ass:noisy-prototype}. Positive rescaling preserves cosine similarity, so transfer depends on how much the perturbation changes each prototype triplet margin. We therefore bound the perturbation relative to the prototype norms and identify which triplet orderings remain intact.
Let $\rho_{\min}:=\min_{k\in V\setminus\{\mathrm{root}\}}\|\bm{\mu}_k\|$ denote the minimum non-root prototype norm.
With $\xi_{k}^{\mathrm{lin}}$ from \Cref{cor:linear-sampled-path-noisy-centroid-form}, define $\xi_{\max}^{\mathrm{lin}}:=\max_{k\in V\setminus\{\mathrm{root}\}}|\xi_k^{\mathrm{lin}}|$ and $\delta_{\mathrm{root}}:=\xi_{\max}^{\mathrm{lin}}\|\bm{\mu}_{\mathrm{root}}\|$, the largest root-centering perturbation norm.

To compare violations across modalities, for a non-root non-leaf anchor $u\in V$, let $\HASViol^{\mathrm{txt}}(u)$ denote \HASViol evaluated on the optimal text embeddings.
For non-zero non-root image prototypes and $\tau\ge0$, let $\HASViol^{\mathrm{img}}_\tau(u)$ denote the same metric evaluated on the prototypes with their threshold $s_u$ lowered to $s_u-\tau$.
Equivalently, $\HASViol^{\mathrm{img}}_\tau(u)$ is the percentage of off-branch nodes at $u$ whose worst-case triplet margin on the image prototypes is at most $\tau$.

\begin{restatable}[\HAS Transfer for the Linear Contrastive Loss]{theorem}{LinearSampledPathNoisyHASThm}\label{thm:linear-sampled-path-noisy-has}
Suppose \Cref{ass:noisy-prototype} (sub-Gaussian prototype variation), \Cref{ass:leaf-uniform-caption} (leaf-uniform caption coverage), and \Cref{ass:prototype-centroid-consistency} (prototype consistency) hold, and suppose $\rho_{\min}>0$.
Let $N_{\min}:=\min_{k\in V\setminus\{\mathrm{root}\}}N_k$ be the minimum non-root caption count.
There exists a universal constant $C>0$ such that the following holds for every $\epsilon\in(0,1)$.
Define the combined error term
\begin{equation}\label{eq:linear-sampled-path-noisy-has-bound}
  D_\epsilon^{\mathrm{lin}}:=\delta_{\mathrm{root}}+C\sigma\left(\sqrt{\tfrac{d+\log(M-1)+\log(4/\epsilon)}{N_{\min}}}+\xi_{\max}^{\mathrm{lin}}\sqrt{\tfrac{d+\log(4/\epsilon)}{N}}\right).
\end{equation}
If $D_\epsilon^{\mathrm{lin}}<\rho_{\min}/4$, then, with probability at least $1-\epsilon$, every non-root non-leaf anchor node $u$ satisfies
\begin{equation}\label{eq:per-triplet-has-bound}
  \HASViol^{\mathrm{txt}}(u)\le\HASViol^{\mathrm{img}}_{\tau_\epsilon^{\mathrm{lin}}}(u),
  \qquad \tau_\epsilon^{\mathrm{lin}}:=\frac{8D_\epsilon^{\mathrm{lin}}}{\rho_{\min}}.
\end{equation}
\end{restatable}
We defer the proof to \Cref{app:proof-linear-sampled-path-noisy-has}.
Under the assumptions, every image-side triplet with margin greater than $\tau_\epsilon^{\mathrm{lin}}$ remains correctly ordered on the text side. Only triplets with image-side margin at most $\tau_\epsilon^{\mathrm{lin}}$ can violate text-side \HAS.
We now turn to the InfoNCE loss~\citep{oordRepresentationLearningContrastive2018}.

\subsection{InfoNCE Loss Analysis}\label{sec:has-infonce}
In the same fixed-image setup, we test whether the transfer mechanism exposed by the linear loss persists under regularized symmetric InfoNCE~\citep{oordRepresentationLearningContrastive2018}:
\begin{align}\label{eq:objective-simplified-finite}
  \cL(\bA) &:= \sum_{j=1}^N \frac{1}{2}\Bigg(
    -\ba_{c_j}^\top \bz_j + \log\sum_{k=1}^M e^{\ba_k^\top \bz_j}
    -\ba_{c_j}^\top \bz_j + \log\sum_{j'=1}^N e^{\ba_{c_j}^\top \bz_{j'}}
  \Bigg) + \frac{\lambda}{2}\|\bA\|_F^2.
\end{align}
The objective is $\lambda$-strongly convex, so its unique minimizer $\bA^{\mathrm{nce},*}$ satisfies the softmax fixed-point equation (\Cref{prop:stationarity}). Because this equation has no closed form, we analyze $\bA^{\mathrm{nce},*}$ as $\lambda \to \infty$.

\begin{restatable}[Large-Regularization Expansion]{proposition}{LargeLambdaExpansionProp}\label{prop:large-lambda}
As $\lambda \to \infty$, for any $k$ with $N_k>0$,
\begin{equation}\label{eq:large-lambda}
  \ba_k^{\mathrm{nce},*}
  =
  \frac{N_k}{\lambda}\bar{\bz}_k
  -
  \frac{1}{\lambda}\left(\frac{N}{2M} + \frac{N_k}{2}\right)\bar{\bz}
  + O(\lambda^{-2}).
\end{equation}
\end{restatable}
As $\lambda$ grows, $\bA^{\mathrm{nce},*}$ shrinks and the softmax weights approach uniform. The fixed-point equation then yields \Cref{eq:large-lambda}. We defer the proof to \Cref{app:proofs-has-infonce-nonthm}.
Substituting $\bar{\bz}_k=\bm{\mu}_k+\bar{\bm{\varepsilon}}_k$ and $\bar{\bz}=\bm{\mu}_{\mathrm{root}}+\bar{\bm{\varepsilon}}$ into \Cref{eq:large-lambda} gives the same prototype decomposition as the linear-loss minimizer.
\begin{corollary}[Prototype Decomposition for the Large-Regularization Expansion]\label{cor:prototype-form}\label{cor:prototype-form-noisy}
Suppose \Cref{ass:leaf-uniform-caption,ass:prototype-centroid-consistency,ass:noisy-prototype} hold. For each non-root $k$, define $\xi_k^{\mathrm{nce}}:=\left(\frac{N}{2M}+\frac{N_k}{2}\right)/N_k$. Then every such $k$ satisfies
\begin{equation}\label{eq:large-lambda-prototype-noisy}
  \ba_k^{\mathrm{nce},*}
  =
  \frac{N_k}{\lambda}\left(
    \bm{\mu}_k
    +
    \bar{\bm{\varepsilon}}_k
    -
    \xi_k^{\mathrm{nce}}\left(
      \bm{\mu}_{\mathrm{root}}
      +
      \bar{\bm{\varepsilon}}
    \right)
  \right)
  + O(\lambda^{-2}).
\end{equation}
\end{corollary}
We defer the proof to \Cref{app:prototype-form}.
The leading term has the same form as \Cref{cor:linear-sampled-path-noisy-centroid-form} with $\xi_k^{\mathrm{nce}}$ in place of $\xi_k^{\mathrm{lin}}$, so the same margin argument gives the following \HAS transfer guarantee for InfoNCE.
For $\epsilon\in(0,1)$, define $D_\epsilon^{\mathrm{nce}}$ by recomputing the error term in \Cref{eq:linear-sampled-path-noisy-has-bound} with $\xi_k^{\mathrm{nce}}$ in place of $\xi_k^{\mathrm{lin}}$ throughout, including in $\xi_{\max}^{\mathrm{lin}}$ and $\delta_{\mathrm{root}}$, and with $\log(8/\epsilon)$ in place of $\log(4/\epsilon)$.

\begin{restatable}[\HAS Transfer for InfoNCE]{theorem}{LargeLambdaNoisyHASThm}\label{thm:large-lambda-has-noisy}
Suppose \Cref{ass:noisy-prototype,ass:leaf-uniform-caption,ass:prototype-centroid-consistency} hold and $\rho_{\min}>0$, and fix $\epsilon\in(0,1)$.
If $0<D_\epsilon^{\mathrm{nce}}<\rho_{\min}/8$, then there is a deterministic threshold $\lambda_0(\epsilon)>0$ such that, for every $\lambda\ge\lambda_0(\epsilon)$, with probability at least $1-\epsilon$, every non-root non-leaf anchor node $u$ satisfies \Cref{eq:per-triplet-has-bound} with $\tau_\epsilon^{\mathrm{lin}}$ replaced by $\tau_\epsilon^{\mathrm{nce}}:=16D_\epsilon^{\mathrm{nce}}/\rho_{\min}$.
\end{restatable}
Thus, regularized InfoNCE inherits the linear-loss transfer mechanism. The large-$\lambda$ condition is required for the analysis but not for the phenomenon: in \Cref{sec:synthetic-experiments} we show that \HAS transfer persists under substantially weaker regularization.
We defer the proof to \Cref{app:proof-large-lambda-has-noisy}. 

\begin{remark}[Connection to the modality gap]\label{rem:modality-gap}
CLIP embeds images and text in separated regions of the shared space, which is known as the \emph{modality gap}~\citep{liangMindGapUnderstanding2022}. Our analysis offers an explanation to this phenomenon. 
In \Cref{eq:linear-sampled-path-noisy-centroid-form,eq:large-lambda-prototype-noisy}, each optimal text embedding subtracts a multiple of the global image mean $\bm{\mu}_{\mathrm{root}}$. Instead of aligning with the global image mean, the text embeddings subtract it, so the two modalities remain separated at the text optimum.
\end{remark}

\myparagraph{Extensions beyond the basic transfer model}
In \Cref{app:caption-relaxation}, we quantify how caption imbalance, inexact prototype consistency, and ancestor omission affect the linear-loss transfer guarantee.
Caption imbalance and prototype inconsistency add deterministic error terms, while independent ancestor omissions reduce the effective sample size and introduce random imbalance.
In \Cref{app:alternating}, we analyze one exact image-side update after the text-side update under the linear loss, and show that averaging over images sharing the same caption reduces stochastic variation in the updated image embeddings.

\myparagraph{Uncaptioned concepts limit hierarchical transfer}
In our fixed-image UFM, both regularized losses treat non-root concepts absent from positive captions symmetrically.
Their unique minimizers therefore assign these concepts identical text embeddings (\Cref{prop:total-ancestor-omission}).
If an anchor has an uncaptioned descendant and an uncaptioned off-branch concept, the cosine similarity between their embeddings cannot satisfy the \HAS condition.
Thus, hierarchical image geometry and the caption co-occurrences are both necessary for the successful transfer of hierarchical structure.
Our result therefore resonates with the view that data coverage is a key factor in successful compositional generalization.

\myparagraph{Limitations}\label{sec:limitations}
Our analysis explains how pre-existing image geometry transfers to text under a simplified model of contrastive learning.
For tractability, we fix the image embeddings and replace CLIP's unit-norm constraints with quadratic regularization (\Cref{rem:normalization}).
The transfer guarantees require perturbations small relative to the prototype norms; for InfoNCE, we additionally require sufficiently large $\lambda$.
In \Cref{sec:synthetic-experiments}, we observe transfer under weaker regularization, but a theoretical guarantee in that regime remains open.
Beyond these fixed-image guarantees, our analysis of one image-side update does not explain the full joint training dynamics or how approximate \HAS becomes exact.
Explaining why hierarchical geometry is already present in image data also remains outside the scope of our transfer analysis.
We discuss further limitations in \Cref{app:model-limitations}.

\subsection{Synthetic Validation at Small Regularization}\label{sec:synthetic-experiments}

\begin{figure}[t]
\centering
\begin{subfigure}{0.49\linewidth}
  \centering
  \includegraphics[width=\linewidth]{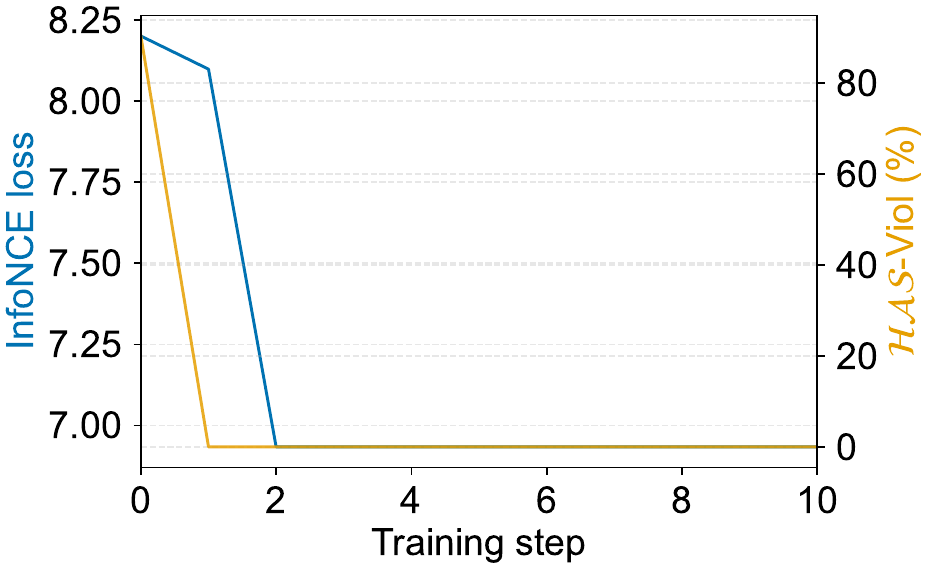}
  \caption{Convergence at $\lambda=10^{-3}$.}
  \label{fig:synthetic-convergence-trace}
\end{subfigure}
\hfill
\begin{subfigure}{0.49\linewidth}
  \centering
  \includegraphics[width=\linewidth]{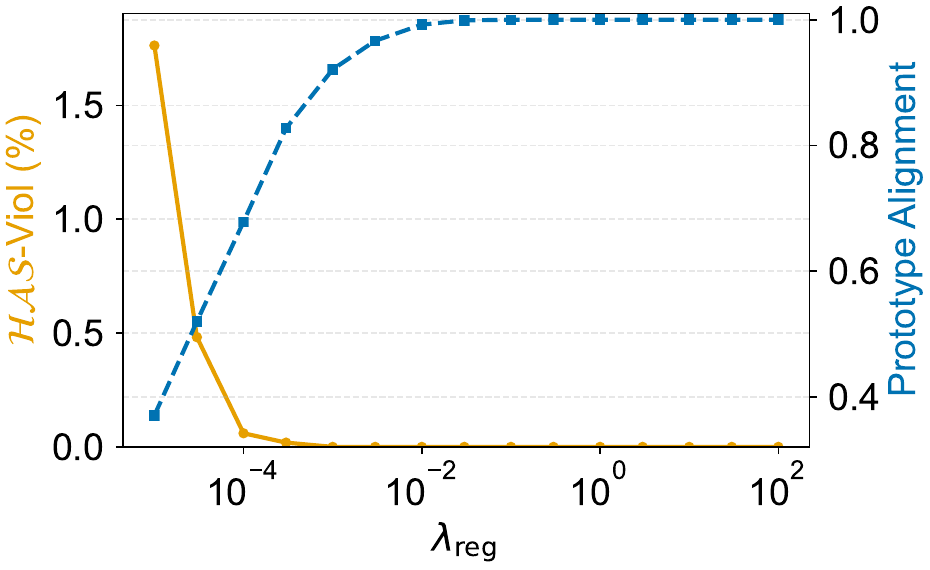}
  \caption{Regularization sweep over $\lambda\in[10^{-5},10^{2}]$.}
  \label{fig:synthetic-lambda-sweep}
\end{subfigure}
\caption{Synthetic validation of \HAS transfer at small regularization with the InfoNCE loss. \emph{(a)} L-BFGS convergence at $\lambda=10^{-3}$: the InfoNCE loss without the regularization term and the text-side $\HASViol$ against the training step. \emph{(b)} At the minimizer, text-side $\HASViol$ reaches zero at smaller $\lambda$ than is needed for prototype alignment to saturate.}
\label{fig:synthetic-validation}
\end{figure}
We test whether \HAS transfer persists beyond the sufficient conditions used in our InfoNCE analysis.
In \Cref{thm:large-lambda-has-noisy}, we guarantee transfer only for sufficiently large regularization $\lambda$ and sufficiently small perturbations.
On a synthetic image-prototype hierarchy, we minimize the InfoNCE objective \eqref{eq:objective-simplified-finite} with L-BFGS over $\lambda\in[10^{-5},10^{2}]$.
At each minimizer, we report $\HASViol$ and \emph{prototype alignment}, the minimum over non-root $k$ of $\cos(\ba_k^{\mathrm{nce},*},\bm{\mu}_k)$.
The full setup is in \Cref{sec:exp-low-lambda}.

\HAS transfer occurs under weaker regularization than required by the analysis.
As $\lambda$ increases, violations vanish at the minimizer before prototype alignment saturates (\Cref{fig:synthetic-lambda-sweep}).
The convergence trace confirms that zero violations persist at the minimizer found by L-BFGS (\Cref{fig:synthetic-convergence-trace}).
\HAS requires preserving similarity orderings, so transfer can occur before text embeddings align with image prototypes.

Transfer also occurs outside the sufficient perturbation bound: the root norm in this experiment violates the bound, yet $\HASViol$ reaches zero.
Violations also remain small across different root norms and variation strengths (\Cref{sec:exp-low-lambda}).

To test whether transfer still depends on hierarchical co-occurrence, we permute hypernym--image pairings while preserving caption frequencies.
Permutation substantially increases text violations under both the linear and InfoNCE losses (\Cref{app:caption-permutation}), so caption frequencies alone does not imply transfer.

\section{Conclusion}\label{sec:conclusion}

In this paper, we studied how concept embeddings learned without explicit hierarchical supervision encode the underlying concept hierarchy. 
We measured this property with \emph{Hierarchical Angular Separation} (\HAS): for each non-leaf concept, a node has a higher cosine similarity to its descendant concepts than to concepts outside its subtree. 
We first showed empirically that \HAS holds approximately in CLIP text embeddings only after pretraining, while it is already approximately present in image embeddings at random initialization.
To explain this phenomenon, we analyzed an Unconstrained Features Model in which we freeze the image embeddings and train the text embeddings using either a linear contrastive loss or the InfoNCE loss with sufficient regularization. 
Under hierarchical co-occurrence, we proved that the text-side \HAS violation percentage at each anchor is bounded by the image-side \HAS violation and near-violation percentage.
Numerical experiments on synthetic concept hierarchies confirm these predictions and indicate that \HAS is induced even with weaker regularization than theoretically required.
Our results suggest that multimodal contrastive learning transfers hierarchical structure already present in image geometry to text embeddings.
Whether this mechanism extends beyond hierarchy to sparse, low-rank, or other compositional structures, and why image data exhibit hierarchy in the first place, remain open.

\newpage
\begin{ack}
The authors thank Ryan Chan, Ryan Pilgrim, Uday Kiran Tadipatri, Buyun Liang, Kyle Poe, Leandro Palma, Yi Ma, and anonymous reviewers for their helpful comments and suggestions.
This work was funded by Simons Foundation grant 814201, NSF grant 2031985, and University of Pennsylvania startup funds.
\end{ack}

{
    \small
    \bibliographystyle{unsrtnat}
    \bibliography{strings,main,nghia}
}

\clearpage
\appendix
\raggedbottom
\crefalias{section}{appendix}\crefalias{subsection}{appendix}\crefalias{subsubsection}{appendix}
\tableofcontents

\section{Formalizations of Hierarchy in Embedding Spaces}\label{app:formalizations-hierarchy}

Several prior works have formalized how an embedding of a rooted tree \emph{encodes} the hierarchy.
We group these formalizations into two categories: 
\begin{itemize}
    \item \emph{Prescriptive} formalizations, which incorporate an explicit geometric constraint (e.g., a cone with a chosen angle) into the training objective to enforce it.
    \item \emph{Descriptive} formalizations, which characterize a property of the learned embeddings without building it into the loss.
\end{itemize}
Our \Cref{def:has} is descriptive: We state a pairwise cosine-similarity inequality to be verified for trained embeddings.
We compare with the closest prior work in \Cref{app:entailment-cones} and note other formalizations in \Cref{app:other-formalizations}.

\subsection{Hyperbolic Entailment Cones}\label{app:entailment-cones}

The most comparable prior formalization is the hyperbolic entailment cone line of work~\citep{ganeaHyperbolicEntailmentCones2018, desaiHyperbolicImageTextRepresentations2024, palCompositionalEntailmentLearning2025}.
Each non-root embedding $\bx$ is assigned a convex cone $\cC_{\bx}$ with a half-angle $\psi(\bx)$ specified as a function of $\bx$,
\footnote{\citet{ganeaHyperbolicEntailmentCones2018} use $\psi(\bx)=\arcsin(K(1-\|\bx\|^2)/\|\bx\|)$ in the Poincar\'e ball. \citet{desaiHyperbolicImageTextRepresentations2024} (MERU) adapt this to the Lorentz model with learnable curvature $c$ via $\arcsin(2K/(\sqrt{c}\,\|\bx_{\text{space}}\|))$. \citet{palCompositionalEntailmentLearning2025} (HyCoCLIP) use the MERU angle with a multiplicative slack $\eta$.} 
and the loss penalizes violations of cone inclusion: descendants of $u$ must lie inside $\cC_{\ba_u}$, while non-descendants must lie outside.
The angle is thus \emph{determined in advance} and enforced as a training margin.

\HAS can be viewed as a Euclidean version of this condition, but \emph{descriptive} rather than prescriptive: the pairwise inequality $\cos(\ba_u,\ba_v)>\cos(\ba_u,\ba_w)$ is equivalent to strict cone containment with the angle determined by the nearest off-branch embedding, so the angle is \emph{determined by the data} rather than imposed by the loss.

\subsection{Other Formalizations}\label{app:other-formalizations}

\myparagraph{Poincar\'e embeddings}
\citet{nickelPoincareEmbeddingsLearning2017} learn embeddings in the Poincar\'e ball via a soft-ranking loss, yielding a hierarchical ordering of concepts (general concepts near the origin, specific ones near the boundary) without an explicit descendant-vs-non-descendant inequality. In \Cref{def:has}, we make the inequality explicit.

\myparagraph{Hyperbolic tree distortion and \texorpdfstring{$\delta$}{delta}-hyperbolicity}
A separate line of work formalizes hierarchy at the metric level: \citet{salaRepresentationTradeoffsHyperbolic2018} bound the distortion of the tree metric under a hyperbolic embedding, and \citet{adcockTreeLikeStructureLarge2013} characterize tree-likeness via Gromov's $\delta$-hyperbolicity. These are \emph{metric} (distance-based) formalizations. In contrast, \HAS is a \emph{directional} (angle-based) one and makes no claim about metric distortion.

\section{Limitations of our analysis}\label{app:model-limitations}
Our analysis isolates how hierarchical image-caption co-occurrence transfers existing image geometry to text. We discuss the scope of this mechanism and the assumptions that limit its application to CLIP training.

\myparagraph{Fixed image embeddings}
The fixed-image UFM captures the direction of transfer suggested by the observed modality asymmetry: hierarchical geometry is already approximately present in image embeddings, and text embeddings acquire it through contrastive learning. By optimizing text concept embeddings directly, we can identify hierarchical image-caption co-occurrence as the mechanism that aligns text embeddings with root-centered image prototypes. The model therefore explains how an alignment objective can transfer hierarchical structure without explicit supervision.
The limitation is that assuming fixed image embeddings does not capture joint training dynamics. We also replace CLIP's unit-norm embeddings with unnormalized embeddings and quadratic regularization for tractability (\Cref{rem:normalization}).

\myparagraph{Image embedding optimization analysis}
The image-side extension in \Cref{app:alternating} identifies noise reduction within caption groups after an exact text update followed by an exact image update under the linear loss. How that mechanism affects subtree-aggregated image prototypes during joint gradient descent, and how the initial image-side hierarchy arises, remain open.

\myparagraph{Approximate \HAS guarantees and empirical measurements}
The main theorems bound text-side violations using the image-side percentage of off-branch nodes whose worst-case triplet margin is at most the transfer threshold. The guarantees require small perturbations relative to the prototype norms. This sufficient condition fails for the real image prototypes examined in \Cref{app:per-triplet-empirical}, so the main theorems give no guarantee for the observed transfer there. Deriving tighter guarantees, therefore, remains an open problem.

\myparagraph{Sufficient regularization for InfoNCE}
The InfoNCE guarantee requires sufficiently large regularization to control the remainder in the expansion around uniform softmax weights. Experiments in \Cref{sec:exp-low-lambda} show HAS transfer at weaker regularization, but do not extend the theorem to that regime. A guarantee for weaker regularization, or for CLIP's normalized objective, remains open.

\section{Supporting Lemmas}\label{app:supporting-lemmas}\label{app:softmax-near-zero}

\subsection{Softmax expansions}

\begin{lemma}[Softmax partial derivatives]\label{lem:softmax-partial-derivatives}
For $K \ge 1$, define $\operatorname{softmax} : \R^K \to \R^K$ by
\begin{equation}
  \bigl(\operatorname{softmax}(\mathbf{s})\bigr)_j
  :=
  \frac{e^{s_j}}{\sum_{r=1}^K e^{s_r}},
  \qquad j \in [K].
\end{equation}
Then, for every $j,\ell \in [K]$,
\begin{equation}
  \frac{\partial}{\partial s_\ell}
  \bigl(\operatorname{softmax}(\mathbf{s})\bigr)_j
  =
  \bigl(\operatorname{softmax}(\mathbf{s})\bigr)_j
  \bigl(\delta_{j\ell} - \bigl(\operatorname{softmax}(\mathbf{s})\bigr)_\ell\bigr),
\end{equation}
where $\delta_{j\ell}$ is the Kronecker delta, equal to $1$ when $j=\ell$ and $0$ otherwise.
\end{lemma}

\begin{proof}
Write
\begin{equation}
  p_j(\mathbf{s})
  :=
  \bigl(\operatorname{softmax}(\mathbf{s})\bigr)_j
  =
  \frac{e^{s_j}}{Z(\mathbf{s})},
  \qquad
  Z(\mathbf{s})
  :=
  \sum_{r=1}^K e^{s_r}.
\end{equation}
Differentiating with respect to $s_\ell$ and using the quotient rule gives
\begin{align}
  \frac{\partial p_j}{\partial s_\ell}
  &=
  \frac{\bigl(\partial_{s_\ell} e^{s_j}\bigr) Z(\mathbf{s}) - e^{s_j}\, \partial_{s_\ell} Z(\mathbf{s})}{Z(\mathbf{s})^2}
  \notag\\
  &=
  \frac{\delta_{j\ell} e^{s_j} Z(\mathbf{s}) - e^{s_j} e^{s_\ell}}{Z(\mathbf{s})^2}
  \notag\\
  &=
  \frac{e^{s_j}}{Z(\mathbf{s})}
  \left(
    \delta_{j\ell} - \frac{e^{s_\ell}}{Z(\mathbf{s})}
  \right)
  \notag\\
  &=
  \operatorname{softmax}(\mathbf{s})_j \left(
    \delta_{j\ell} - \operatorname{softmax}(\mathbf{s})_\ell
  \right).
\end{align}
\end{proof}

\begin{lemma}[Quantitative softmax deviation from uniform]\label{lem:softmax-uniform-lipschitz}
For $K \ge 1$, define $\operatorname{softmax}:\R^K\to\R^K$ as in \Cref{lem:softmax-partial-derivatives}. Then, for every $\mathbf{s}\in\R^K$ and every $j\in[K]$,
\begin{equation}\label{eq:softmax-uniform-lipschitz}
  \left|
    \operatorname{softmax}(\mathbf{s})_j
    -
    \frac{1}{K}
  \right|
  \le
  2\|\mathbf{s}\|.
\end{equation}
\end{lemma}

\begin{proof}
Let $f_j(\mathbf{s}):=\operatorname{softmax}(\mathbf{s})_j$. By \Cref{lem:softmax-partial-derivatives},
\begin{equation}
  \nabla f_j(\mathbf{s})
  =
  f_j(\mathbf{s})\bigl(e_j-\operatorname{softmax}(\mathbf{s})\bigr),
\end{equation}
where $e_j$ is the $j$th standard basis vector in $\R^K$.
Since $0\le f_j(\mathbf{s})\le 1$ and $\|\operatorname{softmax}(\mathbf{s})\|\le 1$, we have $\|\nabla f_j(\mathbf{s})\|\le 2$ for every $\mathbf{s}\in\R^K$.
By the fundamental theorem of calculus along the segment $\{\tau\mathbf{s}:\tau\in[0,1]\}$,
\begin{equation}
  f_j(\mathbf{s})-f_j(\bzero)
  =
  \int_0^1 \nabla f_j(\tau\mathbf{s})^\top \mathbf{s}\,d\tau.
\end{equation}
Therefore
\begin{equation}
  \left|f_j(\mathbf{s})-f_j(\bzero)\right|
  \le
  \int_0^1 \left\|\nabla f_j(\tau\mathbf{s})\right\|\,\|\mathbf{s}\|\,d\tau
  \le
  2\|\mathbf{s}\|.
\end{equation}
Since $f_j(\bzero)=1/K$, this proves \Cref{eq:softmax-uniform-lipschitz}.
\end{proof}

\begin{lemma}[Softmax near the zero vector]\label{lem:softmax-near-zero}
For $K \ge 1$, define $\operatorname{softmax} : \R^K \to \R^K$ by
\begin{equation}
  \bigl(\operatorname{softmax}(\mathbf{s})\bigr)_j
  :=
  \frac{e^{s_j}}{\sum_{\ell=1}^K e^{s_\ell}},
  \qquad j \in [K].
\end{equation}
Let $\|\cdot\|$ denote the Euclidean norm on $\R^K$.
Then $\bigl(\operatorname{softmax}(\bzero)\bigr)_j = \frac{1}{K}$ for every $j \in [K]$. Moreover, if $\mathbf{s} \to \bzero$, then
\begin{equation}\label{eq:softmax-taylor-zero}
  \bigl(\operatorname{softmax}(\mathbf{s})\bigr)_j
  =
  \frac{1}{K}
  + \frac{1}{K}\left(s_j - \frac{1}{K}\sum_{\ell=1}^K s_\ell\right)
  + O(\|\mathbf{s}\|^2),
\end{equation}
uniformly over $j \in [K]$. In particular, if $\|\mathbf{s}\| = O(\varepsilon)$, then
\begin{equation}
  \bigl(\operatorname{softmax}(\mathbf{s})\bigr)_j
  =
  \frac{1}{K}
  + O(\varepsilon),
\end{equation}
uniformly over $j \in [K]$.
\end{lemma}

\begin{proof}
At $\mathbf{s}=\bzero$, all exponentials equal $1$, so
\begin{equation}
  \bigl(\operatorname{softmax}(\bzero)\bigr)_j
  =
  \frac{1}{\sum_{\ell=1}^K 1}
  =
  \frac{1}{K}.
\end{equation}
Since $\operatorname{softmax}$ is smooth on $\R^K$, Taylor's theorem around $\bzero$ with
$
  f_j(\mathbf{s})
  :=
  \bigl(\operatorname{softmax}(\mathbf{s})\bigr)_j
$
gives
\begin{equation}
  f_j(\mathbf{s})
  =
  f_j(\bzero)
  +
  \nabla f_j(\bzero)^\top \mathbf{s}
  + O(\|\mathbf{s}\|^2).
\end{equation}
Equivalently,
\begin{equation}
  \bigl(\operatorname{softmax}(\mathbf{s})\bigr)_j
  =
  \bigl(\operatorname{softmax}(\bzero)\bigr)_j
  +
  \sum_{\ell=1}^K
  \frac{\partial f_j}{\partial s_\ell}(\bzero)\, s_\ell
  + O(\|\mathbf{s}\|^2).
\end{equation}
By \Cref{lem:softmax-partial-derivatives},
\begin{equation}
  \frac{\partial}{\partial s_\ell}
  \bigl(\operatorname{softmax}(\mathbf{s})\bigr)_j
  =
  \bigl(\operatorname{softmax}(\mathbf{s})\bigr)_j
  \bigl(\delta_{j\ell} - \bigl(\operatorname{softmax}(\mathbf{s})\bigr)_\ell\bigr),
\end{equation}
so at $\mathbf{s}=\bzero$ we obtain
\begin{equation}
  \frac{\partial f_j}{\partial s_\ell}(\bzero)
  =
  \frac{1}{K}\left(\delta_{j\ell} - \frac{1}{K}\right).
\end{equation}
Therefore the linear term in the Taylor expansion is
\begin{align}
  \nabla f_j(\bzero)^\top \mathbf{s}
  &=
  \sum_{\ell=1}^K
  \frac{1}{K}\left(\delta_{j\ell} - \frac{1}{K}\right)s_\ell
  \notag\\
  &=
  \frac{1}{K}\left(
    s_j - \frac{1}{K}\sum_{\ell=1}^K s_\ell
  \right).
\end{align}
Moreover,
\begin{align}
  \left|\nabla f_j(\bzero)^\top \mathbf{s}\right|
  &\le
  \frac{1}{K}\left(
    |s_j|
    +
    \frac{1}{K}\left|\sum_{\ell=1}^K s_\ell\right|
  \right)
  \notag\\
  &\le
  \frac{1}{K}\left(
    \|\mathbf{s}\|
    +
    \frac{1}{K}\|\mathbf{1}\|\,\|\mathbf{s}\|
  \right)
  =
  \frac{1}{K}\left(
    1 + \frac{1}{\sqrt{K}}
  \right)\|\mathbf{s}\|,
\end{align}
where $\mathbf{1}\in\R^K$ is the all-ones vector. Thus the linear term is $O(\|\mathbf{s}\|)$. Substituting $f_j(\bzero)=\frac{1}{K}$ and this expression for $\nabla f_j(\bzero)^\top \mathbf{s}$ into the Taylor expansion yields \Cref{eq:softmax-taylor-zero}. If $\|\mathbf{s}\| = O(\varepsilon)$, then the linear term is $O(\varepsilon)$ and the remainder is $O(\varepsilon^2)$, which gives the final claim.
\end{proof}

\subsection{Prototype geometry and margin transfer}

\begin{lemma}[Balanced Descendant Prototype Averaging]\label{lem:prototype-centroid}
Assume the tree is balanced with branching factor $b$, and the prototypes $\{\bm{\mu}_k\}$ are constructed as in \Cref{app:step_by_step_construction}.
Then for every node $k$, the average of the leaf prototypes in the subtree rooted at $k$ is exactly $\bm{\mu}_k$.
\end{lemma}
\begin{proof}
The claim is immediate for leaves.
Now let $k$ be an internal node with children $v_1,\dots,v_b$.
By construction,
\begin{equation}
  \bm{\mu}_{v_j} = \bm{\mu}_k + \bU \bd_j,
  \qquad j=1,\dots,b,
\end{equation}
where the embedded simplex offsets satisfy $\sum_{j=1}^b \bd_j = \bzero$.
Hence the child average is
\begin{equation}
  \frac{1}{b}\sum_{j=1}^b \bm{\mu}_{v_j}
  =
  \bm{\mu}_k + \bU\left(\frac{1}{b}\sum_{j=1}^b \bd_j\right)
  =
  \bm{\mu}_k.
\end{equation}
Because the tree is balanced, each child subtree contains the same number of leaves.
Applying the induction hypothesis within each child subtree therefore shows that the leaf-descendant average of $k$ is the same average of the child averages, hence equal to $\bm{\mu}_k$.
\end{proof}

\begin{lemma}[Cosine stability under small perturbations]\label{lem:cosine-stability}
Let $\bx,\by\in\R^d$ be nonzero, let $\be_x,\be_y\in\R^d$, and set
\begin{equation}
  \rho := \min\{\|\bx\|,\|\by\|\}.
\end{equation}
If
\begin{equation}
  \max\{\|\be_x\|,\|\be_y\|\}\le \delta < \rho,
\end{equation}
then
\begin{equation}\label{eq:cosine-stability}
  \left|
    \cos(\bx+\be_x,\by+\be_y)
    -
    \cos(\bx,\by)
  \right|
  \le
  \frac{4\delta}{\rho}.
\end{equation}
\end{lemma}
\begin{proof}
For any nonzero $\bz\in\R^d$ and perturbation $\be$ with $\|\be\|<\|\bz\|$, define $\nu(\bz):=\bz/\|\bz\|$.
Then
\begin{align}\label{eq:cosine-stability-deviation}
  \|\nu(\bz+\be)-\nu(\bz)\|
  &=
  \left\|
    \frac{\bz+\be}{\|\bz+\be\|}
    -
    \frac{\bz+\be}{\|\bz\|}
    +
    \frac{\bz+\be}{\|\bz\|}
    -
    \frac{\bz}{\|\bz\|}
  \right\|
  \notag\\
  &\le
  \left\|
    \frac{\bz+\be}{\|\bz+\be\|}
    -
    \frac{\bz+\be}{\|\bz\|}
  \right\|
  +
  \left\|
    \frac{\bz+\be}{\|\bz\|}
    -
    \frac{\bz}{\|\bz\|}
  \right\|
  \notag\\
  &=
  \left|
    \frac{\|\bz\|-\|\bz+\be\|}{\|\bz\|}
  \right|
  +
  \frac{\|\be\|}{\|\bz\|}
  \notag\\
  &\le
  \frac{\|\be\|}{\|\bz\|}
  +
  \frac{\|\be\|}{\|\bz\|}
  \notag\\
  &=
  \frac{2\|\be\|}{\|\bz\|},
\end{align}
where the first inequality is the triangle inequality after adding and subtracting $(\bz+\be)/\|\bz\|$, and the second inequality is the reverse triangle inequality
\begin{equation}
  \left|\|\bz\|-\|\bz+\be\|\right|
  \le
  \|\be\|.
\end{equation}
Applying this bound with $(\bz,\be)=(\bx,\be_x)$ and $(\bz,\be)=(\by,\be_y)$ gives
\begin{equation}
  \|\nu(\bx+\be_x)-\nu(\bx)\|
  \le
  \frac{2\delta}{\rho},
  \qquad
  \|\nu(\by+\be_y)-\nu(\by)\|
  \le
  \frac{2\delta}{\rho}.
\end{equation}
Therefore
\begin{align}\label{eq:cosine-stability-final}
  &\left|
    \cos(\bx+\be_x,\by+\be_y)
    -
    \cos(\bx,\by)
  \right|
  =
  \left|
    \nu(\bx+\be_x)^\top\nu(\by+\be_y)
    -
    \nu(\bx)^\top\nu(\by)
  \right|
  \notag\\
  &=
  \left|
    \nu(\bx+\be_x)^\top\nu(\by+\be_y)
    -
    \nu(\bx)^\top\nu(\by+\be_y)
    +
    \nu(\bx)^\top\nu(\by+\be_y)
    -
    \nu(\bx)^\top\nu(\by)
  \right|
  \notag\\
  &\le
  \left|
    \left(\nu(\bx+\be_x)-\nu(\bx)\right)^\top\nu(\by+\be_y)
  \right|
  +
  \left|
    \nu(\bx)^\top\left(\nu(\by+\be_y)-\nu(\by)\right)
  \right|
  \notag\\
  &\le
  \|\nu(\bx+\be_x)-\nu(\bx)\|\,\|\nu(\by+\be_y)\|
  +
  \|\nu(\bx)\|\,\|\nu(\by+\be_y)-\nu(\by)\|
  \notag\\
  &=
  \|\nu(\bx+\be_x)-\nu(\bx)\|
  +
  \|\nu(\by+\be_y)-\nu(\by)\|
  \notag\\
  &\le
  \frac{4\delta}{\rho}.
\end{align}
where the first inequality is the triangle inequality and the second inequality is Cauchy--Schwarz.
\end{proof}

In \Cref{thm:linear-sampled-path-noisy-has,thm:large-lambda-has-noisy}, we compare two embedding families at a common anchor, so each quantity defined below is written with a superscript for the family $\bh=\{\bh_k\}_{k\in V\setminus\{\mathrm{root}\}}$ of nonzero vectors it is evaluated on. The first of the two is the quantity that must be positive in \Cref{def:has}.

\begin{definition}[Admissible triplet and triplet margin]\label{def:per-triplet-margin}
Write $\mathcal{O}_u:=V\setminus\bigl(\mathrm{Sub}(u)\cup\mathrm{Anc}(u)\bigr)$ for the off-branch set at $u$, which forms the denominator of \Cref{eq:violation-percentage}, and call $(u,v,w)$ an \emph{admissible triplet} if $u\in V\setminus\{\mathrm{root}\}$ is non-leaf, $v\in\mathrm{Desc}(u)$, and $w\in\mathcal{O}_u$. The \emph{triplet margin} of an admissible triplet under $\bh$ is
\begin{equation}\label{eq:per-triplet-margin}
  \gamma^{\bh}_{uvw} := \cos(\bh_u,\bh_v) - \cos(\bh_u,\bh_w).
\end{equation}
\end{definition}

When admissible triplets exist for a family of nonzero image prototypes, define their global margin by
\begin{equation}\label{eq:prototype-has-margin}
  \gamma_{\HAS}:=\min_{(u,v,w)\text{ admissible}}\gamma^{\bm{\mu}}_{uvw}.
\end{equation}
The second definition is \Cref{def:violation-percentage} with its descendant threshold lowered by a slack.

\begin{definition}[Slack violation percentage]\label{def:slack-violation-percentage}
Let $s^{\bh}_u:=\min_{v\in\mathrm{Desc}(u)}\cos(\bh_u,\bh_v)$ be the descendant threshold $s_u$ of \Cref{def:violation-percentage} evaluated on $\bh$. For a non-root non-leaf node $u$ and a slack $\tau\ge 0$, the \emph{slack-$\tau$ violation percentage} is
\begin{equation}\label{eq:slack-violation-percentage}
  \HASViol^{\bh}_{\tau}(u)
  := \frac{\left|\left\{w\in\mathcal{O}_u : \cos(\bh_u,\bh_w)\ge s^{\bh}_u-\tau\right\}\right|}{|\mathcal{O}_u|}\times 100\%,
\end{equation}
and we drop the subscript at $\tau=0$, writing $\HASViol^{\bh}(u):=\HASViol^{\bh}_{0}(u)$.
\end{definition}

At $\tau=0$ the threshold is unchanged, so $\HASViol^{\bh}(u)$ is exactly the violation percentage $\HASViol(u)$ of \Cref{def:violation-percentage} for the family $\bh$. In the main text the superscript is a mnemonic rather than a symbol: $\HASViol^{\mathrm{img}}_{\tau}(u)$ is $\HASViol^{\bm{\mu}}_{\tau}(u)$ on the image prototypes, and $\HASViol^{\mathrm{txt}}(u)$ is $\HASViol^{\bh}(u)$ with $\bh$ the optimal text embeddings under consideration.

The two definitions are connected through the identity $\min_{v\in\mathrm{Desc}(u)}\gamma^{\bh}_{uvw}=s^{\bh}_u-\cos(\bh_u,\bh_w)$, which holds for each fixed off-branch node $w$. The percentage $\HASViol^{\bh}_{\tau}(u)$ is therefore the percentage of off-branch nodes $w$ whose worst-case triplet margin at anchor $u$ is at most $\tau$.

\Cref{lem:per-triplet-margin-preservation} is the deterministic core shared by \Cref{thm:linear-sampled-path-noisy-has,thm:large-lambda-has-noisy}. It assumes nothing about the prototype variation model, the observed captions, or the global margin $\gamma_{\HAS}$, and all randomness in either theorem enters only through the bound $\eta$ on the perturbations.

\begin{lemma}[Per-triplet \HAS margin preservation]\label{lem:per-triplet-margin-preservation}
Let $\bm{\nu}=\{\bm{\nu}_k\}_{k\in V\setminus\{\mathrm{root}\}}$ be reference vectors with minimum norm $\rho_{\min}^{\nu}:=\min_{k\neq\mathrm{root}}\|\bm{\nu}_k\|>0$ and, for each non-root node $k$, let $\widetilde{\ba}_k=c_k(\bm{\nu}_k+\be_k)$ for some scaling $c_k>0$ and perturbation $\be_k\in\R^d$. Let $\eta\ge 0$ satisfy
\begin{equation}\label{eq:per-triplet-margin-preservation-eta}
  \max_{k\neq\mathrm{root}}\|\be_k\|\le\eta<\rho_{\min}^{\nu},
\end{equation}
and set $\tau:=8\eta/\rho_{\min}^{\nu}$. Then:
\begin{enumerate}[(i)]
  \item the family $\widetilde{\ba}:=\{\widetilde{\ba}_k\}_{k\neq\mathrm{root}}$ satisfies $\widetilde{\ba}_k\neq\bzero$ for every non-root node $k$, so the cosines in part (ii) and the violation percentage in part (iii) are well defined;
  \item every admissible triplet $(u,v,w)$ with $\gamma^{\bm{\nu}}_{uvw}>\tau$ satisfies $\cos(\widetilde{\ba}_u,\widetilde{\ba}_v)>\cos(\widetilde{\ba}_u,\widetilde{\ba}_w)$;
  \item $\HASViol^{\widetilde{\ba}}(u)\le\HASViol^{\bm{\nu}}_{\tau}(u)$ for every non-root non-leaf node $u$.
\end{enumerate}
\end{lemma}

\begin{proof}
(i) By \Cref{eq:per-triplet-margin-preservation-eta} and the reverse triangle inequality, $\|\bm{\nu}_k+\be_k\|\ge\rho_{\min}^{\nu}-\eta>0$ for every non-root node $k$; since $c_k>0$, this gives $\widetilde{\ba}_k\neq\bzero$. The $\bm{\nu}_k$ are nonzero by hypothesis, so every cosine in parts (ii) and (iii) is taken between nonzero vectors.

(ii) Fix an admissible triplet $(u,v,w)$ with $\gamma^{\bm{\nu}}_{uvw}>\tau$. Positive rescaling leaves cosines unchanged, so
\begin{equation}
  \cos(\widetilde{\ba}_u,\widetilde{\ba}_v)
  =
  \cos(\bm{\nu}_u+\be_u,\bm{\nu}_v+\be_v),
\end{equation}
and similarly with $v$ replaced by $w$. By \Cref{eq:per-triplet-margin-preservation-eta}, we can apply \Cref{lem:cosine-stability} to the pairs $(u,v)$ and $(u,w)$ with $\delta=\eta$ and pairwise minimum norm at least $\rho_{\min}^{\nu}$, so each of the two cosines differs from its reference counterpart by at most $4\eta/\rho_{\min}^{\nu}$. Therefore
\begin{equation}\label{eq:per-triplet-margin-preservation-chain}
  \cos(\widetilde{\ba}_u,\widetilde{\ba}_v)
  -
  \cos(\widetilde{\ba}_u,\widetilde{\ba}_w)
  \ge
  \cos(\bm{\nu}_u,\bm{\nu}_v)
  -
  \cos(\bm{\nu}_u,\bm{\nu}_w)
  -
  \frac{8\eta}{\rho_{\min}^{\nu}}
  =
  \gamma^{\bm{\nu}}_{uvw}-\tau
  >
  0.
\end{equation}

(iii) Fix a non-root non-leaf node $u$ and suppose $w$ is counted in $\HASViol^{\widetilde{\ba}}(u)$, i.e., $\cos(\widetilde{\ba}_u,\widetilde{\ba}_w)\ge s^{\widetilde{\ba}}_u$. Let $v^\star\in\mathrm{Desc}(u)$ attain the minimum defining $s^{\widetilde{\ba}}_u$. Then the \HAS inequality fails for the admissible triplet $(u,v^\star,w)$, so the contrapositive of part (ii) gives $\gamma^{\bm{\nu}}_{uv^\star w}\le\tau$. Therefore
\begin{equation}
  s^{\bm{\nu}}_u-\cos(\bm{\nu}_u,\bm{\nu}_w)
  =
  \min_{v\in\mathrm{Desc}(u)}\gamma^{\bm{\nu}}_{uvw}
  \le
  \gamma^{\bm{\nu}}_{uv^\star w}
  \le
  \tau
  \implies
  \cos(\bm{\nu}_u,\bm{\nu}_w)\ge s^{\bm{\nu}}_u-\tau.
\end{equation}
Thus, the same $w$ is counted in $\HASViol^{\bm{\nu}}_{\tau}(u)$. Both percentages share the denominator $|\mathcal{O}_u|$, so the inclusion of the counted sets yields the claim.
\end{proof}

Applied with $\bm{\nu}=\bm{\mu}$, part (iii) is the transfer inequality of \Cref{eq:per-triplet-has-bound}. In \Cref{app:shared-transfer}, we derive the perturbation bound $\eta$ for each loss.

\subsection{Caption bias and sampling}

The optimizer forms for both losses involve the same difference of empirical means.
In \Cref{lem:caption-mean-decomposition}, we separate that difference into prototype, caption-bias, and variation components, and in \Cref{cor:zero-caption-bias}, we give the conditions under which the caption-bias component vanishes.

\begin{definition}[Empirical caption prototype and caption bias]\label{def:caption-bias}
Fix prototypes $\{\bm{\mu}_k\}_{k\in V}\subset\R^d$, leaf labels $y_j\in V_{\mathrm{leaf}}$, and caption labels $c_j\in V$ for every $j\in[N]$, with $N_k>0$ for every non-root node $k$. The captions need not be path-consistent. Recall from \Cref{sec:has-setup} that $\cP_k:=\{j\in[N]:c_j=k\}$ is the positive set of node $k$ and $N_k:=|\cP_k|$. The \emph{empirical caption prototype} of node $k$ and the \emph{global prototype mean} are
\begin{equation}\label{eq:empirical-caption-prototype}
  \bar{\bm{\mu}}_k := \frac{1}{N_k}\sum_{j\in\cP_k}\bm{\mu}_{y_j},
  \qquad
  \bar{\bm{\mu}} := \frac{1}{N}\sum_{j=1}^{N}\bm{\mu}_{y_j},
\end{equation}
with associated \emph{caption biases} $\bm{b}_k:=\bar{\bm{\mu}}_k-\bm{\mu}_k$ and $\bm{b}:=\bar{\bm{\mu}}-\bm{\mu}_{\mathrm{root}}$.
\end{definition}

\begin{lemma}[Empirical-mean decomposition]\label{lem:caption-mean-decomposition}
Fix prototypes, leaf labels, and caption labels as in \Cref{def:caption-bias}, and let $\xi_k\in\R$ for every non-root node $k$.
Let $\bz_j=\bm{\mu}_{y_j}+\bm{\varepsilon}_j$ with arbitrary $\bm{\varepsilon}_j\in\R^d$ for every $j\in[N]$, and define $\bar{\bm{\varepsilon}}_k:=N_k^{-1}\sum_{j\in\cP_k}\bm{\varepsilon}_j$ for every non-root node $k$ and $\bar{\bm{\varepsilon}}:=N^{-1}\sum_{j=1}^N\bm{\varepsilon}_j$.
Then, for every non-root node $k$,
\begin{equation}\label{eq:caption-bias-empirical-means}
  \bar{\bz}_k=\bm{\mu}_k+\bm{b}_k+\bar{\bm{\varepsilon}}_k,
  \qquad
  \bar{\bz}=\bm{\mu}_{\mathrm{root}}+\bm{b}+\bar{\bm{\varepsilon}},
\end{equation}
and, for every non-root node $k$,
\begin{equation}\label{eq:caption-mean-decomposition}
  \begin{aligned}
    \bar{\bz}_k-\xi_k\bar{\bz}
    ={}&\bm{\mu}_k-\xi_k\bm{\mu}_{\mathrm{root}}
    +\bm{b}_k-\xi_k\bm{b}\\
    &+\bar{\bm{\varepsilon}}_k-\xi_k\bar{\bm{\varepsilon}}.
  \end{aligned}
\end{equation}
\end{lemma}

\begin{proof}
Averaging $\bz_j=\bm{\mu}_{y_j}+\bm{\varepsilon}_j$ over $\cP_k$ and over $[N]$ gives \Cref{eq:caption-bias-empirical-means} by \Cref{def:caption-bias}.
Subtracting $\xi_k\bar{\bz}$ gives \Cref{eq:caption-mean-decomposition}.
\end{proof}

\Cref{lem:caption-mean-decomposition} is an identity, so it holds for any observed captions; \Cref{app:shared-transfer} uses it in that generality and adds the caption bias to the error term. The main theorems instead work in the symmetric setting of \Cref{sec:has-setup}, where the bias vanishes.

\begin{corollary}[Zero caption bias under symmetric captions]\label{cor:zero-caption-bias}
Fix prototypes, leaf labels, and caption labels as in \Cref{def:caption-bias}, suppose the captions are path-consistent, and suppose \Cref{ass:leaf-uniform-caption} (leaf-uniform caption coverage) and \Cref{ass:prototype-centroid-consistency} (prototype consistency) hold.
Then $\bar{\bm{\mu}}_k=\bm{\mu}_k$, and hence $\bm{b}_k=\bzero$, for every non-root node $k$.
If exactly $n\in\mathbb{N}_{>0}$ images are sampled from each leaf, then also $\bar{\bm{\mu}}=\bm{\mu}_{\mathrm{root}}$ and $\bm{b}=\bzero$.
\end{corollary}

\begin{proof}
Path-consistency gives $y_j\in\mathrm{Sub}(k)$ for every $j\in\cP_k$, and by \Cref{ass:leaf-uniform-caption} each such leaf contributes equally many images.
Hence $\bar{\bm{\mu}}_k$ is the uniform average of the leaf prototypes in $\mathrm{Sub}(k)$, which equals $\bm{\mu}_k$ by \Cref{ass:prototype-centroid-consistency}, so $\bm{b}_k=\bzero$.
With exactly $n$ images per leaf, $\bar{\bm{\mu}}$ is the uniform average of all leaf prototypes, which equals $\bm{\mu}_{\mathrm{root}}$ by the same assumption at the root, so $\bm{b}=\bzero$.
\end{proof}

\subsection{Concentration}

\begin{lemma}[Concentration of finitely many empirical variation means]\label{lem:subgaussian-empirical-means}
Let $\cK$ be a finite index set.
Let $\{\bm{\varepsilon}_j\}_{j=1}^{N}$ be independent centered random vectors in $\R^d$.
Assume there exists $\sigma>0$ such that, for every $j\in[N]$ and every unit vector $\bu\in\mathbb{S}^{d-1}$, the scalar $\bu^\top \bm{\varepsilon}_j$ is sub-Gaussian with Orlicz norm at most $\sigma$.
For each $k\in\cK$, let $I_k\subseteq[N]$ with $s_k:=|I_k|>0$, and define
\begin{equation}
  \bar{\bm{\varepsilon}}_k
  :=
  \frac{1}{s_k}\sum_{j\in I_k}\bm{\varepsilon}_j,
  \qquad
  \bar{\bm{\varepsilon}}
  :=
  \frac{1}{N}\sum_{j=1}^{N}\bm{\varepsilon}_j.
\end{equation}
For each $k\in\cK$, let $\xi_k\in\R$, and define
\begin{equation}
  \xi_{\max}
  :=
  \max_{k\in\cK}|\xi_k|,
  \qquad
  s_{\min}
  :=
  \min_{k\in\cK}s_k.
\end{equation}
Then there exists a universal constant $C>0$ such that, for every $u\ge 0$, with probability at least $1-4e^{-u}$,
\begin{equation}\label{eq:finite-sample-noise-centroid-bound}
  \max_{k\in\cK}
  \left\|
    \bar{\bm{\varepsilon}}_k
    -
    \xi_k\bar{\bm{\varepsilon}}
  \right\|
  \le
  C\sigma\left(
    \sqrt{\frac{d+\log|\cK|+u}{s_{\min}}}
    +
    \xi_{\max}\sqrt{\frac{d+u}{N}}
  \right).
\end{equation}
\end{lemma}
\begin{proof}
For each fixed $k\in\cK$, apply the scalar Hoeffding-type tail bound for averages of independent centered sub-Gaussian random variables to $\bu^\top\bar{\bm{\varepsilon}}_k$ for $\bu\in\mathbb{S}^{d-1}$.
A standard $1/2$-net argument on $\mathbb{S}^{d-1}$ then gives a universal constant $C>0$ such that, for every $t\ge 0$,
\begin{equation}
  \Pr\left(
    \left\|\bar{\bm{\varepsilon}}_k\right\|
    >
    C\sigma\sqrt{\frac{d+t}{s_k}}
  \right)
  \le
  2e^{-t}
\end{equation}
The same argument with $I_k=[N]$ and $s_k=N$ gives
\begin{equation}
  \Pr\left(
    \left\|\bar{\bm{\varepsilon}}\right\|
    >
    C\sigma\sqrt{\frac{d+t}{N}}
  \right)
  \le
  2e^{-t}.
\end{equation}
Apply the first bound with $t=u+\log|\cK|$ and take a union bound over $k\in\cK$.
Then, with probability at least $1-2e^{-u}$,
\begin{equation}\label{eq:subgaussian-empirical-means-max-bound}
  \max_{k\in\cK}\left\|\bar{\bm{\varepsilon}}_k\right\|
  \le
  C\sigma\sqrt{\frac{d+\log|\cK|+u}{s_{\min}}}.
\end{equation}
The second bound with $t=u$ gives, with probability at least $1-2e^{-u}$,
\begin{equation}\label{eq:subgaussian-global-mean-bound}
  \left\|\bar{\bm{\varepsilon}}\right\|
  \le
  C\sigma\sqrt{\frac{d+u}{N}}.
\end{equation}
Another union bound shows that both displayed bounds hold simultaneously with probability at least $1-4e^{-u}$.
On that event,
\begin{equation}
  \max_{k\in\cK}
  \left\|
    \bar{\bm{\varepsilon}}_k
    -
    \xi_k\bar{\bm{\varepsilon}}
  \right\|
  \le
  \max_{k\in\cK}\left\|\bar{\bm{\varepsilon}}_k\right\|
  +
  \xi_{\max}\left\|\bar{\bm{\varepsilon}}\right\|,
\end{equation}
where we used the triangle inequality and the bound $|\xi_k|\le \xi_{\max}$ for every $k\in\cK$.
Substituting \Cref{eq:subgaussian-empirical-means-max-bound,eq:subgaussian-global-mean-bound} into this estimate gives \Cref{eq:finite-sample-noise-centroid-bound}.
\end{proof}

\begin{lemma}[Maximum feature norm under sub-Gaussian variation]\label{lem:subgaussian-feature-max-norm}
Suppose \Cref{ass:noisy-prototype} holds, and define $\mu_{\mathrm{leaf},\max}:=\max_{\ell\in V_{\mathrm{leaf}}}\|\bm{\mu}_\ell\|$.
Then there exists a universal constant $C>0$ such that, for every $\epsilon\in(0,1)$, with probability at least $1-\epsilon$,
\begin{equation}\label{eq:subgaussian-feature-max-norm}
  \max_{j\in[N]}\|\bz_j\|
  \le
  \mu_{\mathrm{leaf},\max}
  +
  C\sigma\sqrt{d+\log(2N/\epsilon)}.
\end{equation}
\end{lemma}
\begin{proof}
For each fixed $j\in[N]$, the same Hoeffding bound used in the proof of \Cref{lem:subgaussian-empirical-means} gives $\Pr(\|\bm{\varepsilon}_j\|>C\sigma\sqrt{d+u})\le 2e^{-u}$ for a universal constant $C>0$.
Taking $u=\log(2N/\epsilon)$ and union bounding over $j\in[N]$ gives $\max_{j\in[N]}\|\bm{\varepsilon}_j\|\le C\sigma\sqrt{d+\log(2N/\epsilon)}$ with probability at least $1-\epsilon$.
Absorbing the factor $2$ into the universal constant and using $\|\bz_j\|\le\|\bm{\mu}_{y_j}\|+\|\bm{\varepsilon}_j\|\le\mu_{\mathrm{leaf},\max}+\|\bm{\varepsilon}_j\|$ proves the claim.
\end{proof}

\section{Proofs for Intermediate Results in Linear Contrastive Loss Analysis}\label{app:proofs-linear-sampled-path}

\subsection{Closed-Form Minimizers for the Linear Contrastive Loss}
The same label encoding governs the image-side update in \Cref{app:alternating}, so we state both minimizers together.
\begin{proposition}[Closed-Form Minimizers for the Linear Contrastive Loss with One Side Fixed]\label{prop:linear-sampled-path-minimizer}
Define the assignment matrix $\mathbf{Y}\in\{0,1\}^{M\times N}$ via $Y_{kj}:=\mathbf{1}[c_j=k]$.
Then define the label encoding matrix $\mathbf{B}\in\R^{M\times N}$ by
\begin{equation}\label{eq:linear-sampled-path-matrix}
  B_{k j}
  :=
  \frac{1}{2}\left(
    Y_{k j}
    -
    \frac{1-Y_{k j}}{M-1}
  \right)
  +
  \frac{1}{2}\left(
    Y_{k j}
    -
    \frac{N_k-Y_{k j}}{N-1}
  \right).
\end{equation}
Then the unique minimizer of \Cref{eq:objective-linear-sampled-path} is
\begin{equation}\label{eq:linear-sampled-path-minimizer}
  \bA^{\mathrm{lin},*}
  =
  \frac{1}{\lambda}\bZ\mathbf{B}^\top.
\end{equation}
For any fixed $\bA\in\R^{d\times M}$ and $\lambda_Z>0$, the image-side quadratic $-\mathrm{tr}(\bA^\top\bZ\mathbf{B}^\top)+\frac{\lambda_Z}{2}\|\bZ\|_F^2$ has unique minimizer $\lambda_Z^{-1}\bA\mathbf{B}$ over $\bZ\in\R^{d\times N}$.
\end{proposition}
We first rewrite \Cref{eq:objective-linear-sampled-path} as
\begin{equation}\label{eq:objective-linear-sampled-path-trace}
  \cL^{\mathrm{lin}}(\bA)
  =
  -\mathrm{tr}\!\left(\bA^\top \bZ\mathbf{B}^\top\right)
  +
  \frac{\lambda}{2}\|\bA\|_F^2,
\end{equation}
with $\mathbf{B}$ as in \Cref{eq:linear-sampled-path-matrix}.
\begin{proof}
Write the two terms in \Cref{eq:objective-linear-sampled-path} as $\cL_{\mathrm{i2t}}^{\mathrm{lin}}(\bA)$ and $\cL_{\mathrm{t2i}}^{\mathrm{lin}}(\bA)$.
Since $Y_{k j}=\mathbf{1}[c_j=k]$, the image-to-text term satisfies
\begin{align}
  \cL_{\mathrm{i2t}}^{\mathrm{lin}}(\bA)
  &=
  -\frac{1}{2}\sum_{j=1}^{N}\ba_{c_j}^\top \bz_j
  +
  \frac{1}{2(M-1)}\sum_{j=1}^{N}\sum_{\substack{k\in[M]\\ k\neq c_j}}\ba_k^\top \bz_j
  \notag\\
  &=
  -\sum_{k=1}^{M}\sum_{j=1}^{N}
  \frac{1}{2}\left(
    Y_{k j}
    -
    \frac{1-Y_{k j}}{M-1}
  \right)\ba_k^\top \bz_j.
\end{align}
For the text-to-image term, fix $k\in[M]$ and $j\in[N]$.
The logit $\ba_k^\top \bz_j$ appears once in the positive part when $c_j=k$, and it appears in the negative pool once for each $j'\in[N]\setminus\{j\}$ with $c_{j'}=k$.
Therefore the coefficient of $\ba_k^\top \bz_j$ in $\cL_{\mathrm{t2i}}^{\mathrm{lin}}(\bA)$ is
\begin{equation}
  -\frac{1}{2}Y_{k j}
  +
  \frac{1}{2(N-1)}\sum_{\substack{j'\in[N]\\ j'\neq j}}Y_{k j'}
  =
  -\frac{1}{2}\left(
    Y_{k j}
    -
    \frac{N_k-Y_{k j}}{N-1}
  \right),
\end{equation}
because $\sum_{\substack{j'\in[N]\\ j'\neq j}}Y_{k j'}=N_k-Y_{k j}$ by the definition of $N_k$.
Hence
\begin{align}
  \cL_{\mathrm{t2i}}^{\mathrm{lin}}(\bA)
  &=
  -\sum_{k=1}^{M}\sum_{j=1}^{N}
  \frac{1}{2}\left(
    Y_{k j}
    -
    \frac{N_k-Y_{k j}}{N-1}
  \right)\ba_k^\top \bz_j.
\end{align}
Adding the two terms and using \Cref{eq:linear-sampled-path-matrix} gives
\begin{equation}
  \cL^{\mathrm{lin}}(\bA)
  =
  -\sum_{k=1}^{M}\sum_{j=1}^{N} B_{k j}\ba_k^\top \bz_j
  +
  \frac{\lambda}{2}\|\bA\|_F^2.
\end{equation}
Finally,
\begin{equation}
  \mathrm{tr}\!\left(\bA^\top \bZ\mathbf{B}^\top\right)
  =
  \sum_{k=1}^{M}\ba_k^\top\left(\sum_{j=1}^{N}B_{k j}\bz_j\right)
  =
  \sum_{k=1}^{M}\sum_{j=1}^{N} B_{k j}\ba_k^\top \bz_j,
\end{equation}
so the previous display is exactly \Cref{eq:objective-linear-sampled-path-trace}.
Differentiating with respect to $\bA$ yields
\begin{equation}
  \nabla_{\bA}\cL^{\mathrm{lin}}(\bA)
  =
  \lambda \bA - \bZ\mathbf{B}^\top.
\end{equation}
Setting the gradient to zero gives \Cref{eq:linear-sampled-path-minimizer}, and the quadratic regularizer makes the minimizer unique.
For the image-side quadratic, $\nabla_{\bZ}=\lambda_Z\bZ-\bA\mathbf{B}$; setting this gradient to zero gives the stated image-side minimizer, unique because $\lambda_Z>0$.
\end{proof}

\subsection{Proof of \Cref{cor:linear-sampled-path-centroid-form} (Empirical-Mean Expression)}

\begin{corollary}[Empirical-Mean Expression for the Linear Minimizer]\label{cor:linear-sampled-path-centroid-form}
Every node $k\in[M]$ with $N_k>0$ satisfies
\begin{equation}\label{eq:linear-sampled-path-centroid-form}
  \ba_k^{\mathrm{lin},*}
  =
  \frac{N_k}{\lambda}\left(
    1
    +
    \frac{1}{2(M-1)}
    +
    \frac{1}{2(N-1)}
  \right)\bar{\bz}_k
  -
  \frac{N}{\lambda}\left(
    \frac{1}{2(M-1)}
    +
    \frac{N_k}{2(N-1)}
  \right)\bar{\bz}.
\end{equation}
\end{corollary}

\begin{proof}
Fix a node $k\in[M]$ with $N_k>0$, and write
\begin{equation}
  \cP_k := \{j\in[N] : c_j = k\}.
\end{equation}
By \Cref{eq:linear-sampled-path-matrix},
\begin{equation}
  B_{k j}
  =
  \begin{cases}
    \gamma_k^+, & j\in\cP_k, \\[2pt]
    -\gamma_k^-, & j\notin\cP_k,
  \end{cases}
\end{equation}
where
\begin{equation}
  \gamma_k^+
  :=
  1 - \frac{N_k-1}{2(N-1)},
  \qquad
  \gamma_k^-
  :=
  \frac{1}{2(M-1)} + \frac{N_k}{2(N-1)}.
\end{equation}
Applying \Cref{prop:linear-sampled-path-minimizer},
\begin{align}
  \ba_k^{\mathrm{lin},*}
  &=
  \frac{1}{\lambda}\sum_{j=1}^{N} B_{k j}\bz_j
  \notag\\
  &=
  \frac{\gamma_k^+}{\lambda}\sum_{j\in\cP_k}\bz_j
  -
  \frac{\gamma_k^-}{\lambda}\sum_{j\notin\cP_k}\bz_j.
\end{align}
By definition,
\begin{equation}
  \sum_{j\notin\cP_k}\bz_j
  =
  \sum_{j=1}^{N}\bz_j - \sum_{j\in\cP_k}\bz_j,
\end{equation}
so
\begin{equation}
  \ba_k^{\mathrm{lin},*}
  =
  \frac{\gamma_k^++\gamma_k^-}{\lambda}\sum_{j\in\cP_k}\bz_j
  -
  \frac{\gamma_k^-}{\lambda}\sum_{j=1}^{N}\bz_j.
\end{equation}
By definition of $\bar{\bz}_k$,
\begin{equation}
  \sum_{j\in\cP_k}\bz_j
  =
  N_k\bar{\bz}_k.
\end{equation}
\begin{equation}
  \sum_{j=1}^{N}\bz_j
  =
  N\bar{\bz}.
\end{equation}
Substituting these identities and using
\begin{equation}
  \gamma_k^++\gamma_k^-
  =
  1 + \frac{1}{2(M-1)} + \frac{1}{2(N-1)}
\end{equation}
yields
\begin{equation}
  \ba_k^{\mathrm{lin},*}
  =
  \frac{N_k}{\lambda}\left(
    1
    +
    \frac{1}{2(M-1)}
    +
    \frac{1}{2(N-1)}
  \right)\bar{\bz}_k
  -
  \frac{N}{\lambda}\left(
    \frac{1}{2(M-1)}
    +
    \frac{N_k}{2(N-1)}
  \right)\bar{\bz},
\end{equation}
which is \Cref{eq:linear-sampled-path-centroid-form}.
\end{proof}

\subsection{Proof of \Cref{cor:linear-sampled-path-noisy-centroid-form}}
\LinearSampledPathNoisyCentroidFormCor*
\begin{proof}
The observed captions are path-consistent by the setup in \Cref{sec:has-setup}, and the same setup has exactly $n$ images per leaf.
Under these hypotheses, \Cref{ass:leaf-uniform-caption,ass:prototype-centroid-consistency}, and the decomposition in \Cref{ass:noisy-prototype}, we can apply \Cref{lem:caption-mean-decomposition} and \Cref{cor:zero-caption-bias}.
By \Cref{ass:leaf-uniform-caption}, we also have $N_k>0$ for every non-root node $k$, so we can apply \Cref{cor:linear-sampled-path-centroid-form}.
Factor its empirical-mean expression using $\xi_k^{\mathrm{lin}}$ from \Cref{cor:linear-sampled-path-noisy-centroid-form}, and substitute \Cref{eq:caption-mean-decomposition} with $\bm{b}_k=\bm{b}=\bzero$ to obtain \Cref{eq:linear-sampled-path-noisy-centroid-form}.
\end{proof}

\section{Proofs of Intermediate Results in InfoNCE Loss Analysis}\label{app:proofs-has-infonce-nonthm}

\subsection{Proof of \Cref{lem:strong-convexity} (Strong Convexity)}\label{app:proof-strong-convexity}

\begin{lemma}[Strong Convexity]\label{lem:strong-convexity}
The objective $\cL(\bA)$ in \Cref{eq:objective-simplified-finite} is $\lambda$-strongly convex in $\bA$.
Therefore, it has a unique global minimizer $\bA^{\mathrm{nce},*}$.
\end{lemma}

\begin{proof}
First, each logit $\ba_k^\top \bz_j$ is linear in $\bA$.
Then, the term $-(\text{linear in } \bA) + \log\!\sum \exp(\text{linear in } \bA)$ is convex as the composition of log-sum-exp with linear maps.
Finally, composition with the sum over $j\in[N]$ and the regularizer $\frac{\lambda}{2}\|\bA\|_F^2$ yield $\lambda$-strong convexity.
\end{proof}

\subsection{Proof of \Cref{prop:stationarity} (Stationarity Condition)}\label{app:stationarity-proof}\label{app:stationarity-finite-proof}

\begin{proposition}[Stationarity Condition]\label{prop:stationarity}\label{prop:stationarity-finite}
Define the softmax weights $q_k^{(j)}(\bA) := e^{\ba_k^\top \bz_j} / \sum_{k'=1}^M e^{\ba_{k'}^\top \bz_j}$ and $r_{j'}^{(j)}(\bA) := e^{\ba_{c_j}^\top \bz_{j'}} / \sum_{m=1}^N e^{\ba_{c_j}^\top \bz_m}$.
The unique minimizer $\bA^{\mathrm{nce},*}$ of \Cref{eq:objective-simplified-finite} satisfies, for each $k \in [M]$,
\begin{equation}\label{eq:kkt-finite}
  \ba_k^{\mathrm{nce},*} = \frac{1}{\lambda} \left(
    \sum_{j:\, c_j = k} \bz_j
  - \frac{1}{2}\sum_{j=1}^N q_k^{(j)}(\bA^{\mathrm{nce},*}) \bz_j
  - \frac{1}{2}\sum_{j:\, c_j = k}\sum_{j'=1}^N r_{j'}^{(j)}(\bA^{\mathrm{nce},*}) \bz_{j'}.
  \right)
\end{equation}
\end{proposition}

\begin{proof}
First, we compute the image-to-text gradients.
The total i2t loss is
\begin{equation}
  \cL^{\text{i2t}} = \sum_{j=1}^N \cL^{\text{i2t}}_{j},
\end{equation}
where
\begin{equation}
  \cL^{\text{i2t}}_{j}
  =
  -\ba_{c_j}^\top \bz_j + \log\sum_{k'} e^{\ba_{k'}^\top \bz_j}.
\end{equation}
Each term depends on $\ba_k$ both through the numerator (when $c_j = k$) and the log-sum-exp denominator:
\begin{align}
  \nabla_{\ba_k} \cL^{\text{i2t}}_{j}
  &= -\ind[c_j = k]\,\bz_j
  + \frac{e^{\ba_k^\top \bz_j}}{\sum_{k'} e^{\ba_{k'}^\top \bz_j}}\,\bz_j
  \notag \\
  &= \bigl(-\ind[c_j = k] + q_k^{(j)}\bigr)\,\bz_j.
\end{align}
Summing over $j\in[N]$ gives
\begin{equation}\label{eq:grad-i2t}
  \nabla_{\ba_k} \cL^{\text{i2t}}
  = \sum_{j=1}^N \bigl(-\ind[c_j = k] + q_k^{(j)}\bigr)\,\bz_j
  = -\!\!\sum_{j:\,c_j=k}\!\bz_j
    + \sum_{j=1}^N q_k^{(j)}\,\bz_j.
\end{equation}

Next, we compute the text-to-image gradients.
The total t2i loss is
\begin{equation}
  \cL^{\text{t2i}} = \sum_{j=1}^N \cL^{\text{t2i}}_{j},
\end{equation}
where
\begin{equation}
  \cL^{\text{t2i}}_{j}
  =
  -\ba_{c_j}^\top \bz_j
  + \log\sum_{j'=1}^N e^{\ba_{c_j}^\top \bz_{j'}}.
\end{equation}
Since $\ba_k$ appears in $\cL^{\text{t2i}}_{j}$ only when $c_j = k$:
\begin{align}
  \nabla_{\ba_k} \cL^{\text{t2i}}_{j}
  &= \ind[c_j = k]\left(
      -\bz_j
      + \frac{\sum_{j'=1}^N e^{\ba_{c_j}^\top \bz_{j'}}\,\bz_{j'}}
             {\sum_{m=1}^N e^{\ba_{c_j}^\top \bz_m}}
    \right)
  \notag \\
  &= \ind[c_j = k]\left(
      -\bz_j
      + \sum_{j'=1}^N r_{j'}^{(j)}\,\bz_{j'}
    \right).
\end{align}
Summing over $j\in[N]$ gives
\begin{equation}\label{eq:grad-t2i}
  \nabla_{\ba_k} \cL^{\text{t2i}}
  =
  \sum_{j:\,c_j=k}\!\left(
    -\bz_j
      + \sum_{j'=1}^N r_{j'}^{(j)}\,\bz_{j'}
  \right).
\end{equation}

Finally, we combine \Cref{eq:grad-i2t,eq:grad-t2i} and add the regularization gradient $\lambda\ba_k$.
Since $\bA^{\mathrm{nce},*}$ is the unique minimizer, it satisfies
\begin{equation}
  \nabla_{\ba_k}\cL
  =
  \frac{1}{2}\left(
    \nabla_{\ba_k} \cL^{\text{i2t}}
    + \nabla_{\ba_k} \cL^{\text{t2i}}
  \right)
  + \lambda \ba_k^{\mathrm{nce},*}
  =
  \bzero.
\end{equation}
Substituting \Cref{eq:grad-i2t,eq:grad-t2i} and rearranging yields \Cref{eq:kkt-finite}.
\end{proof}

\subsection{Proof of \Cref{prop:large-lambda} (Large-Regularization Expansion)}\label{app:proof-large-lambda}
\LargeLambdaExpansionProp*
\begin{proof}
Let $Z_{\max} := \max_{j\in[N]}\|\bz_j\|$.
Since $\sum_{k=1}^M q_k^{(j)}(\bA^{\mathrm{nce},*}) = 1$ and $\sum_{j'=1}^N r_{j'}^{(j)}(\bA^{\mathrm{nce},*}) = 1$, we obtain from \Cref{eq:kkt-finite}
\begin{align}
  \|\ba_k^{\mathrm{nce},*}\|
  &\le
  \frac{1}{\lambda}\left(
    N_k Z_{\max}
    + \frac{1}{2}\sum_{j=1}^N q_k^{(j)}(\bA^{\mathrm{nce},*}) Z_{\max}
    + \frac{1}{2}\sum_{j\in\cP_k}\sum_{j'=1}^N r_{j'}^{(j)}(\bA^{\mathrm{nce},*}) Z_{\max}
  \right)
  \notag\\
  &\le
  \frac{1}{\lambda}\left(N_k + \frac{N}{2} + \frac{N_k}{2}\right) Z_{\max}
  =
  O(\lambda^{-1}).
\end{align}
Therefore every logit $(\ba_\ell^{\mathrm{nce},*})^\top\bz_j$ is $O(\lambda^{-1})$, uniformly over $(\ell,j)\in[M]\times[N]$.
Hence the i2t logit vectors
\[
  \mathbf{s}^{(j)}
  :=
  \bigl((\ba_1^{\mathrm{nce},*})^\top\bz_j,\dots,(\ba_M^{\mathrm{nce},*})^\top\bz_j\bigr)
  \in \R^M
\]
and the t2i logit vectors
\[
  \mathbf{u}^{(j)}
  :=
  \bigl((\ba_{c_j}^{\mathrm{nce},*})^\top\bz_{j'}\bigr)_{j'\in[N]}
  \in \R^N
\]
have all entries' magnitudes of order $O(\lambda^{-1})$, and therefore satisfy $\|\mathbf{s}^{(j)}\|, \|\mathbf{u}^{(j)}\| = O(\lambda^{-1})$, uniformly over the relevant indices. In particular, these norms converge to $0$ as $\lambda\to\infty$, so for all sufficiently large $\lambda$ they lie in a fixed neighborhood of $\bzero$ where we can apply \Cref{lem:softmax-near-zero}. Applying that lemma with $K=M$ and $K=N$, respectively, gives
\begin{equation}
  q_k^{(j)}(\bA^{\mathrm{nce},*}) = \frac{1}{M} + O(\lambda^{-1}),
  \qquad
  r_{j'}^{(j)}(\bA^{\mathrm{nce},*}) = \frac{1}{N} + O(\lambda^{-1}),
\end{equation}
uniformly over the relevant indices.
Substituting these expansions into \Cref{eq:kkt-finite} gives
\begin{align}
  \ba_k^{\mathrm{nce},*}
  &=
  \frac{1}{\lambda}\left(
    \sum_{j\in\cP_k}\bz_j
    - \frac{1}{2M}\sum_{j=1}^N \bz_j
    - \frac{N_k}{2N}\sum_{j=1}^N \bz_j
  \right)
  + O(\lambda^{-2})
  \notag\\
  &=
  \frac{N_k}{\lambda}\bar{\bz}_k
  -
  \frac{1}{\lambda}\left(\frac{N}{2M} + \frac{N_k}{2}\right)\bar{\bz}
  + O(\lambda^{-2}),
\end{align}
which is \Cref{eq:large-lambda}.
\end{proof}

\subsection{Proof of \Cref{cor:prototype-form} (Prototype Decomposition)}\label{app:prototype-form}\label{app:prototype-form-noisy}
\begin{proof}
Recall the per-node global-mean coefficient
\begin{equation}\label{eq:large-lambda-xi-def}
  \xi_k^{\mathrm{nce}} := \frac{\frac{N}{2M}+\frac{N_k}{2}}{N_k},
  \qquad k\in V\setminus\{\mathrm{root}\}.
\end{equation}
The observed captions are path-consistent by the setup in \Cref{sec:has-setup}, and the same setup has exactly $n$ images per leaf.
Under these hypotheses, \Cref{ass:leaf-uniform-caption,ass:prototype-centroid-consistency}, and the decomposition in \Cref{ass:noisy-prototype}, we can apply \Cref{lem:caption-mean-decomposition} and \Cref{cor:zero-caption-bias}.
By \Cref{ass:leaf-uniform-caption}, $N_k>0$ for every non-root node $k$, so we can apply \Cref{prop:large-lambda}.
Factor \Cref{eq:large-lambda} using \Cref{eq:large-lambda-xi-def}, and substitute \Cref{eq:caption-mean-decomposition} with $\bm{b}_k=\bm{b}=\bzero$ to obtain \Cref{eq:large-lambda-prototype-noisy}.
\end{proof}

\subsection{Quantitative Remainder for the InfoNCE Expansion}
The asymptotic expansion identifies the leading empirical means; transfer at a specified regularization level also requires a quantitative remainder bound.

\begin{lemma}[Quantitative remainder for the large-$\lambda$ expansion]\label{lem:large-lambda-quant-remainder}
Let $Z_{\max}:=\max_{j\in[N]}\|\bz_j\|$, assume $N_k>0$ for every non-root node $k$, and let $N_{\min}:=\min_{k\in V\setminus\{\mathrm{root}\}}N_k$. For each non-root node $k\in V\setminus\{\mathrm{root}\}$, with $\xi_k^{\mathrm{nce}}$ as in \Cref{eq:large-lambda-xi-def}, define the \emph{prototype-decomposition residual} $\br_{k,\lambda}\in\R^d$ by
\begin{equation}\label{eq:rkl-def}
  \br_{k,\lambda}
  :=
  \frac{\lambda}{N_k}\,\ba_k^{\mathrm{nce},*}
  -
  \bigl(\bar{\bz}_k-\xi_k^{\mathrm{nce}}\bar{\bz}\bigr).
\end{equation}
Under the hypotheses of \Cref{cor:prototype-form-noisy}, $\bar{\bz}_k-\xi_k^{\mathrm{nce}}\bar{\bz}=\widetilde{\bm{\mu}}_k+\bar{\bm{\varepsilon}}_k-\xi_k^{\mathrm{nce}}\bar{\bm{\varepsilon}}$ with $\widetilde{\bm{\mu}}_k:=\bm{\mu}_k-\xi_k^{\mathrm{nce}}\bm{\mu}_{\mathrm{root}}$, so that, equivalently,
\begin{equation}\label{eq:rkl-form}
  \ba_k^{\mathrm{nce},*}
  =
  \frac{N_k}{\lambda}\Big(
    \widetilde{\bm{\mu}}_k
    +
    \bar{\bm{\varepsilon}}_k
    -
    \xi_k^{\mathrm{nce}}\,\bar{\bm{\varepsilon}}
    +
    \br_{k,\lambda}
  \Big).
\end{equation}
Then there exists a deterministic constant $C_{\mathrm{rem}}=C_{\mathrm{rem}}(M,N,N_{\min})$, depending only on $M$, $N$, and $N_{\min}$, such that for every $\lambda>0$,
\begin{equation}\label{eq:large-lambda-quant-remainder}
  \max_{k\in V\setminus\{\mathrm{root}\}}\|\br_{k,\lambda}\|
  \le
  \frac{C_{\mathrm{rem}}\,Z_{\max}^3}{\lambda}.
\end{equation}
\end{lemma}

\begin{proof}
We first bound the difference between the KKT identity and its uniform-softmax approximation. For every non-root node $k\in V\setminus\{\mathrm{root}\}$, define
\begin{equation}\label{eq:large-lambda-rho-def}
  \bm{\rho}_{k,\lambda}
  :=
  -\frac{1}{2\lambda}\sum_{j=1}^N \left(q_k^{(j)}(\bA^{\mathrm{nce},*})-\frac{1}{M}\right)\bz_j
  -
  \frac{1}{2\lambda}\sum_{j\in\cP_k}\sum_{j'=1}^N \left(r_{j'}^{(j)}(\bA^{\mathrm{nce},*})-\frac{1}{N}\right)\bz_{j'}.
\end{equation}
Substituting \Cref{eq:large-lambda-rho-def} into the stationarity condition in \Cref{eq:kkt-finite} gives
\begin{equation}\label{eq:large-lambda-exact-rho}
  \ba_k^{\mathrm{nce},*}
  =
  \frac{N_k}{\lambda}\bar{\bz}_k
  -
  \frac{1}{\lambda}\left(\frac{N}{2M}+\frac{N_k}{2}\right)\bar{\bz}
  +
  \bm{\rho}_{k,\lambda}.
\end{equation}

From \Cref{eq:kkt-finite} and the identities $\sum_k q_k^{(j)}(\bA^{\mathrm{nce},*})=1$, $\sum_{j'} r_{j'}^{(j)}(\bA^{\mathrm{nce},*})=1$, we have
\begin{equation}
  \max_{\ell\in[M]}\|\ba_\ell^{\mathrm{nce},*}\|
  \le
  \frac{1}{\lambda}\left(N+\tfrac{N}{2}+\tfrac{N}{2}\right)Z_{\max}
  =
  \frac{2N\,Z_{\max}}{\lambda}.
\end{equation}
For each image index $j\in[N]$, each entry of the i2t logit vector $\mathbf{s}^{(j)}$ has magnitude at most $\|\ba_\ell^{\mathrm{nce},*}\|\,\|\bz_j\|\le 2N Z_{\max}^2/\lambda$. For each $j\in[N]$, each entry of the t2i logit vector $\mathbf{u}^{(j)}$ has the same upper bound. Hence
\begin{equation}\label{eq:large-lambda-quant-remainder-1}
  \|\mathbf{s}^{(j)}\|
  \le
  \sqrt{M}\cdot\frac{2N\,Z_{\max}^2}{\lambda},
  \qquad
  \|\mathbf{u}^{(j)}\|
  \le
  \sqrt{N}\cdot\frac{2N\,Z_{\max}^2}{\lambda}.
\end{equation}
Applying \Cref{lem:softmax-uniform-lipschitz} with $K=M$ and $K=N$, respectively, gives
\begin{equation}\label{eq:softmax-q-deviation}
  \left|q_k^{(j)}(\bA^{\mathrm{nce},*})-\tfrac{1}{M}\right|
  \le
  2\,\|\mathbf{s}^{(j)}\|
  \le
  c_1(M,N)\,\frac{Z_{\max}^2}{\lambda},
\end{equation}
\begin{equation}\label{eq:softmax-r-deviation}
  \left|r_{j'}^{(j)}(\bA^{\mathrm{nce},*})-\tfrac{1}{N}\right|
  \le
  2\,\|\mathbf{u}^{(j)}\|
  \le
  c_1(M,N)\,\frac{Z_{\max}^2}{\lambda},
\end{equation}
uniformly over the relevant indices, where $c_1(M,N):=4N\sqrt{M+N}$.
Using the fact that $\|\bz_j\|\le Z_{\max}$ and \Crefrange{eq:softmax-q-deviation}{eq:softmax-r-deviation},
\begin{equation}
  \|\bm{\rho}_{k,\lambda}\|
  \le
  \frac{1}{2\lambda}\Big(N+N_k\,N\Big)\cdot c_1(M,N)\frac{Z_{\max}^2}{\lambda}\cdot Z_{\max}
  \le
  \frac{c_2(M,N)\,Z_{\max}^3}{\lambda^2},
\end{equation}
with $c_2(M,N):=N(1+N)\,c_1(M,N)/2$.

By \Cref{eq:large-lambda-xi-def}, the coefficient of $\bar{\bz}$ in \Cref{eq:large-lambda-exact-rho} equals $N_k\xi_k^{\mathrm{nce}}/\lambda$, and therefore we can rewrite \Cref{eq:large-lambda-exact-rho} as
\begin{equation}
  \ba_k^{\mathrm{nce},*}
  =
  \frac{N_k}{\lambda}\left(
    \bar{\bz}_k
    -
    \xi_k^{\mathrm{nce}}\bar{\bz}
  \right)
  +
  \bm{\rho}_{k,\lambda}.
\end{equation}
Comparing this identity with the definition of $\br_{k,\lambda}$ in \Cref{eq:rkl-def} gives
\begin{equation}\label{eq:rkl-rho-identity}
  \frac{N_k}{\lambda}\br_{k,\lambda}
  =
  \bm{\rho}_{k,\lambda},
\end{equation}
and therefore $\|\br_{k,\lambda}\|\le c_2(M,N)Z_{\max}^3/(N_k\lambda)$. Since $N_k\ge N_{\min}$ for every non-root node $k\in V\setminus\{\mathrm{root}\}$,
\begin{equation}
  \max_{k\in V\setminus\{\mathrm{root}\}}\|\br_{k,\lambda}\|
  \le
  \frac{C_{\mathrm{rem}}\,Z_{\max}^3}{\lambda},
\end{equation}
where $C_{\mathrm{rem}}(M,N,N_{\min}):=\frac{c_2(M,N)}{N_{\min}}$.
\end{proof}

\section{Shared Transfer Argument}\label{app:shared-transfer}

For $\lambda>0$, write $\{\ba_k\}$ for the minimizer under consideration, $\bA^{\mathrm{lin},*}$ of \Cref{eq:objective-linear-sampled-path} for the linear loss and $\bA^{\mathrm{nce},*}$ of \Cref{eq:objective-simplified-finite} for InfoNCE.
For both losses $\ba_k$ is a positive multiple of the empirical-mean difference $\bar{\bz}_k-\xi_k\bar{\bz}$: exactly so for the linear loss, and up to a remainder bounded by \Cref{lem:large-lambda-quant-remainder} for InfoNCE.
The two transfer proofs therefore share the same steps, which we present next. We allow a caption bias throughout, so that the main theorems are the special case of zero bias and the relaxations of \Cref{app:caption-relaxation} need only a bound on it.

In \Cref{lem:caption-bias-transfer} below, we need only a bound on the perturbation in a representation $\ba_k=c_k\left(\bm{\mu}_k+\text{perturbation}\right)$ with $c_k>0$.
It has three components, the caption bias, the sampling noise, and the finite regularization, which we define in turn below.

Recall from \Cref{def:caption-bias} that $\bm{b}_k=\bar{\bm{\mu}}_k-\bm{\mu}_k$ and $\bm{b}=\bar{\bm{\mu}}-\bm{\mu}_{\mathrm{root}}$, the gaps between the caption-averaged prototypes and the true ones; by \Cref{eq:caption-mean-decomposition} the bias contributes $\bm{b}_k-\xi_k\bm{b}$ to the empirical-mean difference.
The transfer bound uses that contribution only through an upper bound on its largest norm, so for fixed observed captions with non-root coverage and a fixed family of coefficients $\{\xi_k\in\R\}_{k\in V\setminus\{\mathrm{root}\}}$, we call $\delta_{\mathrm{bias}}\ge0$ a \emph{caption bias bound} if
\begin{equation}\label{eq:caption-bias-level}
  \max_{k\neq\mathrm{root}}\left\|\bm{b}_k-\xi_k\,\bm{b}\right\|
  \le
  \delta_{\mathrm{bias}}.
\end{equation}
The smallest caption bias bound is the left-hand side itself. We allow larger values so that we can apply \Cref{lem:caption-bias-transfer} directly with an upper bound on this maximum in \Cref{app:caption-relaxation}.

The error terms $D_\epsilon^{\mathrm{lin}}$ and $D_\epsilon^{\mathrm{nce}}$ of the main theorems carry over unchanged.
Throughout, $\xi_k$ are the coefficients of the loss under consideration, from \Cref{cor:linear-sampled-path-noisy-centroid-form} for the linear loss and \Cref{eq:large-lambda-xi-def} for InfoNCE, and $\xi_{\max}$, $N_{\min}$, and $\delta_{\mathrm{root}}$ are evaluated at the observed captions and those coefficients, and $\delta_{\mathrm{bias}}$ is a caption bias bound for them.
Write $D_\epsilon^{\mathrm{lin}}$ for the expression in \Cref{eq:linear-sampled-path-noisy-has-bound} and $D_\epsilon^{\mathrm{nce}}$ for its InfoNCE counterpart defined before \Cref{thm:large-lambda-has-noisy}, using the universal constant from \Cref{lem:subgaussian-empirical-means}.
Both expressions are defined for any observed captions with non-root coverage, regardless of caption symmetry, so we use the same $D_\epsilon^{\mathrm{lin}}$ and $D_\epsilon^{\mathrm{nce}}$ in the biased setting here and in the symmetric setting of the main theorems.

For InfoNCE we also choose a regularization threshold that makes the large-$\lambda$ remainder no larger than $D_\epsilon^{\mathrm{nce}}$.
For $\epsilon\in(0,1)$ and $D_\epsilon^{\mathrm{nce}}>0$, define
\begin{equation}\label{eq:large-lambda-threshold}
  R_\epsilon
  :=\max_{\ell\in V_{\mathrm{leaf}}}\|\bm{\mu}_\ell\|
  +C_{\max}\sigma\sqrt{d+\log(4N/\epsilon)},
  \qquad
  \lambda_0(\epsilon)
  :=\frac{C_{\mathrm{rem}}R_\epsilon^3}{D_\epsilon^{\mathrm{nce}}},
\end{equation}
where $C_{\max}>0$ is the universal constant from \Cref{lem:subgaussian-feature-max-norm} and $C_{\mathrm{rem}}=C_{\mathrm{rem}}(M,N,N_{\min})$ is the deterministic constant from \Cref{lem:large-lambda-quant-remainder}.
In particular, $R_\epsilon$ is independent of the captions, whereas $\lambda_0(\epsilon)$ depends on the observed captions through $C_{\mathrm{rem}}$ and $D_\epsilon^{\mathrm{nce}}$.

Recall the minimum non-root prototype norm $\rho_{\min}:=\min_{k\neq\mathrm{root}}\|\bm{\mu}_k\|$, which sets the scale of the threshold below.
For $\epsilon\in(0,1)$, the \emph{total error term} $D_\epsilon^{\mathrm{bias}}$ and the \emph{margin threshold} $\tau_\epsilon^{\mathrm{bias}}$ it induces are
\begin{equation}\label{eq:caption-total-deviation}
  D_\epsilon^{\mathrm{bias}}
  :=
  \begin{cases}
    D_\epsilon^{\mathrm{lin}}+\delta_{\mathrm{bias}}, & \text{linear contrastive loss},\\[2pt]
    2D_\epsilon^{\mathrm{nce}}+\delta_{\mathrm{bias}}, & \text{InfoNCE loss},
  \end{cases}
  \qquad
  \tau_\epsilon^{\mathrm{bias}}:=\frac{8D_\epsilon^{\mathrm{bias}}}{\rho_{\min}}.
\end{equation}
The caption bias enters additively in both cases, and the factor $2$ in the InfoNCE case appears because $D_\epsilon^{\mathrm{nce}}$ bounds each of two terms, the perturbation from the empirical means and the large-$\lambda$ remainder.

In \Cref{lem:caption-bias-transfer}, we assume neither \Cref{ass:leaf-uniform-caption} nor \Cref{ass:prototype-centroid-consistency}, and nothing about the global margin $\gamma_{\HAS}$; the captions enter only through $\delta_{\mathrm{bias}}$.

\newcommand{\CaptionBiasTransferConclusions}[1]{%
\begin{enumerate}[(i)]
  \item $\ba_k\neq\bzero$ for every non-root node $k$;
  \item every admissible triplet $(u,v,w)$ with $\gamma_{uvw}>#1$ satisfies $\cos(\ba_u,\ba_v)>\cos(\ba_u,\ba_w)$;
  \item $\HASViol^{\mathrm{txt}}(u)\le\HASViol^{\mathrm{img}}_{#1}(u)$ for every non-root non-leaf node $u$, with $\HASViol^{\mathrm{txt}}(u)$ computed on $\{\ba_k\}_{k\neq\mathrm{root}}$.
\end{enumerate}%
}

\begin{lemma}[\HAS transfer under biased captions]\label{lem:caption-bias-transfer}
Suppose \Cref{ass:noisy-prototype} (sub-Gaussian prototype variation) holds, $\rho_{\min}>0$, and the observed captions are deterministic with $N_k>0$ for every non-root node $k$. Fix $\epsilon\in(0,1)$, $\lambda>0$, and a caption bias bound $\delta_{\mathrm{bias}}$, and suppose $D_\epsilon^{\mathrm{bias}}<\rho_{\min}/4$; in the InfoNCE case suppose also $D_\epsilon^{\mathrm{nce}}>0$ and $\lambda\ge\lambda_0(\epsilon)$. Then, with probability at least $1-\epsilon$, all of the following hold simultaneously:
\CaptionBiasTransferConclusions{\tau_\epsilon^{\mathrm{bias}}}
\end{lemma}

\begin{proof}
\textbf{Step 1: Decompose the optimal embeddings.}
Fix the observed captions and the coefficients for the loss under consideration.
For every non-root node $k$, set
\begin{equation}\label{eq:caption-transfer-components}
  \be_k:=-\xi_k\bm{\mu}_{\mathrm{root}}+\bar{\bm{\varepsilon}}_k-\xi_k\bar{\bm{\varepsilon}},
  \qquad
  \widetilde{\bm{b}}_k:=\bm{b}_k-\xi_k\bm{b}.
\end{equation}
Under coverage, we apply \Cref{cor:linear-sampled-path-centroid-form} for the linear loss and \Cref{lem:large-lambda-quant-remainder} for InfoNCE to write $\ba_k$ as a positive multiple of $\bar{\bz}_k-\xi_k\bar{\bz}$ plus the remainder.
Under \Cref{ass:noisy-prototype}, every sample satisfies $\bz_j=\bm{\mu}_{y_j}+\bm{\varepsilon}_j$, which is the form required by \Cref{lem:caption-mean-decomposition}.
Substituting \Cref{eq:caption-mean-decomposition} into those empirical-mean expressions gives
\begin{equation}\label{eq:caption-bias-prototype-form}
  \ba_k
  =c_k\left(\bm{\mu}_k+\be_k+\widetilde{\bm{b}}_k+\br_{k,\lambda}\right),
\end{equation}
where $c_k=\frac{N_k}{\lambda}\left(1+\frac{1}{2(M-1)}+\frac{1}{2(N-1)}\right)$ and $\br_{k,\lambda}=\bzero$ for the linear loss, whereas $c_k=N_k/\lambda$ and $\br_{k,\lambda}$ is defined by \Cref{eq:rkl-def} for InfoNCE.
In both cases, coverage and $\lambda>0$ give $c_k>0$, and by \Cref{eq:caption-bias-level}, $\max_{k\neq\mathrm{root}}\|\widetilde{\bm{b}}_k\|\le\delta_{\mathrm{bias}}$.

\textbf{Step 2: Bound the stochastic deviation.}
For a failure probability $p\in(0,1)$, define
\begin{equation}\label{eq:caption-transfer-noise-level}
  \eta_p:=C\sigma\left(
    \sqrt{\frac{d+\log(M-1)+\log(4/p)}{N_{\min}}}
    +\xi_{\max}\sqrt{\frac{d+\log(4/p)}{N}}
  \right),
\end{equation}
and let $\cE_p:=\{\max_{k\neq\mathrm{root}}\|\bar{\bm{\varepsilon}}_k-\xi_k\bar{\bm{\varepsilon}}\|\le\eta_p\}$.
The observed captions are deterministic and have coverage, so the index sets $\cP_k$ are fixed and nonempty, and the variations $\bm{\varepsilon}_j$ are independent, centered, and sub-Gaussian by \Cref{ass:noisy-prototype}.
Apply \Cref{lem:subgaussian-empirical-means} with $\cK=V\setminus\{\mathrm{root}\}$, $I_k=\cP_k$ for every $k\in\cK$, the fixed coefficients $\xi_k$, and $u=\log(4/p)$.
Since $|\cK|=M-1$ and $\min_{k\in\cK}|I_k|=N_{\min}$, we obtain $\Pr(\cE_p)\ge1-p$ from \Cref{eq:finite-sample-noise-centroid-bound}.
On $\cE_p$, \Cref{eq:caption-transfer-components} and the triangle inequality give
\begin{equation}\label{eq:caption-transfer-base-deviation}
  \max_{k\neq\mathrm{root}}\|\be_k\|
  \le\delta_{\mathrm{root}}+\eta_p.
\end{equation}
For the linear loss, choose $p=\epsilon$.
Then $\delta_{\mathrm{root}}+\eta_\epsilon=D_\epsilon^{\mathrm{lin}}$ and the remainder is zero, so the total perturbation in \Cref{eq:caption-bias-prototype-form} is at most $D_\epsilon^{\mathrm{lin}}+\delta_{\mathrm{bias}}=D_\epsilon^{\mathrm{bias}}$ with probability at least $1-\epsilon$.

\textbf{Step 3: Control the InfoNCE remainder.}
For InfoNCE, choose $p=\epsilon/2$ in Step 2, so $\delta_{\mathrm{root}}+\eta_{\epsilon/2}=D_\epsilon^{\mathrm{nce}}$.
Under \Cref{ass:noisy-prototype}, we can also apply \Cref{lem:subgaussian-feature-max-norm} with failure probability $p=\epsilon/2$.
Thus $\cF_\epsilon:=\{\max_{j\in[N]}\|\bz_j\|\le R_\epsilon\}$ has probability at least $1-\epsilon/2$.
On $\cF_\epsilon$, by \Cref{lem:large-lambda-quant-remainder} and \Cref{eq:large-lambda-threshold}, for every $\lambda\ge\lambda_0(\epsilon)$,
\begin{equation}\label{eq:max-rem-linear-analysis}
  \max_{k\neq\mathrm{root}}\|\br_{k,\lambda}\|
  \le\frac{C_{\mathrm{rem}}R_\epsilon^3}{\lambda}
  \le D_\epsilon^{\mathrm{nce}}.
\end{equation}
The union bound gives $\Pr(\cE_{\epsilon/2}\cap\cF_\epsilon)\ge1-\epsilon$; independence of these events is not needed.
On their intersection, the total perturbation is at most $2D_\epsilon^{\mathrm{nce}}+\delta_{\mathrm{bias}}=D_\epsilon^{\mathrm{bias}}$.

\textbf{Step 4: Apply the per-triplet transfer bound.}
For either loss, we have established an event of probability at least $1-\epsilon$ on which
\begin{equation}
  \max_{k\neq\mathrm{root}}\|\be_k+\widetilde{\bm{b}}_k+\br_{k,\lambda}\|
  \le D_\epsilon^{\mathrm{bias}}<\rho_{\min}/4<\rho_{\min}.
\end{equation}
Apply \Cref{lem:per-triplet-margin-preservation} with reference vectors $\bm{\nu}_k=\bm{\mu}_k$, the positive scalings $c_k$ in \Cref{eq:caption-bias-prototype-form}, and the total perturbations $\be_k+\widetilde{\bm{b}}_k+\br_{k,\lambda}$.
With $\eta=D_\epsilon^{\mathrm{bias}}$ its threshold is $\tau=8D_\epsilon^{\mathrm{bias}}/\rho_{\min}=\tau_\epsilon^{\mathrm{bias}}$, so parts (i), (ii), and (iii) of that lemma are claims (i), (ii), and (iii).
\end{proof}

\section{Proof of Hierarchical Angular Separation Transfer for the Linear Contrastive Loss}\label{app:proof-linear-sampled-path-noisy-has}

\LinearSampledPathNoisyHASThm*
\begin{proof}
\textbf{Step 1: Establish zero caption bias.}
The observed captions are path-consistent by the setup in \Cref{sec:has-setup}, and the same setup has exactly $n$ images per leaf.
By \Cref{cor:zero-caption-bias} together with \Cref{ass:leaf-uniform-caption,ass:prototype-centroid-consistency}, $\bm{b}_k=\bzero$ for every non-root node $k$ and $\bm{b}=\bzero$.
Therefore, by \Cref{eq:caption-bias-level}, $\delta_{\mathrm{bias}}=0$ is a caption bias bound for the linear-loss coefficients.

\textbf{Step 2: Bound the error term.}
We obtain non-root coverage from \Cref{ass:leaf-uniform-caption} and the required variation model from \Cref{ass:noisy-prototype}, and we assume $\rho_{\min}>0$ and $D_\epsilon^{\mathrm{lin}}<\rho_{\min}/4$ in the theorem.
With $\delta_{\mathrm{bias}}=0$, we have $D_\epsilon^{\mathrm{bias}}=D_\epsilon^{\mathrm{lin}}<\rho_{\min}/4$ by \Cref{eq:caption-total-deviation}.

\textbf{Step 3: Apply the shared transfer lemma.}
We can apply \Cref{lem:caption-bias-transfer} with $\tau_\epsilon^{\mathrm{bias}}=8D_\epsilon^{\mathrm{lin}}/\rho_{\min}=\tau_\epsilon^{\mathrm{lin}}$.
Its part (iii) gives \Cref{eq:per-triplet-has-bound} with probability at least $1-\epsilon$; parts (i) and (ii) also establish nonzero optimal embeddings and preservation of every admissible triplet with margin greater than $\tau_\epsilon^{\mathrm{lin}}$.
\end{proof}

\section{Proof of Hierarchical Angular Separation Transfer for the InfoNCE Loss}\label{app:proof-large-lambda-has-noisy}

\LargeLambdaNoisyHASThm*
\begin{proof}
\textbf{Step 1: Establish zero caption bias.}
The observed captions are path-consistent by the setup in \Cref{sec:has-setup}, and the same setup has exactly $n$ images per leaf.
By \Cref{cor:zero-caption-bias} together with \Cref{ass:leaf-uniform-caption,ass:prototype-centroid-consistency}, $\bm{b}_k=\bzero$ for every non-root node $k$ and $\bm{b}=\bzero$.
Therefore, by \Cref{eq:caption-bias-level}, $\delta_{\mathrm{bias}}=0$ is a caption bias bound for the InfoNCE coefficients.

\textbf{Step 2: Bound the error term.}
We obtain non-root coverage from \Cref{ass:leaf-uniform-caption} and the required variation model from \Cref{ass:noisy-prototype}.
The theorem hypothesis $0<D_\epsilon^{\mathrm{nce}}<\rho_{\min}/8$ gives $D_\epsilon^{\mathrm{bias}}=2D_\epsilon^{\mathrm{nce}}<\rho_{\min}/4$ and makes the deterministic threshold in \Cref{eq:large-lambda-threshold} positive and finite.

\textbf{Step 3: Apply the shared transfer lemma.}
For every $\lambda\ge\lambda_0(\epsilon)$, we can therefore apply \Cref{lem:caption-bias-transfer} with $\tau_\epsilon^{\mathrm{bias}}=16D_\epsilon^{\mathrm{nce}}/\rho_{\min}=\tau_\epsilon^{\mathrm{nce}}$.
Its part (iii) gives \Cref{eq:per-triplet-has-bound} at $\tau_\epsilon^{\mathrm{nce}}$ with probability at least $1-\epsilon$; parts (i) and (ii) also establish nonzero optimal embeddings and preservation of every admissible triplet with margin greater than $\tau_\epsilon^{\mathrm{nce}}$.
\end{proof}

\section{Step-by-step construction of a Hierarchical Concept Embedding}\label{app:step_by_step_construction}

In this section, we provide an explicit construction for the hierarchical prototypes $\{\bm{\mu}_k\}_{k\in V}$ used in our theoretical analysis, introduced by \citet{nguyenHierarchicalConceptEmbedding2026}. Throughout this section the tree $\cT$ of \Cref{sec:hierarchical-angular-separation-definition} is balanced with depth $L\ge 2$ and branching factor $b\ge 2$, so every non-leaf node has $b$ children and there are $m=b^L$ leaves. The construction has the following properties: (i) the displacement $\bm{\mu}_k-\bm{\mu}_p$ from each strict ancestor $p$ to a non-root node $k$ is orthogonal to $\bm{\mu}_p$, (ii) sibling prototypes $\{\bm{\mu}_{u^{(j)}}\}_{j=1}^b$ form a regular simplex around their parent $\bm{\mu}_u$, and (iii) every child $\bm{\mu}_{u^{(j)}}$ is located in a fixed half-angle cone around its parent $\bm{\mu}_u$. The prototypes satisfy the averaging property in \Cref{lem:prototype-centroid}, while the angle schedule in \Cref{prop:hce-positive-margin} also ensures a positive \HAS margin.

Fix an internal node $u\in V$ at depth $l:=d(u)\ge 0$ with children $u^{(1)},\dots,u^{(b)}$, and write $\pi_0(u),\pi_1(u),\dots,\pi_l(u)$ for the ancestors of $u$ along the root-to-$u$ path, with $\pi_l(u)=u$. Assume the ancestor prototypes $\bm{\mu}_{\pi_0(u)},\dots,\bm{\mu}_{\pi_l(u)}$ have already been constructed and stack them into the matrix
\begin{equation}\label{eq:Al-def}
  \bA_u
  :=
  \begin{bmatrix}
    \bm{\mu}_{\pi_0(u)} & \bm{\mu}_{\pi_1(u)} & \cdots & \bm{\mu}_{\pi_l(u)}
  \end{bmatrix}
  \in \R^{d\times(l+1)}.
\end{equation}
The construction below produces the $b$ child prototypes $\{\bm{\mu}_{u^{(j)}}\}_{j=1}^{b}$ and proceeds inductively starting from any nonzero root prototype $\bm{\mu}_{\mathrm{root}}\in\R^d$.

\subsection{Geometric requirements}
\label{app:construction-requirements}

The child prototypes are required to satisfy the following three conditions, where $\theta_l\in(0,\pi/2)$ is a fixed cone half-angle for depth-$l$ children.

\myparagraph{(R1) Hierarchical orthogonality}
For every ancestor $\pi_k(u)$, $k=0,\dots,l$, and every child $j=1,\dots,b$,
\begin{equation}\label{eq:multi_ortho}
  \bigl(\bm{\mu}_{u^{(j)}}-\bm{\mu}_{\pi_k(u)}\bigr)^{\!\top}\bm{\mu}_{\pi_k(u)} = 0.
\end{equation}

\myparagraph{(R2) Regular $(b{-}1)$-simplex on siblings}
With the parent $\bm{\mu}_u$ as the simplex centroid,
\begin{equation}\label{eq:multi_simplex_ip}
  \bigl(\bm{\mu}_{u^{(j)}}-\bm{\mu}_u\bigr)^{\!\top}\bigl(\bm{\mu}_{u^{(k)}}-\bm{\mu}_u\bigr)
  =
  \begin{cases}
    \lambda_l^{2}, & j=k,\\[2pt]
    -\lambda_l^{2}/(b-1), & j\neq k,
  \end{cases}
\end{equation}
for a scale $\lambda_l>0$ chosen below.

\myparagraph{(R3) Cone half-angle around the parent}
For every $j=1,\dots,b$,
\begin{equation}\label{eq:multi_cone}
  \angle\bigl(\bm{\mu}_{u^{(j)}},\bm{\mu}_u\bigr) = \theta_l.
\end{equation}

\subsection{Feasible subspace and depth--dimension condition}
\label{app:construction-feasible-subspace}

The ancestor orthogonality conditions in (R1) intersect in the affine subspace
\begin{equation}\label{eq:Vl_def}
  \cV_u
  :=
  \bigl\{\bx\in\R^{d}\,:\,\bA_u^{\top}\bx = \bz_u\bigr\},
  \qquad
  \bz_u := \bigl[\|\bm{\mu}_{\pi_0(u)}\|^{2},\dots,\|\bm{\mu}_{\pi_l(u)}\|^{2}\bigr]^{\top}.
\end{equation}
When the ancestor prototypes are linearly independent, $\dim\cV_u = d-(l+1)$. Embedding a regular $(b{-}1)$-simplex inside $\cV_u$ for every internal node up to depth $L-1$ therefore requires the depth--dimension condition
\begin{equation}\label{eq:depth_dim_cond}
  d \;\ge\; L + b - 1,
\end{equation}
which makes the trade-off explicit: $L$ dimensions are absorbed by the ancestor-orthogonality constraints down to depth $L-1$, while the regular $(b-1)$-simplex of siblings requires $b-1$ additional dimensions~\citep{nguyenHierarchicalConceptEmbedding2026}.

\myparagraph{Limits on hierarchy depth}
\Cref{eq:depth_dim_cond} is necessary but not sufficient. Because each child is separated from its parent by a fixed angle $\theta_l$ (see (R3)), the children's cones must be nested inside their parent's cone. This, in turn, forces the half-angles $\{\theta_l\}_{l=0}^{L-1}$ to shrink monotonically with depth and to satisfy a recursive feasibility constraint coupling $\theta_l$, $\theta_{l-1}$, and the branching factor $b$. As $L$ increases, the admissible cone angles collapse and the inter-prototype margin $\gamma_{\HAS}$ degrades, establishing an effective upper bound on the depth $L$ for a fixed embedding dimension $d$ while preserving \HAS.
For the full analysis on the precise depth-angle-dimension trade-off and the resulting bound on $L$, we refer the reader to \citet{nguyenHierarchicalConceptEmbedding2026}.
The dimension condition in \Cref{eq:depth_dim_cond} makes the orthogonal simplex construction feasible, but a positive \HAS margin also requires control of the angles across depths. In \Cref{prop:hce-positive-margin}, we give a sufficient decay schedule for this construction.

\subsection{Construction}
\label{app:construction-steps}

We construct $\{\bm{\mu}_{u^{(j)}}\}_{j=1}^{b}$ in five steps.

\begin{enumerate}
  \item \textbf{Origin in $\cV_u$ {[\,for (R1)\,]}.}
    Take the parent itself, $\bg_u := \bm{\mu}_u$. Hierarchical orthogonality of $\bm{\mu}_u$ to all its ancestors (inductively) gives $\bm{\mu}_u^\top\bm{\mu}_{\pi_k(u)} = \|\bm{\mu}_{\pi_k(u)}\|^{2}$ for every $k\le l$, so $\bg_u\in\cV_u$.
  \item \textbf{Orthonormal basis of the offset directions {[\,for (R1)\,]}.}
    Because $d-l-1\ge b-1$ for every internal node at depth $l\le L-1$ by \Cref{eq:depth_dim_cond}, choose $\bU_u\in\R^{d\times(b-1)}$ with $\bA_u^{\!\top}\bU_u=\bm{0}$ and $\bU_u^{\!\top}\bU_u = I_{b-1}$, e.g.\ by taking any $b-1$ left singular vectors spanning $\ker(\bA_u^\top)$. Any vector in $\bg_u + \mathrm{range}(\bU_u)$ lies in $\cV_u$ and therefore satisfies (R1).
  \item \textbf{Canonical regular simplex in $\R^{b-1}$ {[\,for (R2)\,]}.}
    Choose vectors $\tilde{\bd}_1,\dots,\tilde{\bd}_b\in\R^{b-1}$ forming a centered regular simplex:
    \begin{equation}\label{eq:canonical-simplex-offsets}
      \sum_{j=1}^{b}\tilde{\bd}_j=\bzero,
      \qquad
      \tilde{\bd}_j^\top\tilde{\bd}_k
      =
      \mathbf{1}[j=k]-\frac{1}{b},
      \qquad
      j,k\in[b].
    \end{equation}
  \item \textbf{Scale to enforce the cone condition {[\,for (R3)\,]}.}
    Set
    \begin{equation}\label{eq:lambda_choice}
      \lambda_l \;:=\; \|\bm{\mu}_u\|\tan\theta_l,
      \qquad
      \bd_j \;:=\; \tfrac{\lambda_l}{\sqrt{(b-1)/b}}\,\tilde{\bd}_j,
      \qquad j\in[b],
    \end{equation}
    so that $\|\bd_j\|=\lambda_l$ and $\bd_j^\top\bd_k=-\lambda_l^2/(b-1)$ for $j\neq k$, satisfying (R2).
  \item \textbf{Embed and translate.}
    \begin{equation}\label{eq:child-embedding}
      \bm{\mu}_{u^{(j)}}
      \;:=\;
      \bm{\mu}_u + \bU_u\,\bd_j,
      \qquad j=1,\dots,b.
    \end{equation}
\end{enumerate}

\subsection{Properties}
\label{app:construction-properties}

\myparagraph{Compatibility with the assumptions}
Iterating the construction from the root to all leaves gives a set of prototypes $\{\bm{\mu}_k\}_{k\in V}$ where (i) for every non-root node $k$ and strict ancestor $p$, the displacement $\bm{\mu}_k-\bm{\mu}_p$ is orthogonal to $\bm{\mu}_p$, and (ii) the children of any internal node form a centered regular simplex around their parent. Property~(i) yields $\|\bm{\mu}_k\|>0$ for every non-root $k$ provided the depth-wise scales $\|\bm{\mu}_u\|\tan\theta_l$ are positive, so $\rho_{\min}>0$. Under the angle schedule in \Cref{prop:hce-positive-margin}, the prototypes also have $\gamma_{\HAS}>0$. Property~(ii) is exactly the centered-simplex hypothesis used in \Cref{lem:prototype-centroid}, so the descendant average of leaf prototypes equals the ancestor prototype. As this holds for every internal node, the resulting set of prototypes is consistent with \Cref{ass:prototype-centroid-consistency}.

The centroid property holds for any choice of the angles, whereas a positive \HAS margin needs separation between descendants and off-branch nodes.
\begin{proposition}[Positive \HAS margin under geometric angle decay]\label{prop:hce-positive-margin}
Suppose the prototypes are constructed as in \Cref{app:construction-steps} with $0<\theta_0<\pi/2$ and some $q$ such that
\begin{align}
  0<q&\le1/4,\label{eq:hce-angle-rate}\\
  \theta_{r+1}&\le q\theta_r\quad(r\in\{0,\dots,L-2\}).\label{eq:hce-angle-recursion}
\end{align}
Then the global margin $\gamma_{\HAS}$ of \Cref{eq:prototype-has-margin} is positive, so the prototype family satisfies \HAS.
\end{proposition}

\begin{proof}
Fix an admissible triplet $(u,v,w)$, let $p$ be the lowest common ancestor of $u$ and $w$, and write $l:=d(p)$. The nodes $u$ and $w$ lie below distinct children $a$ and $a'$ of $p$. By \Cref{eq:multi_simplex_ip,eq:multi_cone}, their sibling angle $\alpha_l:=\angle(\bm{\mu}_a,\bm{\mu}_{a'})$ satisfies
\begin{align}\label{eq:alpha_l_angle_bound}
  \cos\alpha_l
  &=\frac{\bm{\mu}_a^\top\bm{\mu}_{a'}}{\|\bm{\mu}_a\|\|\bm{\mu}_{a'}\|}
  &&\text{(definition of $\alpha_l$)}\notag\\
  &=\frac{\|\bm{\mu}_p\|^2+(\bm{\mu}_a-\bm{\mu}_p)^\top(\bm{\mu}_{a'}-\bm{\mu}_p)}{\|\bm{\mu}_p\|^2+\lambda_l^2}
  &&\text{(by \Cref{eq:multi_ortho,eq:multi_simplex_ip})}\notag\\
  &=\frac{\|\bm{\mu}_p\|^2-\lambda_l^2/(b-1)}{\|\bm{\mu}_p\|^2+\lambda_l^2}
  &&\text{(by \Cref{eq:multi_simplex_ip})}\notag\\
  &=\frac{1-\lambda_l^2/((b-1)\|\bm{\mu}_p\|^2)}{1+\lambda_l^2/\|\bm{\mu}_p\|^2}
  &&\text{(dividing by $\|\bm{\mu}_p\|^2$)}\notag\\
  &=\frac{1-\tan^2\theta_l/(b-1)}{1+\tan^2\theta_l}
  &&\text{(by \Cref{eq:lambda_choice})}\notag\\
  &=\cos^2\theta_l-\frac{\sin^2\theta_l}{b-1}
  &&\text{(multiplying by $\cos^2\theta_l$)}\notag\\
  &<\cos\theta_l
  &&\text{(since $0<\cos\theta_l<1$ and $b\ge2$)}\notag\\
  \iff \alpha_l&>\theta_l
  &&\text{(since cosine decreases on $[0,\pi]$)}.
\end{align}
Set $T_l:=\sum_{j=l+1}^{L-1}\theta_j$. Applying the angle triangle inequality first through $u$ and then through $w$ gives
\begin{align*}
  \alpha_l
  &\le\angle(\bm{\mu}_a,\bm{\mu}_u)+\angle(\bm{\mu}_u,\bm{\mu}_{a'})\\
  &\le\angle(\bm{\mu}_a,\bm{\mu}_u)+\angle(\bm{\mu}_u,\bm{\mu}_w)+\angle(\bm{\mu}_w,\bm{\mu}_{a'}).
\end{align*}
By \Cref{eq:multi_cone} and repeated angle triangle inequalities along the three ancestor--descendant paths,
\begin{align*}
  \angle(\bm{\mu}_u,\bm{\mu}_a)
  &\le\textstyle\sum_{j=l+1}^{d(u)-1}\theta_j\le T_l,\\
  \angle(\bm{\mu}_w,\bm{\mu}_{a'})
  &\le\textstyle\sum_{j=l+1}^{d(w)-1}\theta_j\le T_l,\\
  \angle(\bm{\mu}_u,\bm{\mu}_v)
  &\le\textstyle\sum_{j=d(u)}^{d(v)-1}\theta_j\le T_l.
\end{align*}
Rearranging the sibling-angle bound and applying these path bounds gives
\begin{align}
  \angle(\bm{\mu}_u,\bm{\mu}_w)
  &\ge\alpha_l-\angle(\bm{\mu}_u,\bm{\mu}_a)-\angle(\bm{\mu}_w,\bm{\mu}_{a'})
  \notag\\
  &\ge\alpha_l-T_l-\angle(\bm{\mu}_w,\bm{\mu}_{a'})
  &&\text{(since $\angle(\bm{\mu}_u,\bm{\mu}_a)\le T_l$)}\notag\\
  &\ge\alpha_l-2T_l
  &&\text{(since $\angle(\bm{\mu}_w,\bm{\mu}_{a'})\le T_l$)}\notag\\
  \angle(\bm{\mu}_u,\bm{\mu}_w)-\angle(\bm{\mu}_u,\bm{\mu}_v)
  &\ge\alpha_l-2T_l-\angle(\bm{\mu}_u,\bm{\mu}_v)\notag\\
  &\ge\alpha_l-3T_l
  &&\text{(since $\angle(\bm{\mu}_u,\bm{\mu}_v)\le T_l$)}.
\end{align}
It remains to show that $3T_l<\alpha_l$. Since $q>0$ by \Cref{eq:hce-angle-rate}, iterating \Cref{eq:hce-angle-recursion} gives $\theta_{l+m}\le q^m\theta_l$ for every $m\in\{1,\dots,L-l-1\}$. Thus
\begin{align}\label{eq:three_T_l_bound}
  3T_l
  &=3\sum_{j=l+1}^{L-1}\theta_j
  &&\text{(definition of $T_l$)}\notag\\
  &\le 3\theta_l\sum_{m=1}^{L-l-1}q^m
  &&\text{(by \Cref{eq:hce-angle-recursion})}\notag\\
  &\le 3\theta_l\sum_{m=1}^{\infty}q^m
  &&\text{(since $q>0$ by \Cref{eq:hce-angle-rate})}\notag\\
  &=\frac{3q}{1-q}\theta_l
  &&\text{(geometric series)}\notag\\
  &\le\theta_l
  &&\text{(since $q\le1/4$ by \Cref{eq:hce-angle-rate})}\notag\\
  &<\alpha_l
  &&\text{(by \Cref{eq:alpha_l_angle_bound})}.
\end{align}
Thus the angle margin is positive. Cosine decreases with angle on $[0,\pi]$, so the triplet margin is positive. Since $L\ge2$ and $b\ge2$, admissible triplets exist, and their margin is positive.
\end{proof}

\section{Additional Experiments}\label{app:additional-experiments}
All experiments are run on a server equipped with NVIDIA RTX A5000 GPUs (24\,GB memory each). The CLIP empirical evaluations of \Cref{sec:hierarchical-angular-separation-empirical} use publicly released CLIP checkpoints over the ImageNet-1K training set and the WordNet hierarchy. The synthetic experiments of \Cref{sec:exp-low-lambda} run on a single A5000.

\subsection{Hierarchical Angular Separation on CLIP embeddings}\label{app:additional-experiments-has}

\subsubsection{Controls and robustness checks for approximate image-side \HAS}\label{app:raw-pixel-has}
\Cref{tab:raw-pixel-has} measures every representation under the disjoint-split protocol of \Cref{app:image-has-disjoint-split-control}: we partition the images in each ImageNet-1K leaf into even and odd halves, estimate every node embedding from each half separately by descendant-leaf-weighted instance aggregation, and average \HASViol over the two directions. For raw pixels, we apply CLIP's resize, center-crop, and channel-normalization preprocessing without an encoder and flatten each image into a $150{,}528$-dimensional vector. The Gaussian control draws independent isotropic Gaussian vectors in $\R^{768}$, normalizes each to unit length, and uses the same per-leaf sample counts. Every condition uses the WordNet hierarchy, anchor set, and $95$th-percentile descendant threshold of \Cref{sec:hierarchical-angular-separation-empirical}, and its hierarchy-shuffled null reassigns leaf groups within that hierarchy while preserving the representation and the aggregation procedure.

For random ViT, the original-assignment value reports mean $\pm$ sample standard deviation over three weight initializations, while the null pools five shuffles per initialization.
For Gaussian vectors, the original-assignment value reports mean $\pm$ standard deviation over $40$ independent draws, while the null pools five shuffles for each of the first five draws.
The raw-pixel and pretrained CLIP nulls use five shuffles of their fixed representations.

The randomly initialized row of \Cref{tab:raw-pixel-has} averages three ViT-L/14 random initializations, so it leaves open whether approximate \HAS at random initialization depends on the architecture. We therefore repeat the measurement across visual backbones, with at least three independent random initializations each.

\begin{table}[H]
\centering
\caption{Approximate image-side \HAS at random initialization across visual backbones, measured with the disjoint-split protocol. Each backbone is evaluated with $S$ independent random initializations, and both columns report the mean $\pm$ sample standard deviation across those samplings; each sampling's null is itself a mean over five hierarchy shuffles. The ViT-L/14 row is the randomly initialized row of \Cref{tab:raw-pixel-has}.}
\label{tab:image-has-backbone-seed}
\begin{tabular}{lccc}
\toprule
Backbone & $S$ & \HASViol (\%) $\downarrow$ & \HASViol shuffled (\%) $\downarrow$ \\
\midrule
ViT-B/16 & $6$ & $28.55 \pm 0.34$ & $57.20 \pm 0.07$ \\
ViT-B/32 & $3$ & $29.58 \pm 0.13$ & $57.39 \pm 0.15$ \\
ViT-L/14 & $3$ & $27.88 \pm 0.24$ & $56.80 \pm 0.12$ \\
ViT-H/14 & $3$ & $27.63 \pm 0.56$ & $56.62 \pm 0.03$ \\
RN50 & $3$ & $28.53 \pm 1.66$ & $59.45 \pm 1.44$ \\
\midrule
ViT-L/14@336 & $1$ & $27.83$ & $56.83 \pm 0.33$ \\
\bottomrule
\end{tabular}
\end{table}

Across architectures and random seeds, the violation percentage of every backbone is far below its own shuffled null.

\subsubsection{Robustness to the prototype construction and the taxonomy}\label{app:construction-taxonomy}
Image-side \HAS could depend on two choices in the setup of \Cref{sec:hierarchical-angular-separation-empirical}: the descendant-leaf weighting used to build prototypes, and the WordNet taxonomy itself. We repeat the disjoint-split measurement with each choice replaced in turn. First, we replace the descendant-leaf mean by a uniform mean over child prototypes, so that every child contributes equally regardless of how many images it has. Second, we replace ImageNet-1K and WordNet by CUB-200~\citep{wah2011caltech} with its own bird taxonomy, an independent hierarchy of different depth and branching. Chance level depends on both choices, so we compare each condition against its own hierarchy-shuffled null rather than against a common baseline.

\begin{table}[H]
\centering
\caption{Image-side \HAS is robust to the prototype construction and to the taxonomy. Each condition is paired with its own hierarchy-shuffled null (mean $\pm$ standard deviation over five shuffles). The first row restates the ViT-L/14 measurements of \Cref{tab:raw-pixel-has} at the precision used there; its randomly initialized entries are the seed-$0$ initialization rather than the mean over three initializations reported there.}
\label{tab:image-has-construction-taxonomy}
\resizebox{\linewidth}{!}{%
\begin{tabular}{lcccc}
\toprule
& \multicolumn{2}{c}{Pretrained OpenAI CLIP} & \multicolumn{2}{c}{Randomly initialized} \\
\cmidrule(lr){2-3}\cmidrule(lr){4-5}
Setting & \HASViol (\%) & \HASViol shuffled (\%) & \HASViol (\%) & \HASViol shuffled (\%) \\
\midrule
ImageNet / WordNet, descendant-leaf mean & $2.95$ & $32.93 \pm 0.48$ & $28.15$ & $56.70 \pm 0.42$ \\
ImageNet / WordNet, uniform child mean & $3.9$ & $34.5 \pm 0.3$ & $29.1$ & $56.8 \pm 0.5$ \\
CUB-200 / bird taxonomy & $3.8$ & $41.2 \pm 0.8$ & $33.6$ & $66.3 \pm 1.4$ \\
\bottomrule
\end{tabular}
}
\end{table}

In all three conditions, the violation percentage of the pretrained encoder is roughly an order of magnitude below its own null, while that of the randomly initialized encoder is roughly half of its null. Neither the prototype construction nor the choice of taxonomy therefore changes the conclusion: pretraining accounts for roughly the same share of the achievable separation in every setting, and approximate separation at random initialization appears in every setting.

\begin{figure}[t]
\centering
\begin{subfigure}{0.32\linewidth}
  \centering
  \includegraphics[width=\linewidth]{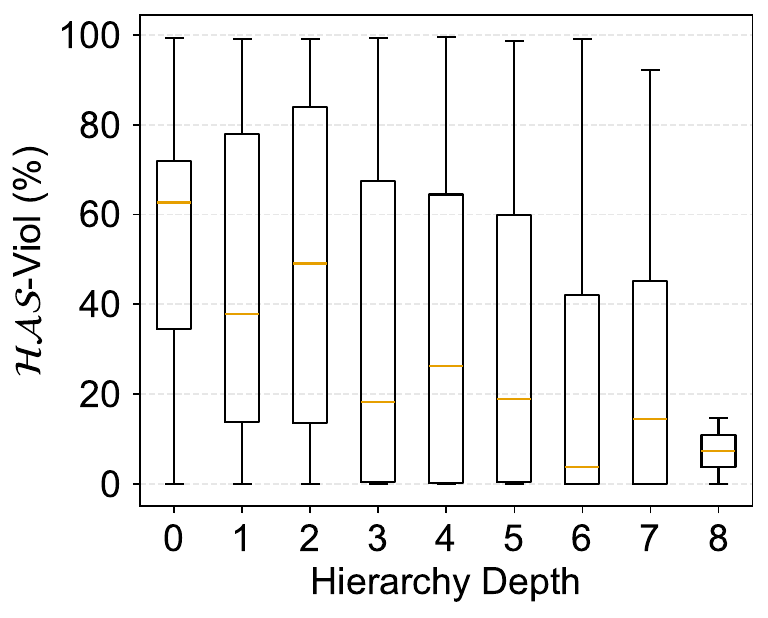}
  \caption{Randomly initialized.}
\end{subfigure}
\hfill
\begin{subfigure}{0.32\linewidth}
  \centering
  \includegraphics[width=\linewidth]{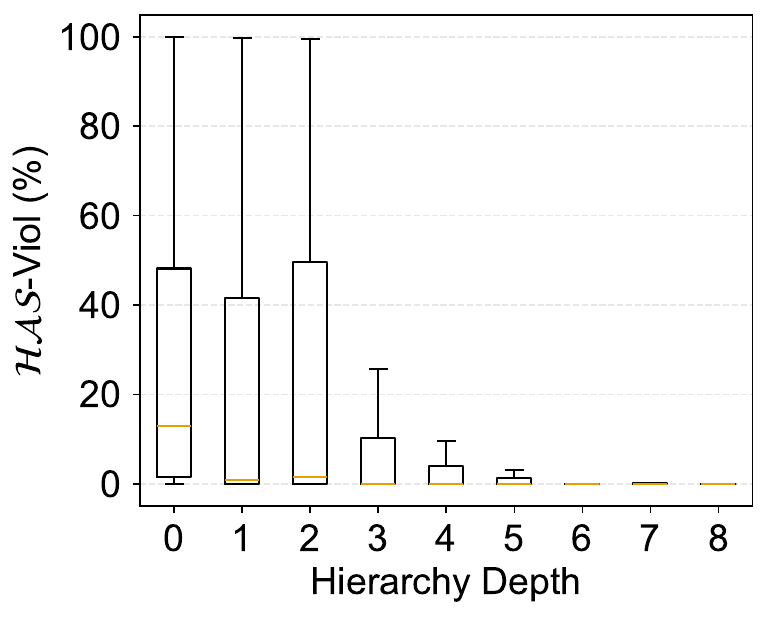}
  \caption{OpenAI.}
\end{subfigure}
\hfill
\begin{subfigure}{0.32\linewidth}
  \centering
  \includegraphics[width=\linewidth]{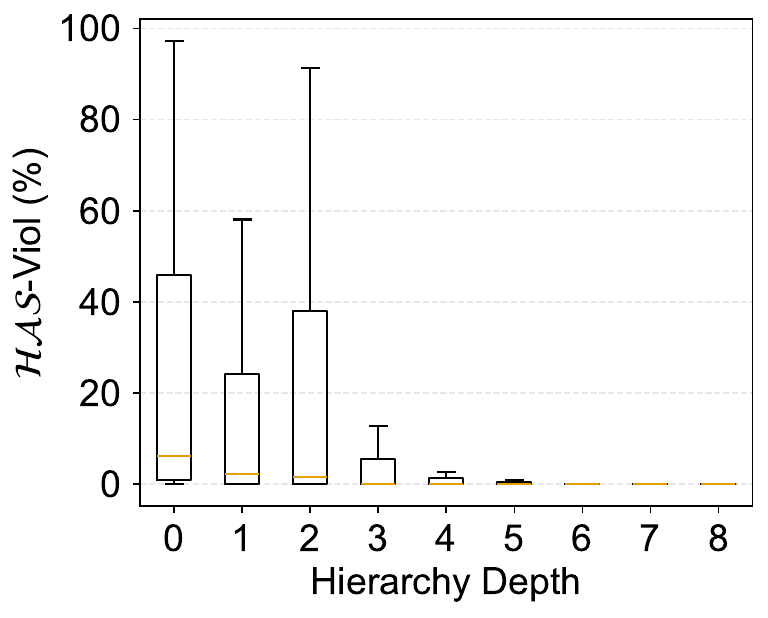}
  \caption{OpenCLIP (DataComp-1B).}
\end{subfigure}
\caption{\HAS violation percentage by depth on ViT-L-14 CLIP \emph{text} embeddings (instance aggregation): \HAS appears with pretraining.}
\label{fig:has-instance-agg-violation}
\end{figure}

\myparagraph{Instance aggregation on the text side (ViT-L/14)}
\Cref{fig:has-instance-agg-violation} repeats the text experiment of \Cref{sec:hierarchical-angular-separation-empirical} estimating concept embeddings by instance aggregation instead of direct concept encoding, again on ViT-L-14, and reaches the same conclusion: the median violation percentages are concentrated near zero for both pretrained models and substantially larger for the randomly initialized model.

\begin{figure}[t]
\centering
\begin{minipage}[c]{0.025\linewidth}
  \centering
  \raisebox{2.7\baselineskip}{\rotatebox{90}{\small Text}}
\end{minipage}%
\hfill
\begin{subfigure}[c]{0.475\linewidth}
  \centering
  \includegraphics[width=\linewidth]{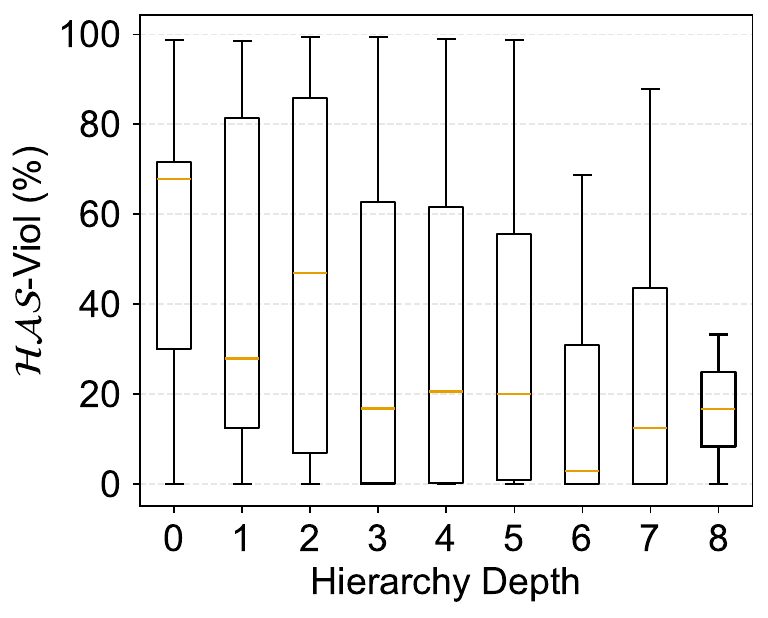}
  \caption{Randomly initialized.}
\end{subfigure}
\hfill
\begin{subfigure}[c]{0.475\linewidth}
  \centering
  \includegraphics[width=\linewidth]{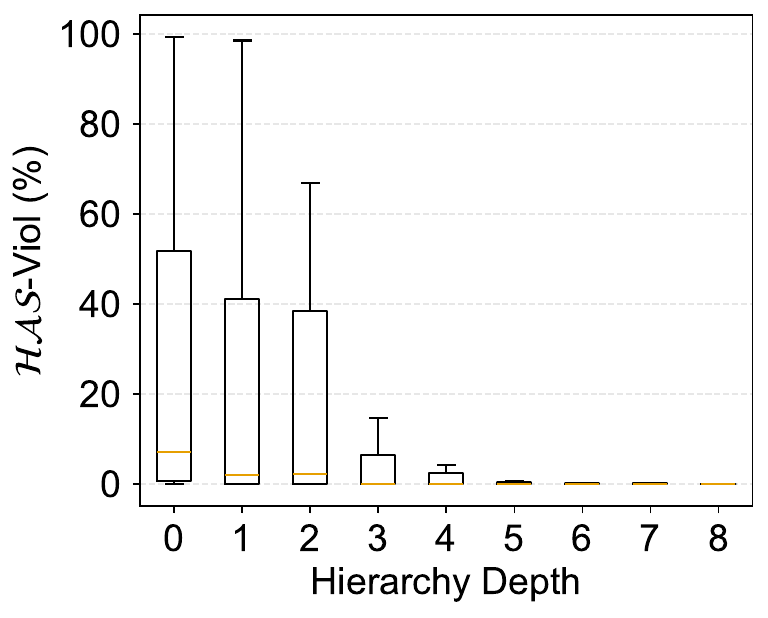}
  \caption{OpenCLIP (DFN-5B).}
\end{subfigure}

\vspace{0.5em}

\begin{minipage}[c]{0.025\linewidth}
  \centering
  \raisebox{2.7\baselineskip}{\rotatebox{90}{\small Image}}
\end{minipage}%
\hfill
\begin{subfigure}[c]{0.475\linewidth}
  \centering
  \includegraphics[width=\linewidth]{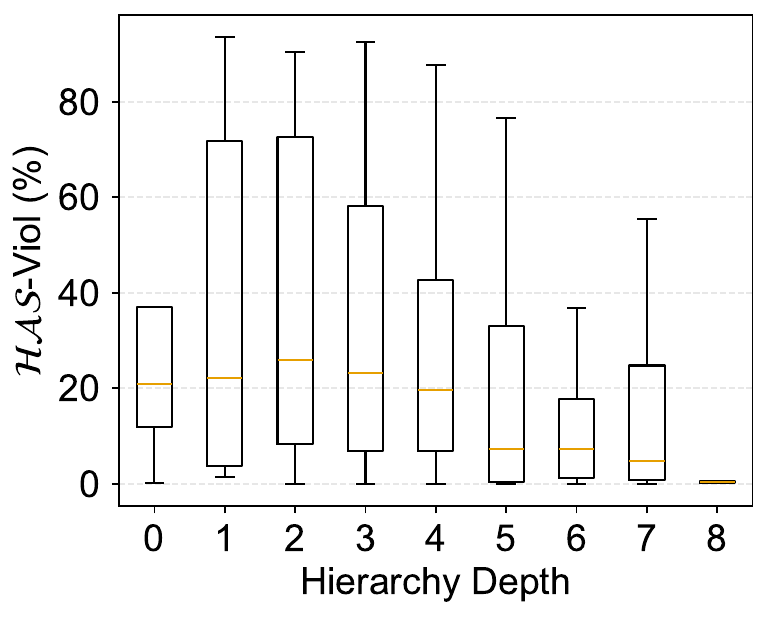}
  \caption{Randomly initialized.}
\end{subfigure}
\hfill
\begin{subfigure}[c]{0.475\linewidth}
  \centering
  \includegraphics[width=\linewidth]{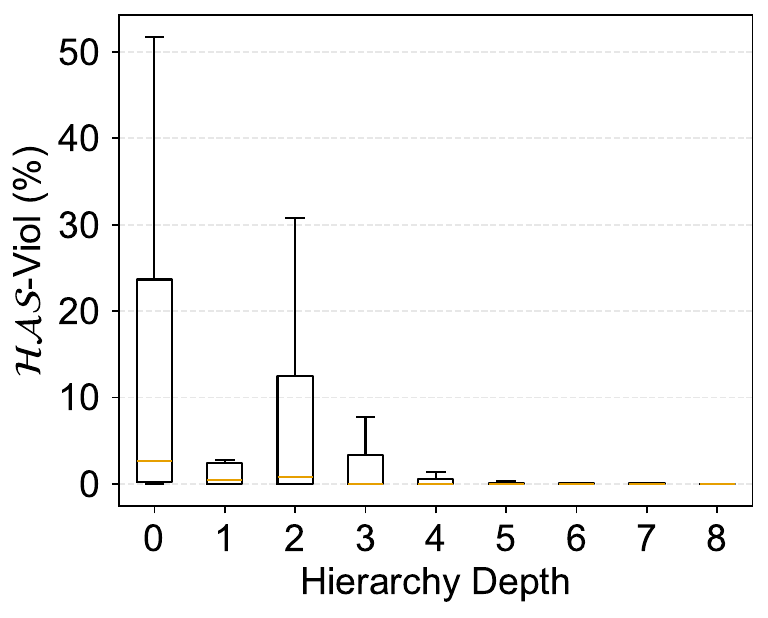}
  \caption{OpenCLIP (DFN-5B).}
\end{subfigure}
\caption{\HAS violation percentage by depth on ViT-H/14 CLIP embeddings. \emph{Top:} \emph{text} embeddings (instance aggregation) -- \HAS appears with pretraining. \emph{Bottom:} \emph{image} embeddings (instance aggregation): approximate \HAS appears at random initialization and strengthens after pretraining.}
\label{fig:has-violation-vith}
\end{figure}

\myparagraph{ViT-H/14 results}
We further repeat the experiment on ViT-H/14, comparing OpenCLIP trained on DFN-5B~\citep{fangDataFilteringNetworks2023} against a randomly initialized ViT-H/14. \Cref{fig:has-violation-vith} groups text embeddings (top; instance aggregation) and image embeddings (bottom; instance aggregation) and reproduces the same qualitative pattern.

\begin{figure}[t]
\centering
\begin{minipage}[c]{0.025\linewidth}
  \centering
  \raisebox{2.7\baselineskip}{\rotatebox{90}{\small Text}}
\end{minipage}%
\hfill
\begin{subfigure}[c]{0.31\linewidth}
  \centering
  \includegraphics[width=\linewidth]{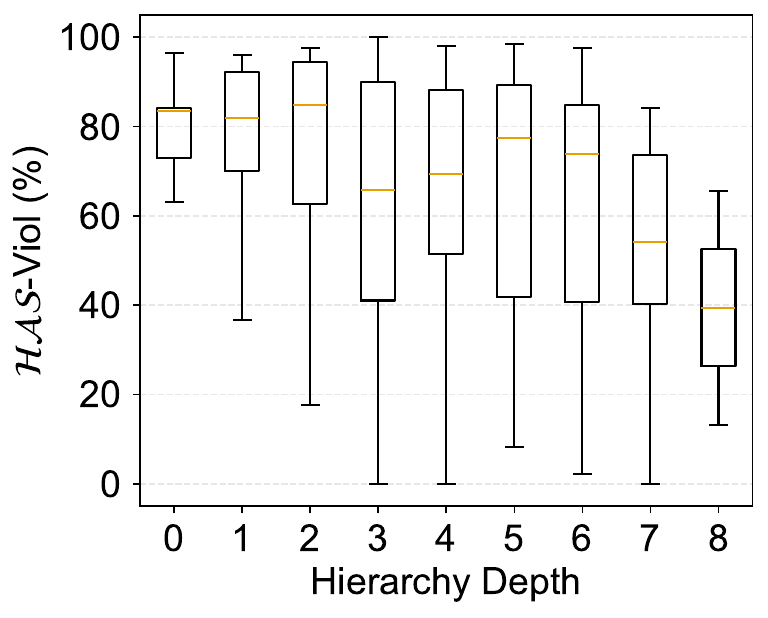}
  \caption{Randomly initialized.}
\end{subfigure}
\hfill
\begin{subfigure}[c]{0.31\linewidth}
  \centering
  \includegraphics[width=\linewidth]{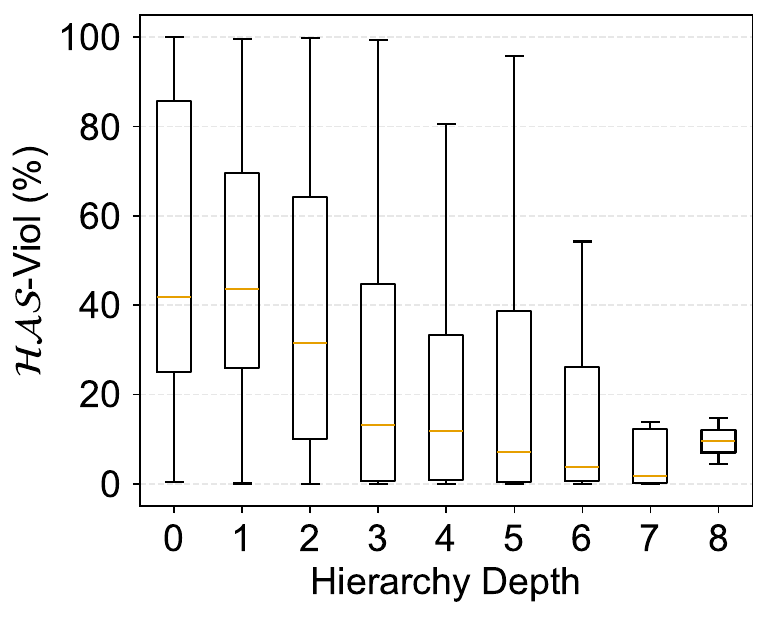}
  \caption{OpenAI.}
\end{subfigure}
\hfill
\begin{subfigure}[c]{0.31\linewidth}
  \centering
  \includegraphics[width=\linewidth]{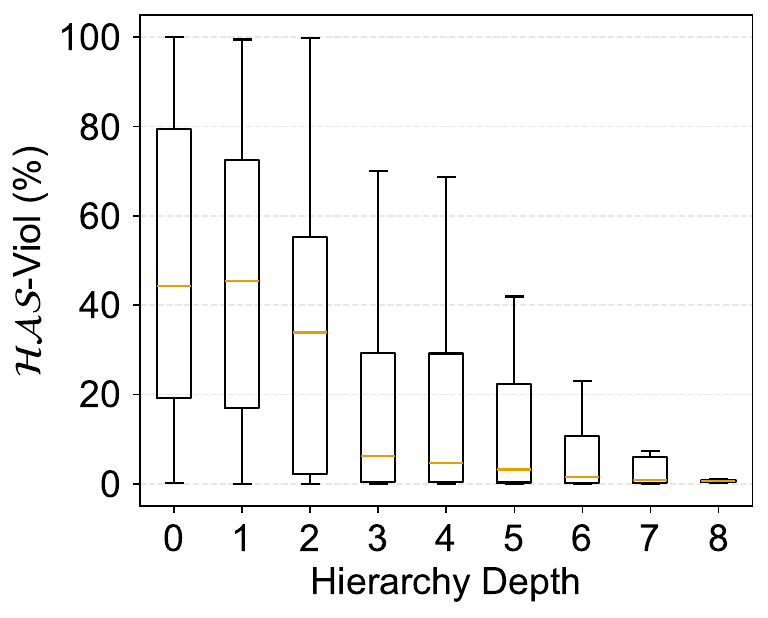}
  \caption{OpenCLIP (DataComp-1B).}
\end{subfigure}

\vspace{0.5em}

\begin{minipage}[c]{0.025\linewidth}
  \centering
  \raisebox{2.7\baselineskip}{\rotatebox{90}{\small Image}}
\end{minipage}%
\hfill
\begin{subfigure}[c]{0.31\linewidth}
  \centering
  \includegraphics[width=\linewidth]{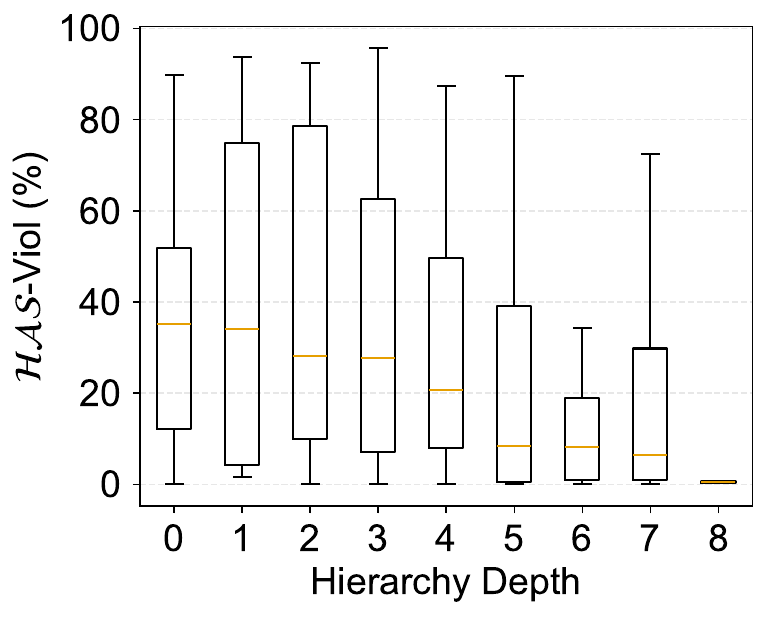}
  \caption{Randomly initialized.}
\end{subfigure}
\hfill
\begin{subfigure}[c]{0.31\linewidth}
  \centering
  \includegraphics[width=\linewidth]{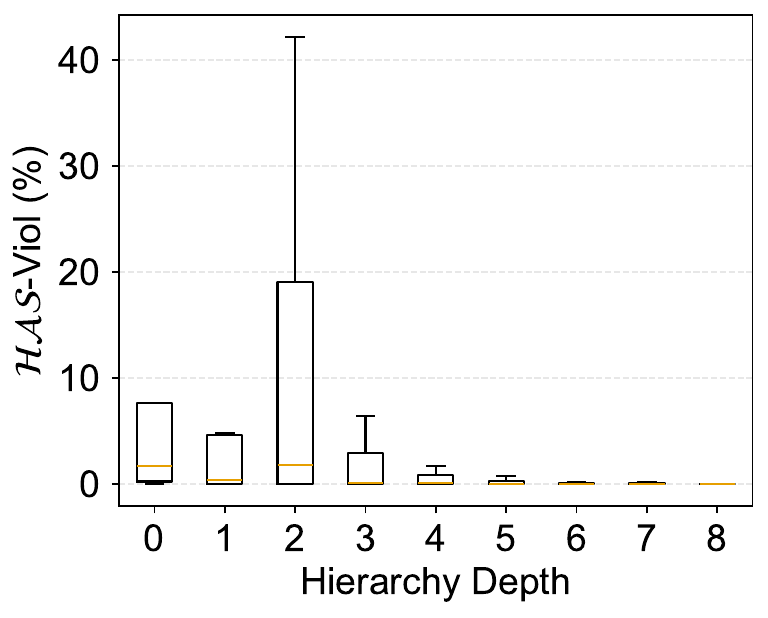}
  \caption{OpenAI.}
\end{subfigure}
\hfill
\begin{subfigure}[c]{0.31\linewidth}
  \centering
  \includegraphics[width=\linewidth]{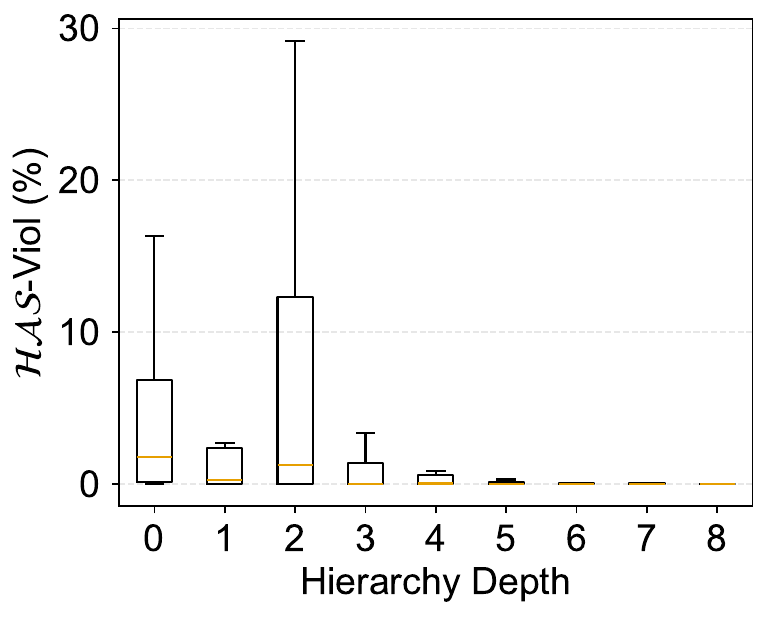}
  \caption{OpenCLIP (DataComp-1B).}
\end{subfigure}
\caption{\HAS violation percentage by depth on ViT-B/16 CLIP embeddings. \emph{Top:} \emph{text} embeddings (direct concept encoding): \HAS appears with pretraining. \emph{Bottom:} \emph{image} embeddings (instance aggregation): approximate \HAS appears at random initialization and strengthens after pretraining.}
\label{fig:has-violation-vitb}
\end{figure}

\begin{figure}[t]
\centering
\begin{subfigure}{0.32\linewidth}
  \centering
  \includegraphics[width=\linewidth]{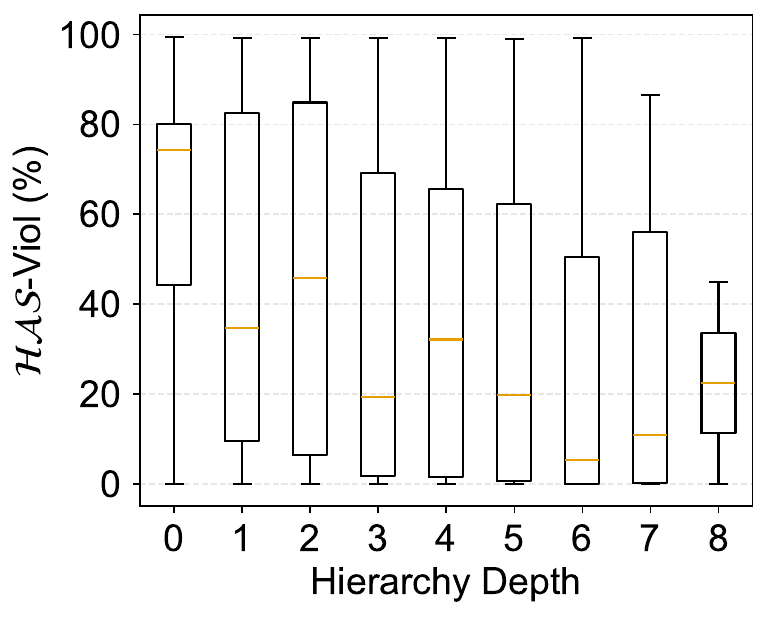}
  \caption{Randomly initialized.}
\end{subfigure}
\hfill
\begin{subfigure}{0.32\linewidth}
  \centering
  \includegraphics[width=\linewidth]{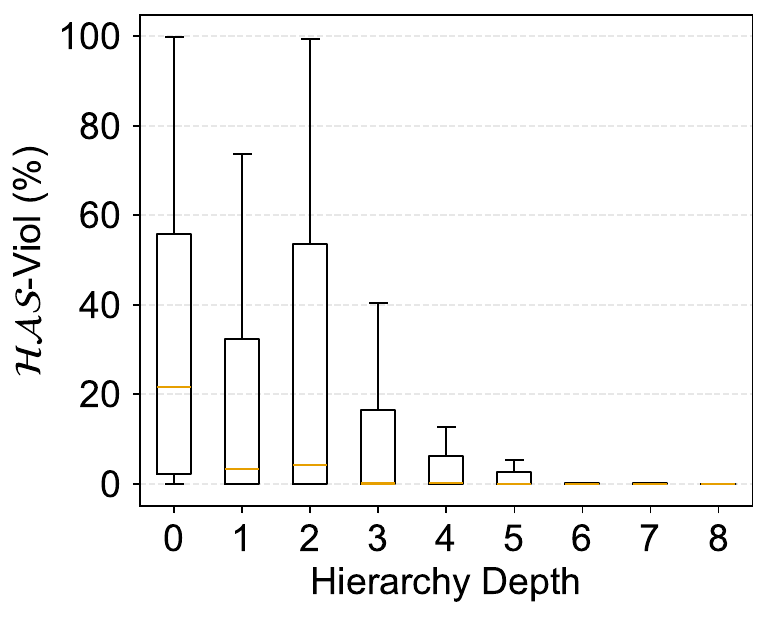}
  \caption{OpenAI.}
\end{subfigure}
\hfill
\begin{subfigure}{0.32\linewidth}
  \centering
  \includegraphics[width=\linewidth]{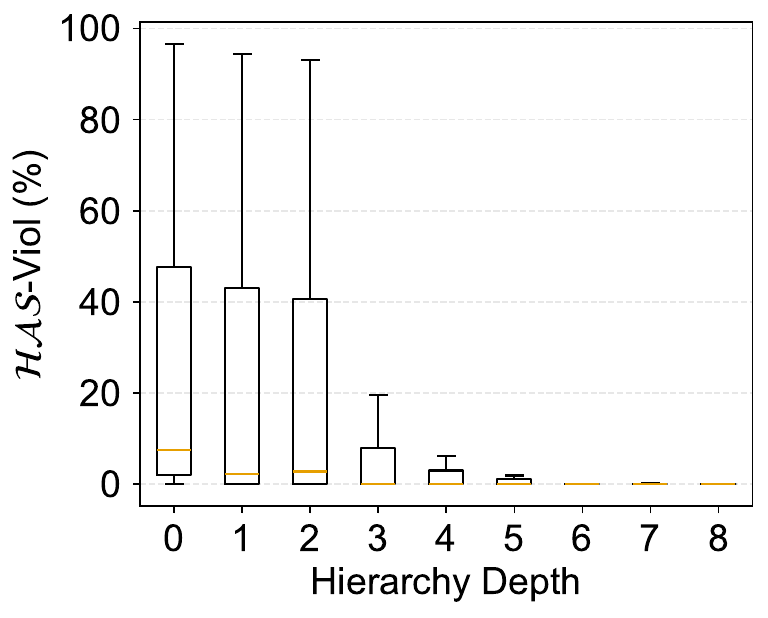}
  \caption{OpenCLIP (DataComp-1B).}
\end{subfigure}
\caption{\HAS violation percentage by depth on ViT-B/16 CLIP \emph{text} embeddings (instance aggregation): \HAS appears with pretraining.}
\label{fig:has-text-violation-vitb-instance-agg}
\end{figure}

\myparagraph{ViT-B/16 results}
We additionally repeat the main-text experiment on a smaller backbone, ViT-B/16, comparing OpenAI CLIP~\citep{radford2021learning} and OpenCLIP~\citep{cherti2023reproducible} trained on DataComp-1B~\citep{gadre2023datacomp} against a randomly initialized ViT-B/16. \Cref{fig:has-violation-vitb} (text via direct concept encoding on top, image via instance aggregation on bottom) and \Cref{fig:has-text-violation-vitb-instance-agg} (text, instance aggregation) reproduce the same qualitative pattern as on ViT-L/14: text \HAS appears with pretraining, while image embeddings show approximate \HAS at random initialization.

\begin{figure}[t]
\centering
\begin{minipage}[c]{0.025\linewidth}
  \centering
  \raisebox{2.7\baselineskip}{\rotatebox{90}{\small Text}}
\end{minipage}%
\hfill
\begin{subfigure}[c]{0.31\linewidth}
  \centering
  \includegraphics[width=\linewidth]{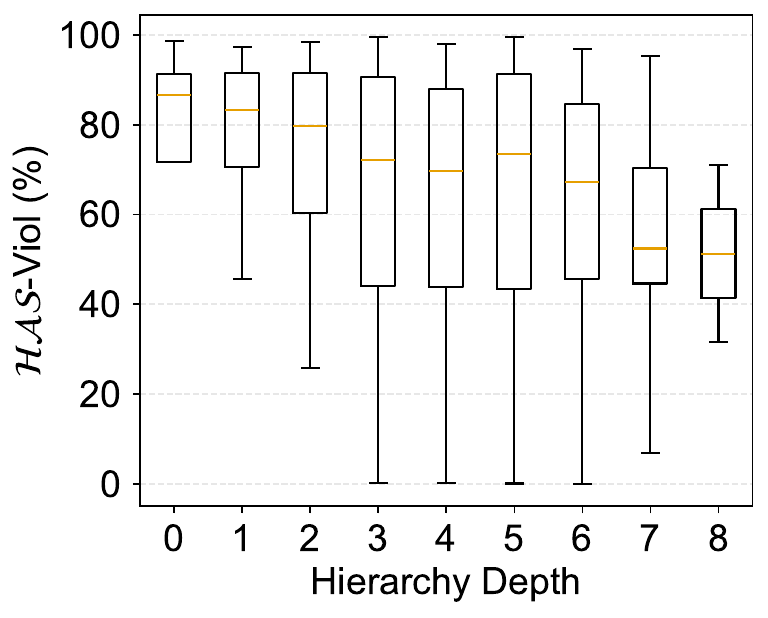}
  \caption{Randomly initialized.}
\end{subfigure}
\hfill
\begin{subfigure}[c]{0.31\linewidth}
  \centering
  \includegraphics[width=\linewidth]{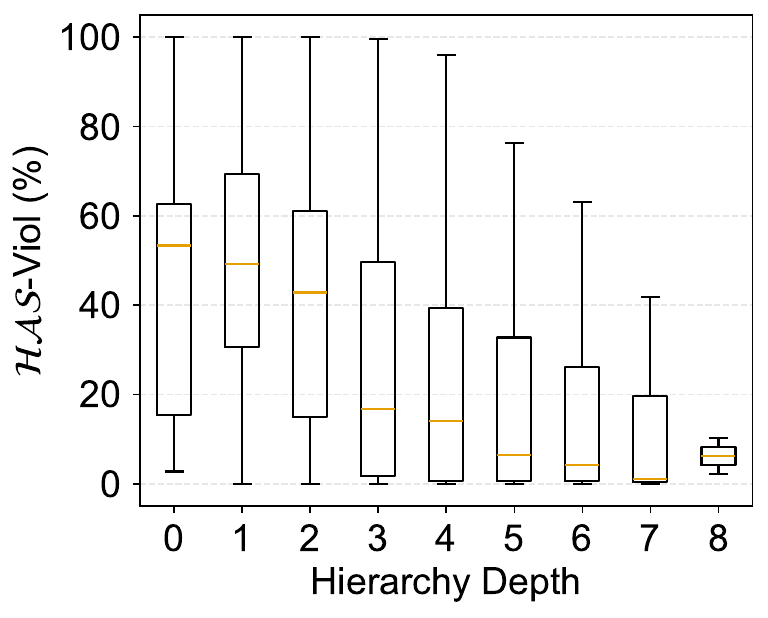}
  \caption{OpenAI.}
\end{subfigure}
\hfill
\begin{subfigure}[c]{0.31\linewidth}
  \centering
  \includegraphics[width=\linewidth]{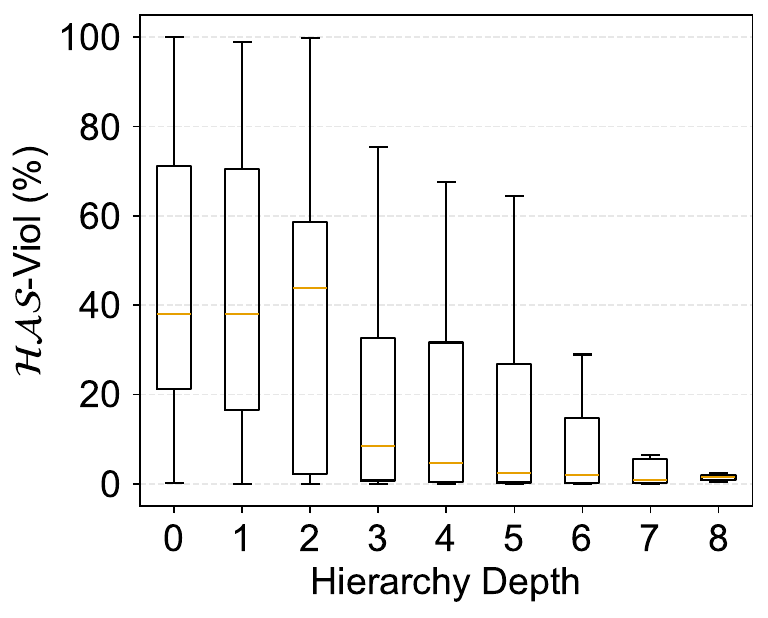}
  \caption{OpenCLIP (DataComp-1B).}
\end{subfigure}

\vspace{0.5em}

\begin{minipage}[c]{0.025\linewidth}
  \centering
  \raisebox{2.7\baselineskip}{\rotatebox{90}{\small Image}}
\end{minipage}%
\hfill
\begin{subfigure}[c]{0.31\linewidth}
  \centering
  \includegraphics[width=\linewidth]{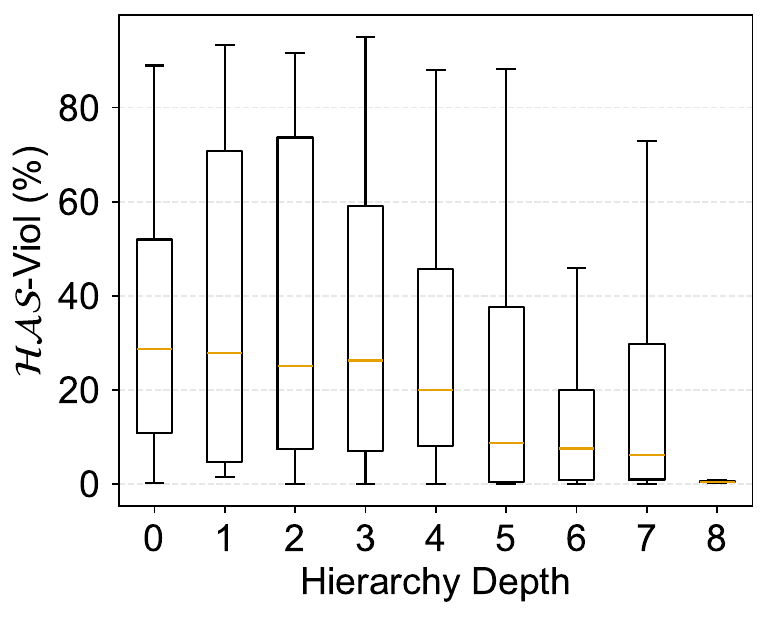}
  \caption{Randomly initialized.}
\end{subfigure}
\hfill
\begin{subfigure}[c]{0.31\linewidth}
  \centering
  \includegraphics[width=\linewidth]{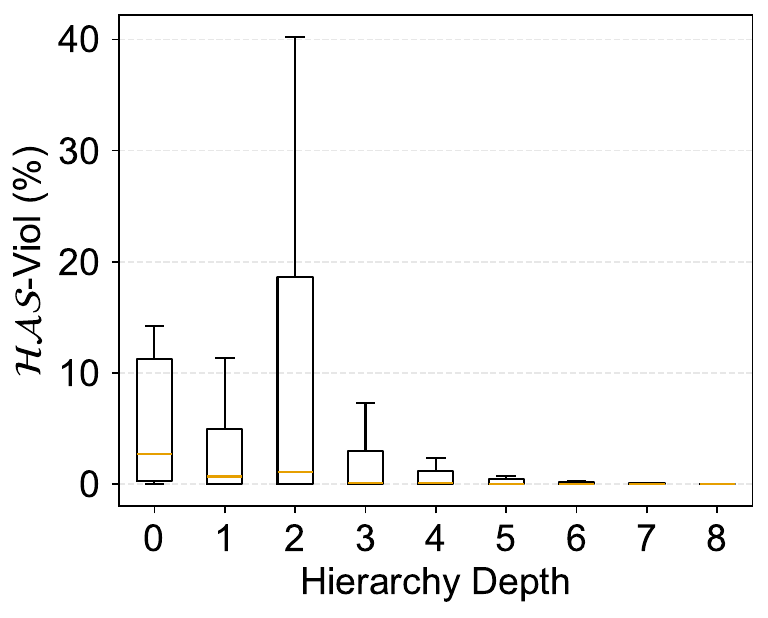}
  \caption{OpenAI.}
\end{subfigure}
\hfill
\begin{subfigure}[c]{0.31\linewidth}
  \centering
  \includegraphics[width=\linewidth]{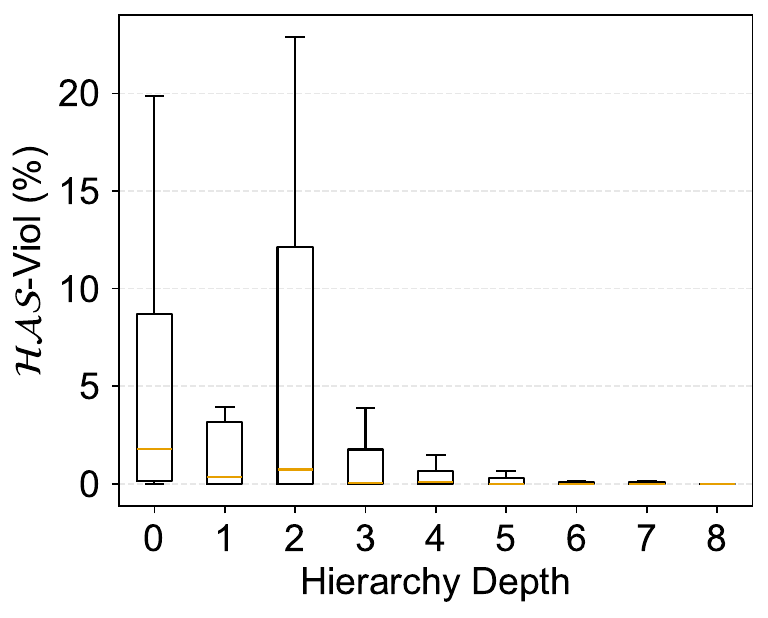}
  \caption{OpenCLIP (DataComp-1B).}
\end{subfigure}
\caption{\HAS violation percentage by depth on ViT-B/32 CLIP embeddings. \emph{Top:} \emph{text} embeddings (direct concept encoding): \HAS appears with pretraining. \emph{Bottom:} \emph{image} embeddings (instance aggregation): approximate \HAS appears at random initialization and strengthens after pretraining.}
\label{fig:has-violation-vitb32}
\end{figure}

\myparagraph{ViT-B/32 results}
We further repeat the experiment on ViT-B/32, comparing OpenAI CLIP and OpenCLIP (DataComp-1B) against a randomly initialized ViT-B/32. \Cref{fig:has-violation-vitb32} groups text embeddings (top; direct concept encoding) and image embeddings (bottom; instance aggregation) and reproduces the same qualitative pattern.

\begin{figure}[t]
\centering
\begin{minipage}[c]{0.025\linewidth}
  \centering
  \raisebox{2.7\baselineskip}{\rotatebox{90}{\small Text}}
\end{minipage}%
\hfill
\begin{subfigure}[c]{0.475\linewidth}
  \centering
  \includegraphics[width=\linewidth]{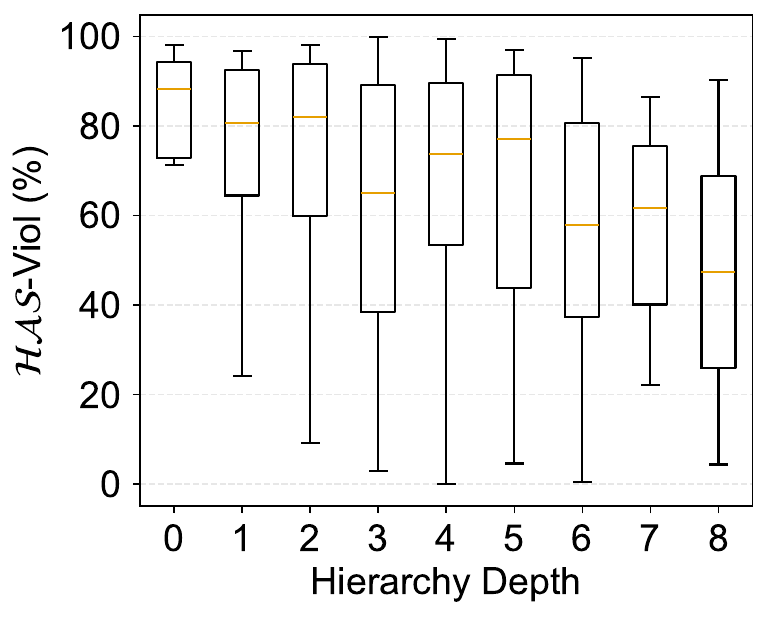}
  \caption{Randomly initialized.}
\end{subfigure}
\hfill
\begin{subfigure}[c]{0.475\linewidth}
  \centering
  \includegraphics[width=\linewidth]{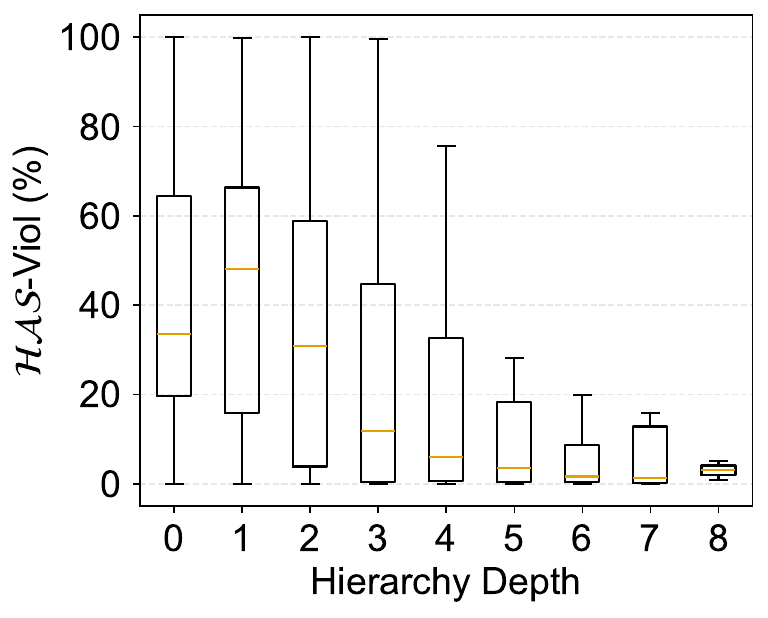}
  \caption{OpenAI.}
\end{subfigure}

\vspace{0.5em}

\begin{minipage}[c]{0.025\linewidth}
  \centering
  \raisebox{2.7\baselineskip}{\rotatebox{90}{\small Image}}
\end{minipage}%
\hfill
\begin{subfigure}[c]{0.475\linewidth}
  \centering
  \includegraphics[width=\linewidth]{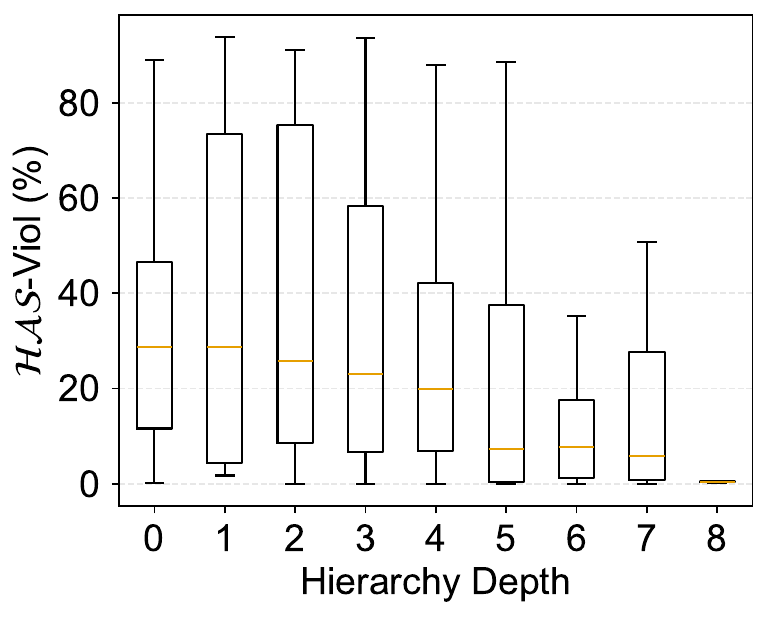}
  \caption{Randomly initialized.}
\end{subfigure}
\hfill
\begin{subfigure}[c]{0.475\linewidth}
  \centering
  \includegraphics[width=\linewidth]{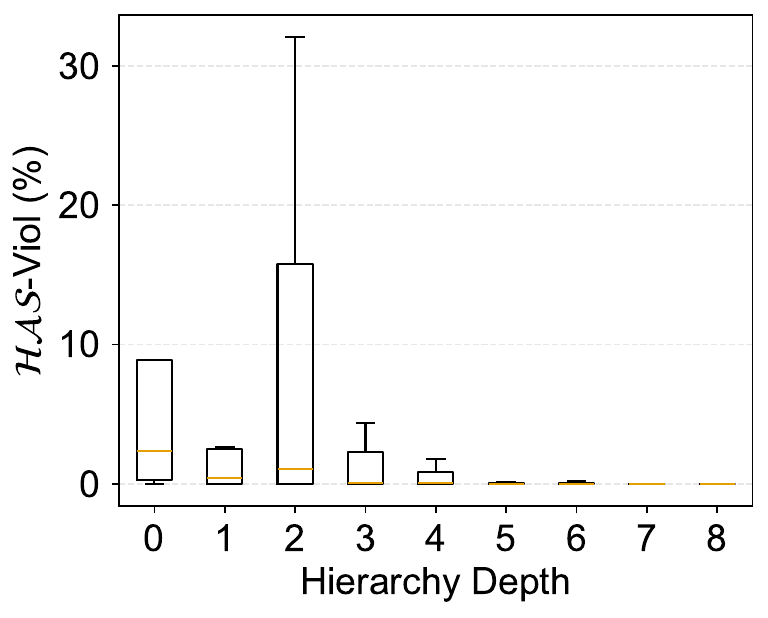}
  \caption{OpenAI.}
\end{subfigure}
\caption{\HAS violation percentage by depth on ViT-L/14@336px CLIP embeddings. \emph{Top:} \emph{text} embeddings (direct concept encoding): \HAS appears with pretraining. \emph{Bottom:} \emph{image} embeddings (instance aggregation): approximate \HAS appears at random initialization and strengthens after pretraining.}
\label{fig:has-violation-vitl336}
\end{figure}

\myparagraph{ViT-L/14@336px results}
We also evaluate the higher-resolution ViT-L/14@336px backbone, for which only the OpenAI checkpoint ships at 336px in OpenCLIP; we compare it against a randomly initialized ViT-L/14@336px. \Cref{fig:has-violation-vitl336} groups text embeddings (top; direct concept encoding) and image embeddings (bottom; instance aggregation) and reproduces the same qualitative pattern.

\begin{figure}[t]
\centering
\begin{minipage}[c]{0.025\linewidth}
  \centering
  \raisebox{2.7\baselineskip}{\rotatebox{90}{\small Text}}
\end{minipage}%
\hfill
\begin{subfigure}[c]{0.475\linewidth}
  \centering
  \includegraphics[width=\linewidth]{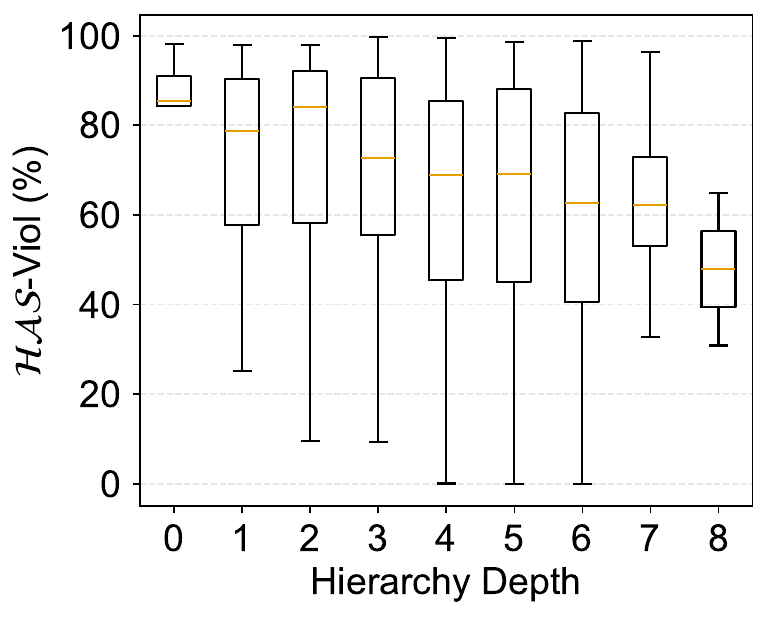}
  \caption{Randomly initialized.}
\end{subfigure}
\hfill
\begin{subfigure}[c]{0.475\linewidth}
  \centering
  \includegraphics[width=\linewidth]{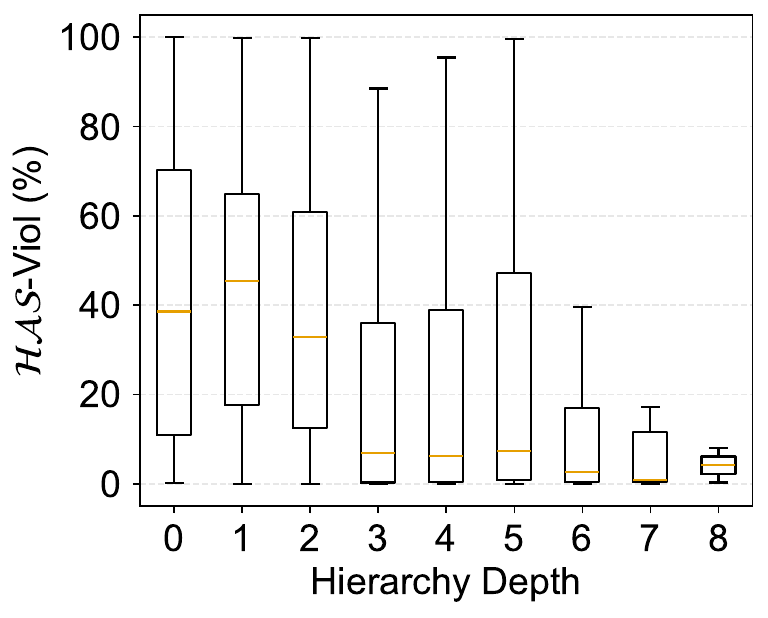}
  \caption{SigLIP2 pretrained.}
\end{subfigure}

\vspace{0.5em}

\begin{minipage}[c]{0.025\linewidth}
  \centering
  \raisebox{2.7\baselineskip}{\rotatebox{90}{\small Image}}
\end{minipage}%
\hfill
\begin{subfigure}[c]{0.475\linewidth}
  \centering
  \includegraphics[width=\linewidth]{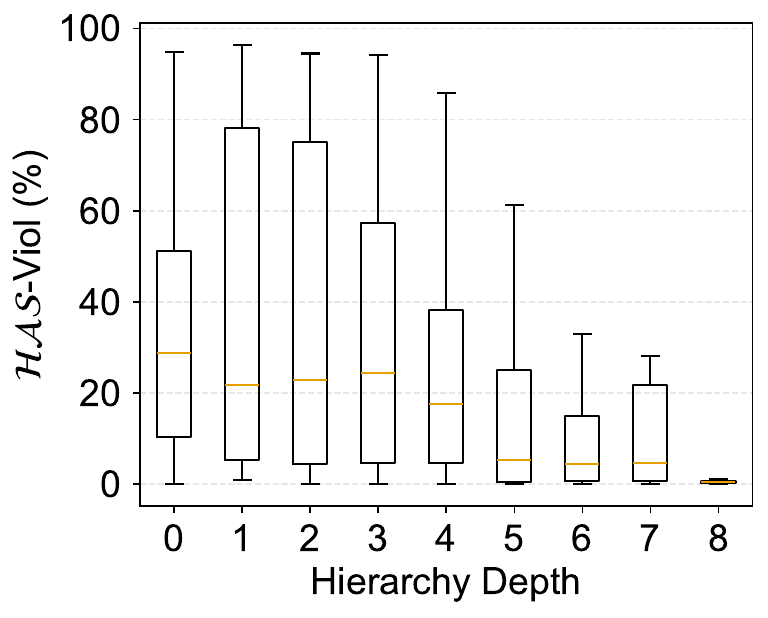}
  \caption{Randomly initialized.}
\end{subfigure}
\hfill
\begin{subfigure}[c]{0.475\linewidth}
  \centering
  \includegraphics[width=\linewidth]{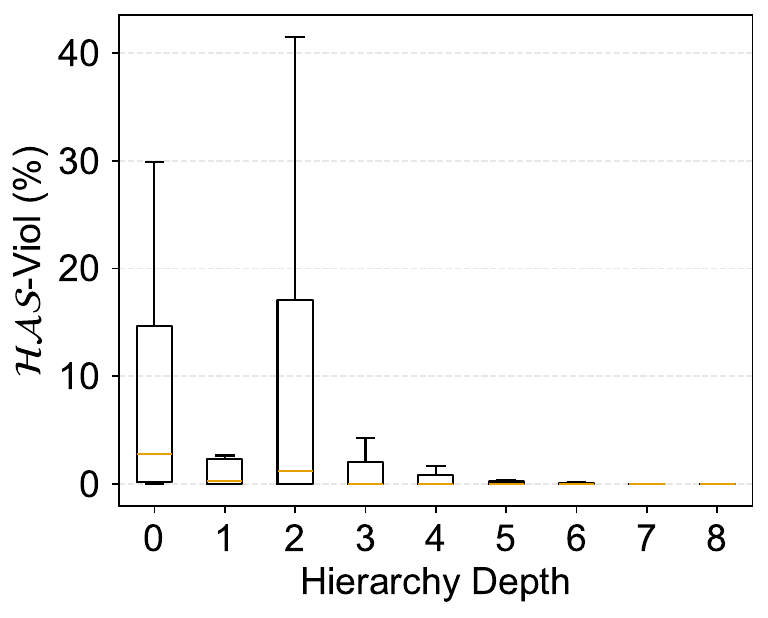}
  \caption{SigLIP2 pretrained.}
\end{subfigure}
\caption{\HAS violation percentage by depth on SigLIP2 embeddings. \emph{Top:} \emph{text} embeddings (direct concept encoding): \HAS appears with pretraining. \emph{Bottom:} \emph{image} embeddings (instance aggregation): approximate \HAS appears at random initialization and strengthens after pretraining.}
\label{fig:has-violation-siglip2}
\end{figure}

\myparagraph{SigLIP2 results}
We evaluate SigLIP2 (ViT-B/16 at 256px), comparing the pretrained SigLIP2 checkpoint against a randomly initialized ViT-B/16 at 256px. \Cref{fig:has-violation-siglip2} (text via direct concept encoding on top, image via instance aggregation on bottom) reproduces the same qualitative pattern: text \HAS appears with pretraining, while image embeddings show approximate \HAS at random initialization.

\begin{figure}[t]
\centering
\begin{minipage}[c]{0.025\linewidth}
  \centering
  \raisebox{2.7\baselineskip}{\rotatebox{90}{\small Text}}
\end{minipage}%
\hfill
\begin{subfigure}[c]{0.475\linewidth}
  \centering
  \includegraphics[width=\linewidth]{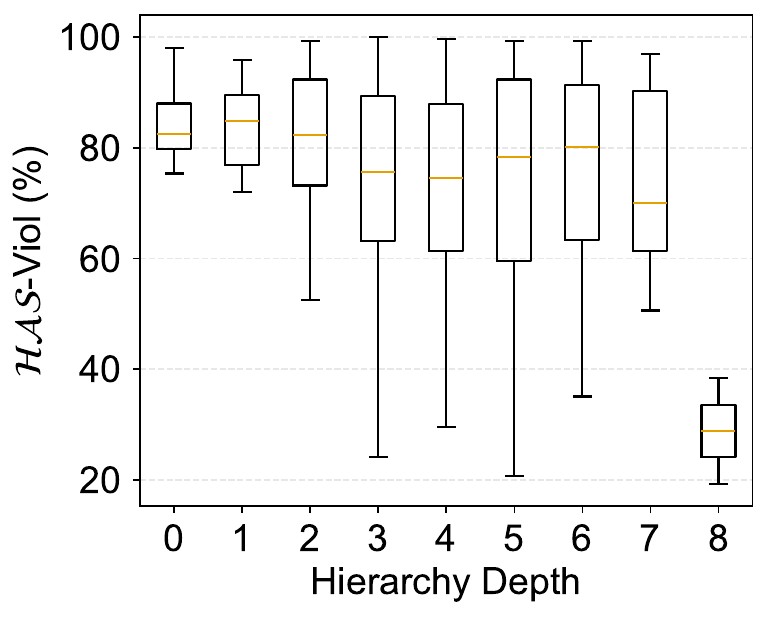}
  \caption{Randomly initialized.}
\end{subfigure}
\hfill
\begin{subfigure}[c]{0.475\linewidth}
  \centering
  \includegraphics[width=\linewidth]{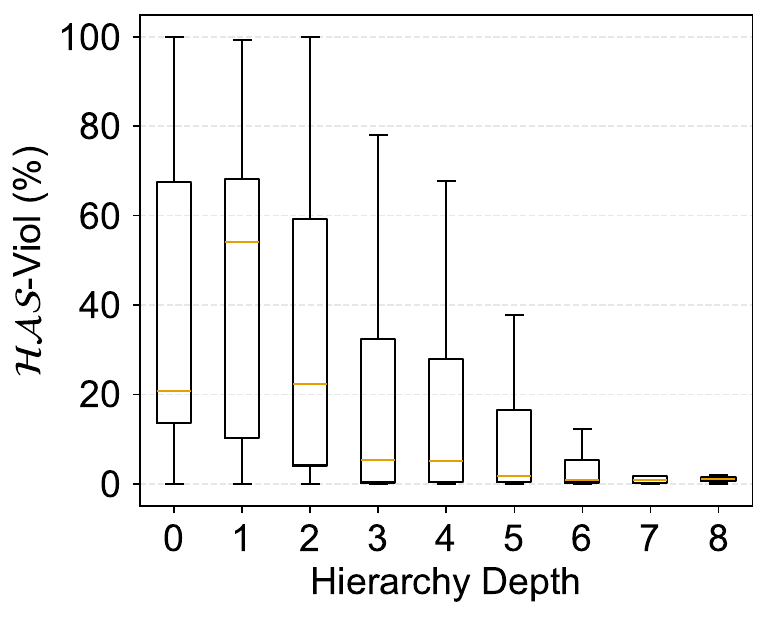}
  \caption{MetaCLIP2 pretrained.}
\end{subfigure}

\vspace{0.5em}

\begin{minipage}[c]{0.025\linewidth}
  \centering
  \raisebox{2.7\baselineskip}{\rotatebox{90}{\small Image}}
\end{minipage}%
\hfill
\begin{subfigure}[c]{0.475\linewidth}
  \centering
  \includegraphics[width=\linewidth]{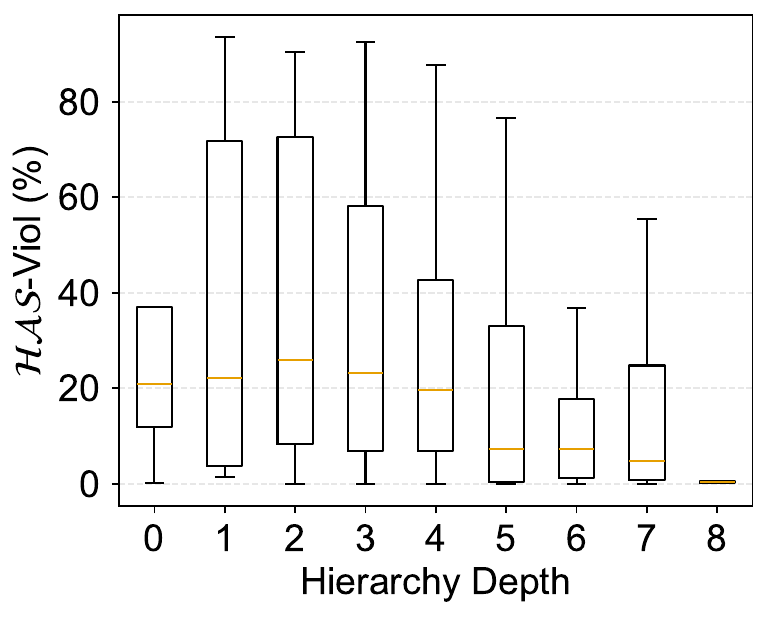}
  \caption{Randomly initialized.}
\end{subfigure}
\hfill
\begin{subfigure}[c]{0.475\linewidth}
  \centering
  \includegraphics[width=\linewidth]{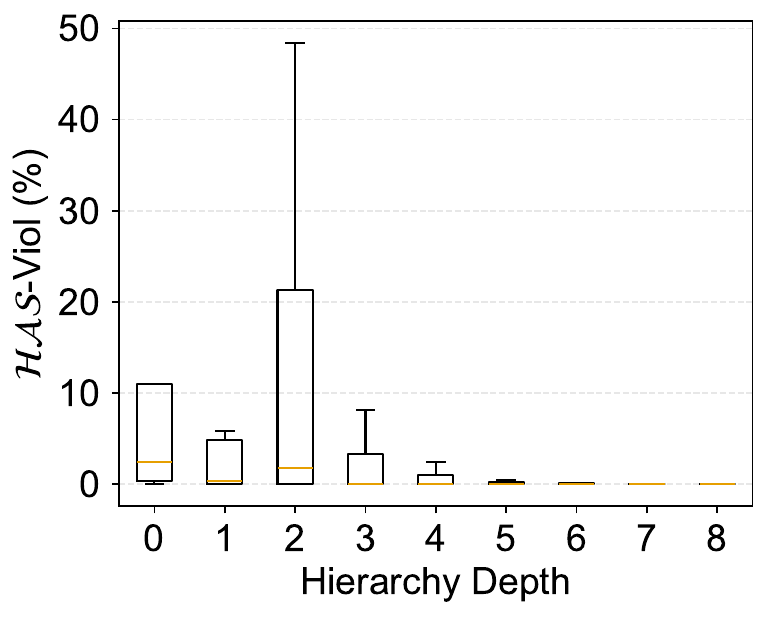}
  \caption{MetaCLIP2 pretrained.}
\end{subfigure}
\caption{\HAS violation percentage by depth on MetaCLIP2 embeddings. \emph{Top:} \emph{text} embeddings (direct concept encoding): \HAS appears with pretraining. \emph{Bottom:} \emph{image} embeddings (instance aggregation): approximate \HAS appears at random initialization and strengthens after pretraining.}
\label{fig:has-violation-metaclip2}
\end{figure}

\myparagraph{MetaCLIP2 results}
We evaluate MetaCLIP2 (ViT-H/14 at 224px), comparing the pretrained MetaCLIP2 worldwide checkpoint against a randomly initialized ViT-H/14. \Cref{fig:has-violation-metaclip2} (text via direct concept encoding on top, image via instance aggregation on bottom) reproduces the same qualitative pattern: text \HAS appears with pretraining, while image embeddings show approximate \HAS at random initialization.

\subsection{Disjoint-Split Control against Instance Overlap for \HAS in Image Embeddings}\label{app:image-has-disjoint-split-control}

In hierarchical instance aggregation, ancestor prototypes naturally share image instances with their descendants, creating an inherent source of correlated sampling noise.
When prototypes are estimated by averaging instance vectors over descendant leaves, every image contributing to a descendant node also enters the sample mean of all its ancestors.
This shared sample overlap inflates anchor--descendant cosine similarities and narrows the apparent descendant cone, potentially producing low \HAS violation rates even in representations lacking semantic hierarchy.
Because standard hierarchy shuffling permutes leaf assignments across the tree while preserving this internal parent--descendant overlap, a shuffled null alone cannot detect this artifact.

To eliminate sample overlap while preserving tree aggregation, we evaluate \HAS using a split-sample protocol.
For each ImageNet training leaf, we partition the available images into even and odd splits, rebuilding all internal node prototypes independently within each split.
We then evaluate anchor prototypes from one split against descendant and off-branch prototypes from the other split, and average \HASViol across both split directions.
Under this procedure, every anchor prototype and its comparison targets are estimated from strictly disjoint sets of images.
We employ this disjoint-split protocol for every representation and its matched hierarchy-shuffled null in \Cref{tab:raw-pixel-has}.

To measure the precise impact of instance overlap, we conduct a paired experiment comparing shared-instance and disjoint-split evaluations alongside an uninformative Gaussian baseline.
In \Cref{tab:image-has-instance-overlap}, we evaluate raw pixels, randomly initialized ViT-L/14, pretrained CLIP, and isotropic Gaussian noise vectors ($\bm{z}_j \sim \cN(\bzero, \bI_d)$, normalized to unit length).
Because Gaussian vectors are distributed independently of the semantic taxonomy, any apparent separation under shared instances isolates the purely geometric artifact of sample reuse, while their behavior under disjoint splits establishes the uninformative chance level.
For random ViT and Gaussian vectors, reported values and spreads indicate means and sample standard deviations across three weight initializations and $40$ independent Gaussian draws, respectively; raw pixels and pretrained CLIP report single evaluations paired with five hierarchy permutations.

\begin{table}[H]
\centering
\small
\setlength{\tabcolsep}{5pt}
\caption{Mean image-side \HASViol (\%) under shared-instance versus disjoint-split instance aggregation, paired with matched hierarchy-shuffled nulls. Spreads ($\pm$) denote sample standard deviations across random seeds or permutations (details in text). Disjoint splitting eliminates the apparent separation of the Gaussian noise baseline while preserving the hierarchical geometry of genuine visual representations.}
\label{tab:image-has-instance-overlap}
\begin{tabular}{lcccc}
\toprule
& \multicolumn{2}{c}{Shared instances} & \multicolumn{2}{c}{Disjoint splits} \\
\cmidrule(lr){2-3} \cmidrule(lr){4-5}
Representation & \shortstack{Original leaf\\assignments} & \shortstack{Shuffled leaf\\assignments} & \shortstack{Original leaf\\assignments} & \shortstack{Shuffled leaf\\assignments} \\
\midrule
Raw pixels & $25.75$ & $49.87 \pm 0.38$ & $26.59$ & $51.06 \pm 0.41$ \\
Random ViT-L/14 & $27.17 \pm 0.19$ & $56.04 \pm 0.12$ & $28.03 \pm 0.23$ & $56.95 \pm 0.13$ \\
OpenAI CLIP ViT-L/14 & $2.76$ & $32.81 \pm 0.56$ & $2.76$ & $32.87 \pm 0.56$ \\
i.i.d. Gaussian ($d=768$) & $1.19 \pm 0.07$ & $1.19 \pm 0.04$ & $76.02 \pm 1.03$ & $76.24 \pm 0.59$ \\
\bottomrule
\end{tabular}
\end{table}

The paired results in \Cref{tab:image-has-instance-overlap} demonstrate that instance overlap artificially induces apparent separation for unstructured noise, but does not account for the hierarchical structure observed in visual representations.
For isotropic Gaussian vectors, shared instance aggregation produces an artificially low violation rate ($1.19\%$), which is indistinguishable from its matched shuffled null ($1.19\%$).
Under disjoint splits, where sample overlap is eliminated, Gaussian violations jump to chance level ($76.02\%$), again matching the shuffled baseline ($76.24\%$) and unmasking the shared-instance artifact.
In sharp contrast, all three visual representations (raw pixels, randomly initialized ViT-L/14, and pretrained CLIP) remain largely unaffected by disjoint splitting, maintaining violation rates far below their respective shuffled nulls under both protocols.

\begin{figure}[t]
\centering
\begin{minipage}[c]{0.025\linewidth}
  \centering
  \raisebox{2.7\baselineskip}{\rotatebox{90}{\small Disjoint}}
\end{minipage}%
\hfill
\begin{subfigure}[c]{0.31\linewidth}
  \centering
  \includegraphics[width=\linewidth]{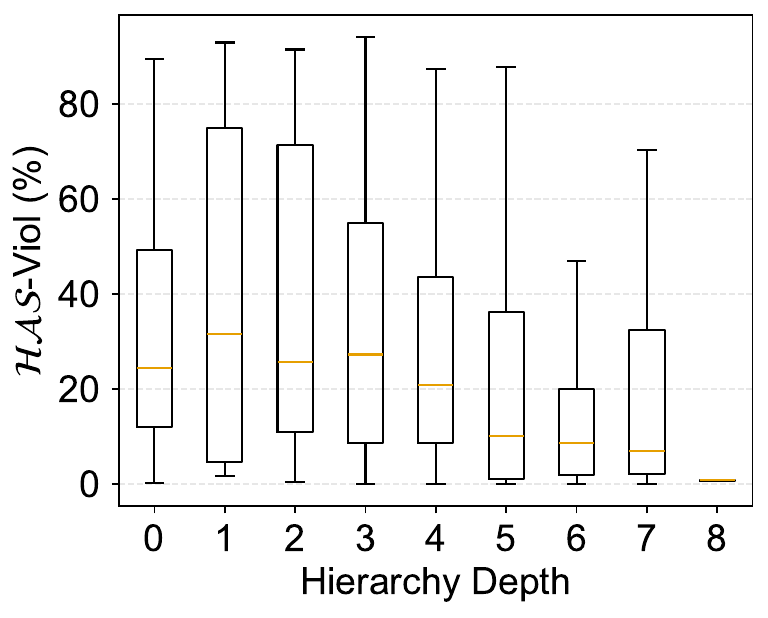}
  \caption{Randomly initialized.}
\end{subfigure}
\hfill
\begin{subfigure}[c]{0.31\linewidth}
  \centering
  \includegraphics[width=\linewidth]{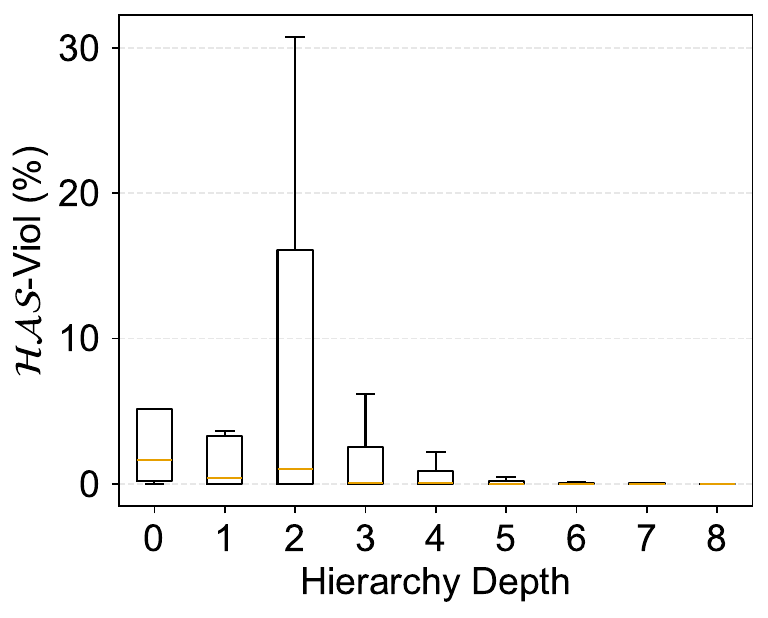}
  \caption{OpenAI.}
\end{subfigure}
\hfill
\begin{subfigure}[c]{0.31\linewidth}
  \centering
  \includegraphics[width=\linewidth]{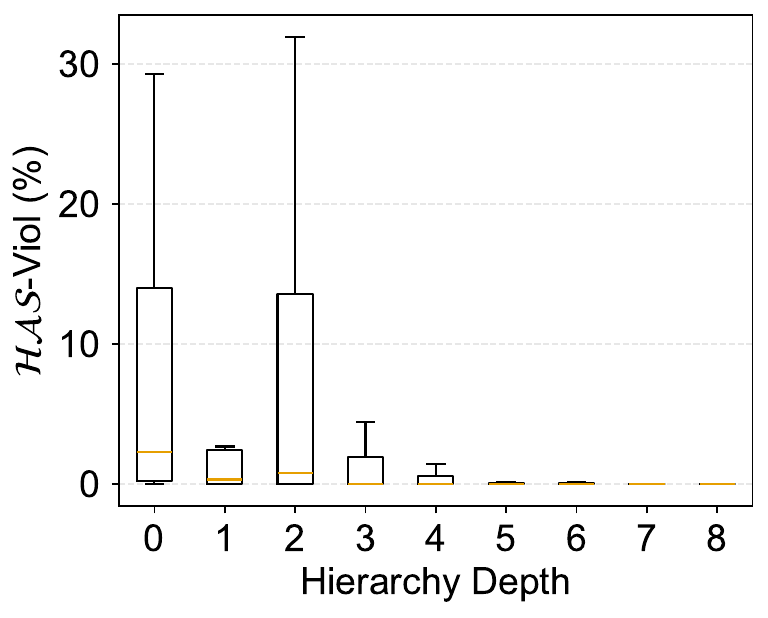}
  \caption{OpenCLIP (DataComp-1B).}
\end{subfigure}

\vspace{0.5em}

\begin{minipage}[c]{0.025\linewidth}
  \centering
  \raisebox{2.7\baselineskip}{\rotatebox{90}{\small Original}}
\end{minipage}%
\hfill
\begin{subfigure}[c]{0.31\linewidth}
  \centering
  \includegraphics[width=\linewidth]{figures/experiments/hierarchical-cone-separation/image/ViT-L/descendant_leaves/random_vit_violation_by_depth.pdf}
  \caption{Randomly initialized.}
\end{subfigure}
\hfill
\begin{subfigure}[c]{0.31\linewidth}
  \centering
  \includegraphics[width=\linewidth]{figures/experiments/hierarchical-cone-separation/image/ViT-L/descendant_leaves/openai_violation_by_depth.pdf}
  \caption{OpenAI.}
\end{subfigure}
\hfill
\begin{subfigure}[c]{0.31\linewidth}
  \centering
  \includegraphics[width=\linewidth]{figures/experiments/hierarchical-cone-separation/image/ViT-L/descendant_leaves/datacomp1b_violation_by_depth.pdf}
  \caption{OpenCLIP (DataComp-1B).}
\end{subfigure}
\caption{\HAS violation percentages by depth on ViT-L/14 CLIP \emph{image} embeddings remain similar under the disjoint-split control. \emph{Top:} disjoint-split control, where parent prototypes and comparison prototypes are estimated from disjoint image instances. \emph{Bottom:} original instance-aggregation evaluation.}
\label{fig:image-has-disjoint-split-control}
\end{figure}

This geometric stability holds across all hierarchy levels, ruling out sample overlap as the source of approximate image-side \HAS.
In \Cref{fig:image-has-disjoint-split-control}, the depth-resolved violation profiles for ViT-L/14 under disjoint splits mirror those under shared-instance aggregation across all depths.
While the disjoint-split protocol eliminates sample sharing between anchor and comparison prototypes, parent prototypes remain averages over descendant leaf distributions.
The measured separation therefore reflects genuine consistency across visual subtrees and the hierarchical structure of the data distribution, rather than a statistical artifact of instance reuse.

\clearpage

\subsection{Numerical Validation of \HAS Transfer at Small Regularization}\label{sec:exp-low-lambda}\label{app:exp-low-lambda-setup}

In \Cref{thm:large-lambda-has-noisy}, we guarantee approximate \HAS transfer only for sufficiently large $\lambda$, but leave open whether transfer persists outside the asymptotic regime used in its proof. We test this possibility on a synthetic image-prototype hierarchy by minimizing \Cref{eq:objective-simplified-finite} across a sweep of $\lambda$. We verify whether transfer happens outside the large-$\lambda$ regime, whether the optimal text embeddings approach the root-centered prototype decomposition of \Cref{cor:prototype-form} as $\lambda$ increases, and whether the sufficient transfer condition is conservative. We report \HASViol at the minimizer found by L-BFGS and define prototype alignment as the minimum, over non-root nodes $k$, of the cosine between $\ba_k^{\mathrm{nce},*}$ and $\bm{\mu}_k$.

\myparagraph{Experimental setup}
For convenience, we restate the data generating process from \Cref{ass:noisy-prototype} as
\begin{equation}\label{eq:has-data-generating-process}
  \bz_j = \bm{\mu}_{y_j} + \bm{\varepsilon}_j,
  \qquad \E[\bm{\varepsilon}_j]=\bzero,\ \ j\in[N].
\end{equation}
We instantiate \Cref{eq:has-data-generating-process} with isotropic Gaussian variation $\bm{\varepsilon}_j\sim\mathcal{N}(\bzero,\sigma^2\bI_d)$ at depth $L=5$, branching factor $3$, variation parameter $\sigma=10^{-5}$, feature-dimension-to-number-of-nodes ratio $d/\abs{V}=1$, and $150$ images per leaf, using prototypes generated by geometric-decay sibling angles ($\theta_0=85^{\circ}$). We optimize the symmetric InfoNCE objective (temperature $1$, no learnable temperature) with L-BFGS to find the global minimizer. We assign captions deterministically, cycling each leaf's images through its root-to-leaf path; we call this the balanced-caption setup.

\myparagraph{Transfer precedes the asymptotic regime} Sweeping $\lambda\in[10^{-5},10^{2}]$ separates the onset of transfer from the accuracy of the large-$\lambda$ expansion. \HASViol reaches zero before prototype alignment saturates (\Cref{fig:synthetic-lambda-sweep}). The convergence trace confirms that zero \HASViol holds at the minimizer found by L-BFGS rather than only at an early iterate (\Cref{fig:synthetic-convergence-trace}).

\myparagraph{The large-$\lambda$ expansion predicts the limiting geometry} Prototype alignment increases with $\lambda$ and approaches one in the large-$\lambda$ regime (\Cref{fig:synthetic-lambda-sweep}). The trend is consistent with the directional prediction of \Cref{cor:prototype-form}, while the earlier disappearance of violations shows that transfer does not require near-perfect prototype alignment.

\begin{figure}[H]
\centering
\includegraphics[width=0.9\linewidth]{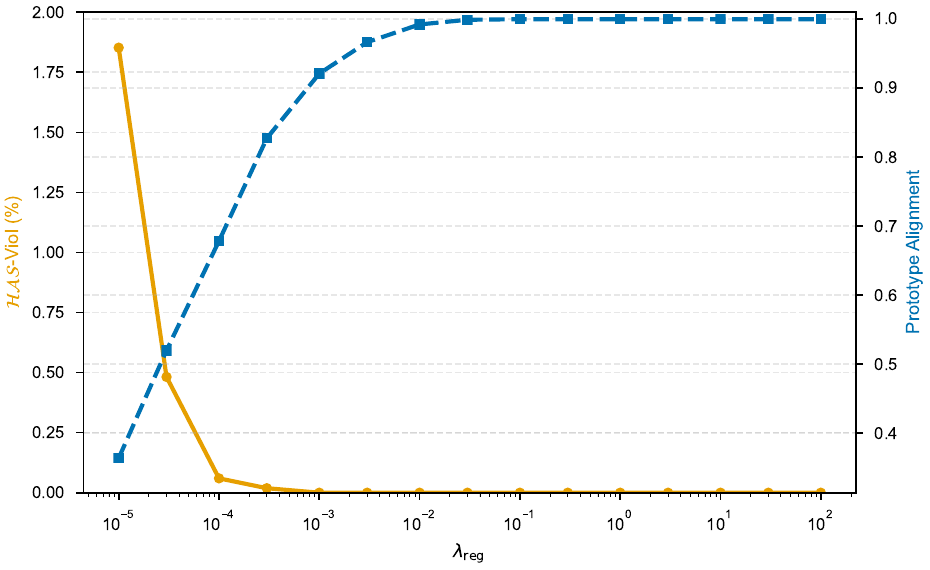}
\caption{$\lambda$ sweep at root norm $\|\bm{\mu}_{\mathrm{root}}\|=10^{-2}$ in the same balanced-caption setup as \Cref{fig:synthetic-lambda-sweep}. The x-axis is $\lambda$; the left y-axis is the final \HASViol averaged over non-leaf nodes, and the right y-axis is the final minimum prototype alignment over non-root nodes. Although this root norm violates the sufficient transfer condition $D_\epsilon^{\mathrm{nce}}<\rho_{\min}\gamma_{\HAS}/16$ by a factor of $161$ on this set of prototypes, because $D_\epsilon^{\mathrm{nce}}\ge\xi_{\max}\|\bm{\mu}_{\mathrm{root}}\|$ and $\xi_{\max}=2.50$, the drop in violation percentage is nearly unchanged: it again reaches zero by $\lambda=10^{-3}$, while prototype alignment saturates just below $1$, reflecting the remaining root prototype $\bm{\mu}_{\mathrm{root}}$.}
\label{fig:issue40-noncentered-root-outofbudget}
\end{figure}

\begin{figure}[H]
\centering
\includegraphics[width=0.9\linewidth]{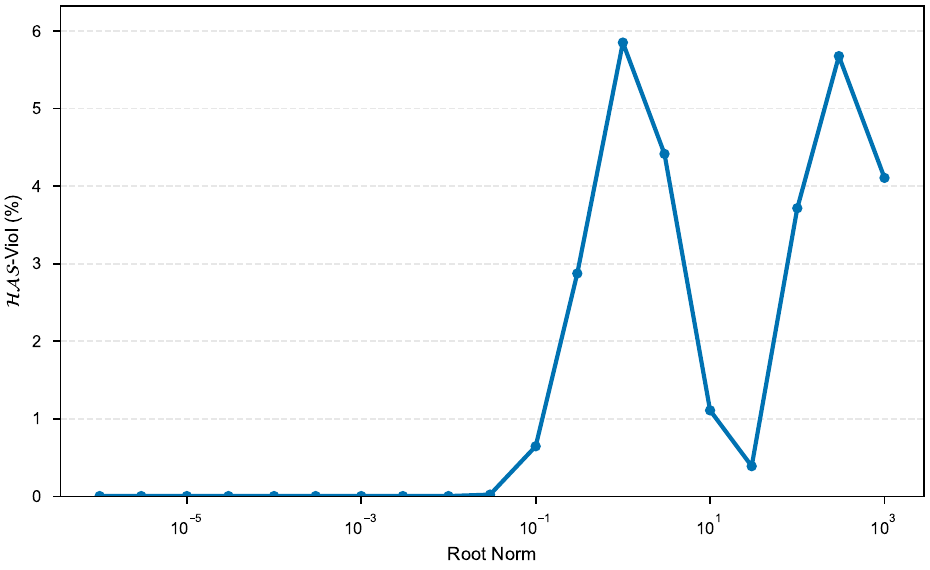}
\caption{Root-norm sweep at $\lambda=10^{-3}$ in the same balanced-caption setup as \Cref{sec:exp-low-lambda}. The x-axis is the root norm $\|\bm{\mu}_{\mathrm{root}}\|$, and the y-axis is the final \HASViol averaged over non-leaf nodes. The violation percentage remains nearly flat at small root norm, then departs strongly from near-zero once the root norm is large, with non-monotone behavior in the extreme large-root-norm regime.}
\label{fig:issue40-rootnorm-fixed-lambda}
\end{figure}

\begin{figure}[H]
\centering
\begin{subfigure}[t]{0.32\linewidth}
  \centering
  \includegraphics[width=\linewidth]{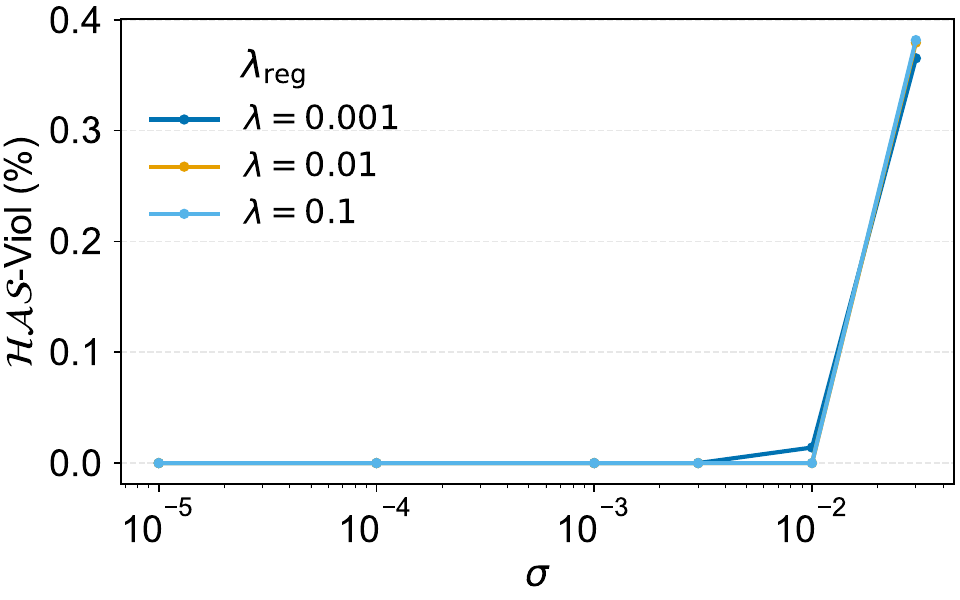}
\end{subfigure}
\hfill
\begin{subfigure}[t]{0.32\linewidth}
  \centering
  \includegraphics[width=\linewidth]{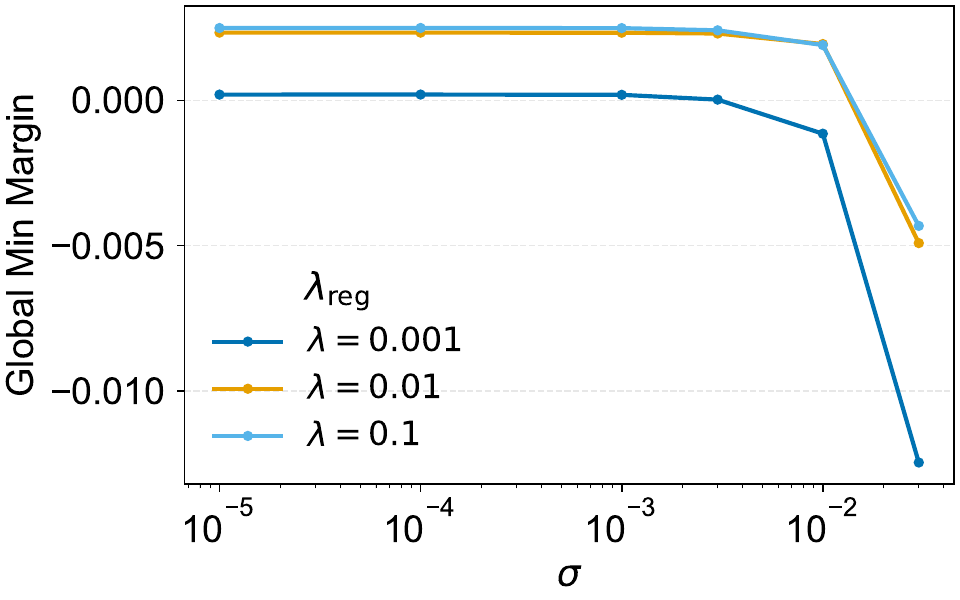}
\end{subfigure}
\hfill
\begin{subfigure}[t]{0.32\linewidth}
  \centering
  \includegraphics[width=\linewidth]{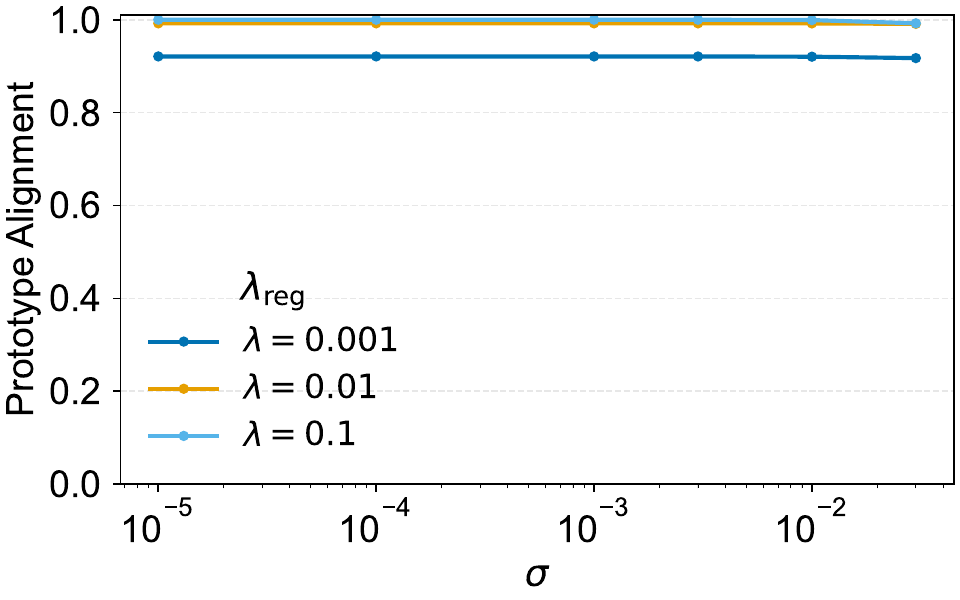}
\end{subfigure}
\caption{$\sigma$-robustness sweep in the same balanced-caption setup as \Cref{sec:exp-low-lambda}, with root norm $\|\bm{\mu}_{\mathrm{root}}\|=10^{-4}$ and $\lambda\in\{10^{-3},10^{-2},10^{-1}\}$. \textit{Left:} \HASViol. \textit{Middle:} global minimum branch margin. \textit{Right:} minimum non-root prototype alignment. All three regularization levels remain stable through $\sigma=3\!\times\!10^{-3}$. At $\sigma=10^{-2}$, only $\lambda=10^{-3}$ has a nonzero violation percentage ($0.014\%$) and a negative global minimum margin; the other two settings retain zero violations and positive margins. At $\sigma=3\!\times\!10^{-2}$, every $\lambda$ has a negative global minimum margin and a violation percentage between $0.37\%$ and $0.38\%$.}
\label{fig:issue40-sigma-robustness}
\end{figure}

\myparagraph{The sufficient transfer condition is conservative} Under \Cref{thm:large-lambda-has-noisy}, transfer is certified when $D_\epsilon^{\mathrm{nce}}<\rho_{\min}\gamma_{\HAS}/16$. Both tested root norms, $\|\bm{\mu}_{\mathrm{root}}\|\in\{10^{-4},10^{-2}\}$, violate this condition yet produce nearly the same transition (\Cref{fig:synthetic-lambda-sweep,fig:issue40-noncentered-root-outofbudget}). At fixed $\lambda=10^{-3}$, violations remain small across the root-norm sweep and increase only under substantially larger shifts (\Cref{fig:issue40-rootnorm-fixed-lambda}). Across the noise sweep at $\lambda\in\{10^{-3},10^{-2},10^{-1}\}$, degradation appears first under the weakest regularization and later under all three regularization values (\Cref{fig:issue40-sigma-robustness}). We therefore find that the theorem's transfer condition is sufficient but not necessary in this synthetic setting.

\subsection{Predicted and Realized Slack on Real ViT-L/14 Prototypes}\label{app:per-triplet-empirical}

We use these experiments to compare the guarantee of \Cref{thm:linear-sampled-path-noisy-has} with the behavior of its exact linear-loss minimizer on real prototypes. We first evaluate the theorem's threshold $\tau_\epsilon^{\mathrm{lin}}=8D_\epsilon^{\mathrm{lin}}/\rho_{\min}$, then measure the realized transfer and explain the gap between them. Although the theorem's bound gives no information in this setting, approximate \HAS{} still transfers.

We use the ImageNet-1K / WordNet image prototypes of \Cref{sec:hierarchical-angular-separation-empirical} ($1375$ nodes and $N=1{,}281{,}167$ training images) from three ViT-L/14 backbones: randomly initialized, OpenAI CLIP, and OpenCLIP trained on DataComp-1B. Holding the image prototypes fixed at $\bZ$, we compute the closed-form text-side minimizer $\bA^{\mathrm{lin},*}$ of \Cref{prop:linear-sampled-path-minimizer}. Each leaf prototype is its exact empirical image mean: we neither inject nor fit noise. The calculation in this subsection is therefore deterministic given the prototypes and caption counts.

\myparagraph{The inequality required by the theorem does not hold}
For this calculation, the reference vector is the empirical prototype itself, $\bm{\mu}_k=\bar{\bz}_k$, taken as the descendant-leaf-weighted instance aggregation of \Cref{sec:hierarchical-angular-separation-empirical}. The stochastic part of the perturbation therefore vanishes, leaving $\be_k=-\xi_k^{\mathrm{lin}}\bar{\bz}$. The theorem's error term consequently reduces to the root-centering contribution alone, $D=\xi_{\max}^{\mathrm{lin}}\|\bar{\bz}\|$, where $\xi_k^{\mathrm{lin}}\in[0.50,8.54]$ is fixed by the ImageNet caption counts and shared across encoders.

Using expected counts has a negligible effect on the prototypes or violation percentages. We use $N_k=\sum_{\ell}n_\ell/|\mathrm{path}(\ell)|$ under \Cref{ass:leaf-uniform-caption} rather than a sampled caption assignment, which makes the calculation deterministic. Under that assumption the empirical mean $\bar{\bz}_k$ of the images captioned $k$ weights each descendant leaf $\ell$ by $n_\ell/|\mathrm{path}(\ell)|$ instead of $n_\ell$. When we rebuild every internal prototype under that weighting, the prototypes and the text-side violation percentages of \Cref{tab:caption-permutation-real} remain essentially unchanged.

For every backbone, the inequality $D<\rho_{\min}/4$ required by \Cref{thm:linear-sampled-path-noisy-has} does not hold, so we obtain no guarantee from the theorem for these prototypes. Moreover, the resulting threshold exceeds the largest possible cosine-margin difference, which is $2$, so the bound does not certify any triplet. The mismatch is structural: the stochastic term is zero here, and the universal constant $C$ multiplies only that term.

\myparagraph{The sufficient bound is loose, but approximate \HAS{} still transfers}
We therefore measure transfer descriptively rather than certify it. For each anchor, we find the smallest slack that satisfies \Cref{eq:per-triplet-has-bound}, then report the largest of these anchor-specific values. This worst-anchor slack is the single post hoc slack needed for the inequality to hold at every anchor.

The quantity $\HASViol^{\mathrm{img}}_{\tau}(u)$ is \Cref{def:violation-percentage} evaluated on the image prototypes with the descendant radius $s^{\bm{\mu}}_u$ relaxed by $\tau$. The theoretical bound and the empirical metric of \Cref{sec:hierarchical-angular-separation-empirical} therefore have the same units: a percentage of off-branch nodes per anchor. The same bound holds for the percentile-based descendant radius used there for outlier robustness, provided both sides replace the minimum by the same order statistic, the $r$-th smallest descendant cosine for a fixed $r$. If $w$ is counted on the text side, then at least $r$ descendants $v$ satisfy $\cos(\widetilde{\ba}_u,\widetilde{\ba}_v)\le\cos(\widetilde{\ba}_u,\widetilde{\ba}_w)$. By part (ii) of \Cref{lem:per-triplet-margin-preservation}, $\gamma_{uvw}\le\tau$ for each of them, so at least $r$ descendants satisfy $\cos(\bm{\mu}_u,\bm{\mu}_v)\le\cos(\bm{\mu}_u,\bm{\mu}_w)+\tau$, and $w$ is counted on the image side.

As in \Cref{sec:hierarchical-angular-separation-empirical}, we compute the $95$th percentile of the descendant angles with linear interpolation rather than as an exact order statistic. Both sides have the same rank, but neither radius is one of the measured cosines, so the order-statistic argument applies only approximately. When we recompute both sides with the order statistic, every conclusion in this appendix remains intact.

To assess whether individual image-side violations persist after transfer, we pool the $505{,}938$ (anchor, off-branch node) pairs scored by \Cref{def:violation-percentage} over the $375$ anchors. We compare the probability that such a pair violates \HAS in the learned text embeddings with the conditional probability that it violates \HAS on the text side given that it violates \HAS on the image side.

\begin{table}[t]
\centering
\caption{The inequality required by \Cref{thm:linear-sampled-path-noisy-has} and measured approximate \HAS{} transfer on the same real prototypes. The inequality $D<\rho_{\min}/4$ does not hold for any of the three backbones, so we obtain no guarantee from the theorem here. We evaluate the image-side rows on the fixed prototypes and the text-side rows on the closed-form linear-loss optimum $\bA^{\mathrm{lin},*}$. For the worst-anchor slack, one value of $\tau$ must satisfy \Cref{eq:per-triplet-has-bound} at every anchor, so we report the largest anchor-specific slack. This slack is post hoc. For the last three rows, we pool the $505{,}938$ (anchor, off-branch node) pairs over the $375$ anchors. For the two $\HASViol$ rows, we report per-anchor percentages measured from all training images per node in one pass, not with the disjoint even/odd split of \Cref{tab:raw-pixel-has}.}
\label{tab:per-triplet-slack}
\resizebox{\linewidth}{!}{%
\begin{tabular}{lccc}
\toprule
Quantity & Randomly initialized ViT-L/14 & OpenAI CLIP ViT-L/14 & OpenCLIP ViT-L/14 (DataComp-1B) \\
\midrule
\multicolumn{4}{l}{\textit{Inequality required by \Cref{thm:linear-sampled-path-noisy-has}}} \\
$D<\rho_{\min}/4$ holds? & No & No & No \\
\midrule
\multicolumn{4}{l}{\textit{Realized transfer}} \\
Mean $\HASViol^{\mathrm{img}}(u)$ across anchors (\%) & 25.55 & 2.95 & 2.42 \\
Mean $\HASViol^{\mathrm{txt}}(u)$ across anchors (\%) & 27.11 & 9.58 & 8.35 \\
Worst-anchor slack (must hold at every anchor) & 0.123 & 0.184 & 0.259 \\
Median image margin & 0.03 & 0.19 & 0.28 \\
$\Pr[\text{text triplet violates}]$ (\%) & 26.6 & 9.2 & 7.9 \\
$\Pr[\text{text triplet violates}\mid\text{image triplet violates}]$ (\%) & 62 & 79 & 86 \\
$\Pr[\text{text violates}\mid\text{image violates}]/\Pr[\text{text violates}]$ & $2.3\times$ & $8.6\times$ & $10.9\times$ \\
\bottomrule
\end{tabular}%
}
\end{table}

\Cref{tab:per-triplet-slack} shows that the theorem's bound is loose for these prototypes. Although $D<\rho_{\min}/4$ does not hold for any backbone, \Cref{eq:per-triplet-has-bound} holds at every anchor with a post hoc slack well below the largest possible cosine-margin difference of $2$. The worst-anchor slack is comparable to the median image margin for the two pretrained backbones and exceeds it for the randomly initialized encoder.

For the randomly initialized encoder, the image and text geometries are most similar because their mean violation percentages nearly match. The worst-anchor slack shows that this agreement is not uniform across anchors.

\myparagraph{The slack bound is loose because the perturbation is structured}
The realized perturbation $\be_k=-\xi_k^{\mathrm{lin}}\bar{\bz}$ is rank one: every text embedding moves along the single direction $-\bar{\bz}$, differing only in magnitude. By contrast, in \Cref{lem:cosine-stability} we bound perturbations in arbitrary directions by $4\eta/\rho_{\min}$ per cosine and $8\eta/\rho_{\min}$ for the cosine difference in \Cref{def:has}. In the worst case, the perturbations at $u$, $v$, and $w$ point in directions that maximize the change in the cosine difference. Here the parallel shifts largely cancel in the cosine difference. The gap between the theorem's bound and the observed transfer is therefore due to the worst-case bound, which allows perturbations in any direction, not a failure of the transfer mechanism.

To obtain a useful a priori threshold for these real embeddings, we may need an analysis that accounts for the perturbations sharing one direction, which we leave open.

\myparagraph{Image-side violations persist, but new text-side violations also arise}
If image- and text-side violations were independent, the conditional and marginal text-side violation probabilities would be equal. Instead, their ratio exceeds one for every backbone in \Cref{tab:per-triplet-slack}. Thus an image-side violation is more likely to remain after transfer than an arbitrary triplet is to violate on the text side.

A high conditional violation probability does not mean that text-side violations occur only where image-side ones do. On the trained backbones, the per-anchor text violation percentage substantially exceeds the image percentage, so many text-side violations have no image-side counterpart. We conclude only that, where \Cref{thm:linear-sampled-path-noisy-has} gives no guarantee, substantial approximate \HAS{} structure remains after transfer while new violations also arise.

\subsection{Caption Permutation Disrupts Approximate \HAS{} Transfer}\label{app:caption-permutation}

We test whether approximate \HAS{} transfer depends on caption--image pairings or only on the marginal caption distribution. Under the path-consistent setup of \Cref{sec:has-setup}, each caption co-occurs only with images from its descendant leaves.

Using the ImageNet-1K / WordNet prototypes from three ViT-L/14 backbones in \Cref{app:per-triplet-empirical}, we compute the closed-form linear-loss text minimizer $\bA^{\mathrm{lin},*}$ from \Cref{prop:linear-sampled-path-minimizer} with path-consistent captions and with hypernyms randomly permuted across images. The permutation preserves the image prototypes and marginal caption distribution but disrupts path consistency. We repeat the ablation in the synthetic InfoNCE setting of \Cref{sec:exp-low-lambda}, varying $\lambda$ to test whether stronger regularization compensates for the permutation.

\begin{table}[tbp]
\centering
\caption{Caption-permutation ablations for (a) the closed-form linear-loss text optimum on fixed real image prototypes and (b) the synthetic InfoNCE setting of \Cref{sec:exp-low-lambda}. Permuted entries are mean $\pm$ standard deviation over five permutation seeds.}
\label{tab:caption-permutation-real}
\label{tab:caption-permutation-synthetic}
\setlength{\tabcolsep}{5pt}
\textit{(a) Real prototypes with the linear contrastive loss}\par\smallskip
\begin{tabular}{lccc}
\toprule
& & \multicolumn{2}{c}{$\HASViol^{\mathrm{txt}}$ (\%)} \\
\cmidrule(lr){3-4}
Fixed image embeddings & $\HASViol^{\mathrm{img}}$ (\%) & path-consistent & permuted \\
\midrule
Randomly initialized ViT-L/14 & $25.55$ & $27.11$ & $61.63 \pm 0.24$ \\
OpenAI CLIP ViT-L/14 & $2.95$ & $9.58$ & $54.06 \pm 0.21$ \\
OpenCLIP ViT-L/14 (DataComp-1B) & $2.42$ & $8.35$ & $54.17 \pm 0.16$ \\
\bottomrule
\end{tabular}
\par\medskip
\textit{(b) Synthetic prototypes with the InfoNCE loss}\par\smallskip
\begin{tabular}{lcc}
\toprule
& \multicolumn{2}{c}{$\HASViol^{\mathrm{txt}}$ (\%)} \\
\cmidrule(lr){2-3}
$\lambda$ & path-consistent & permuted \\
\midrule
$10^{-5}$ & $1.76$ & $64.3 \pm 4.6$ \\
$10^{-4}$ & $0.06$ & $59.7 \pm 4.9$ \\
$10^{-3}$ & $0.00$ & $59.3 \pm 4.5$ \\
\bottomrule
\end{tabular}
\end{table}

In \Cref{tab:caption-permutation-real}(a), permutation more than doubles the text-side violation percentage for every backbone, raising it above half. In \Cref{tab:caption-permutation-real} (b), increasing $\lambda$ reduces the path-consistent violation percentage to zero, but not the permuted percentage. Thus, in both settings, preserving the marginal caption distribution alone does not preserve approximate \HAS{} transfer.

\subsection{Text-side \HAS{} is strongest with path-consistent captions under joint CLIP training}\label{app:trained-clip-captions}

In \Cref{app:caption-permutation}, we fix the image embeddings and permute hypernym captions to test whether text-side \HAS{} depends on their image pairings. Here we train both CLIP encoders on ImageNet and compare caption pairings.

\myparagraph{Training procedure}
We train CLIP ViT-B/32~\citep{radford2021learning,dosovitskiy2020image} from scratch with OpenCLIP~\citep{cherti2023reproducible} on the $1{,}281{,}167$ ImageNet-1K training images for $32$ epochs. We use AdamW ($\beta_1=0.9$, $\beta_2=0.98$, $\epsilon=10^{-6}$), learning rate $10^{-3}$, and weight decay $0.1$. We use batch size $1024$, $2000$ warmup steps, and a learned temperature. For each image, we fix one caption across epochs by placing a hierarchy node's name in one of the $80$ ImageNet prompt templates of \citet{radford2021learning}.

\myparagraph{Caption pairings}
We compare the following caption pairings:
\begin{itemize}
\item \emph{path-consistent}: draw uniformly from the image's leaf and its ancestors in the WordNet hierarchy of \Cref{sec:hierarchical-angular-separation-empirical}, following \Cref{sec:has-setup};
\item \emph{leaf-only}: choose the image's leaf;
\item \emph{permuted}: start with the path-consistent draw, then shuffle non-leaf caption labels across images within each depth, preserving each caption count.
\end{itemize}

\begin{figure}[tbp]
\centering
\includegraphics[width=\linewidth]{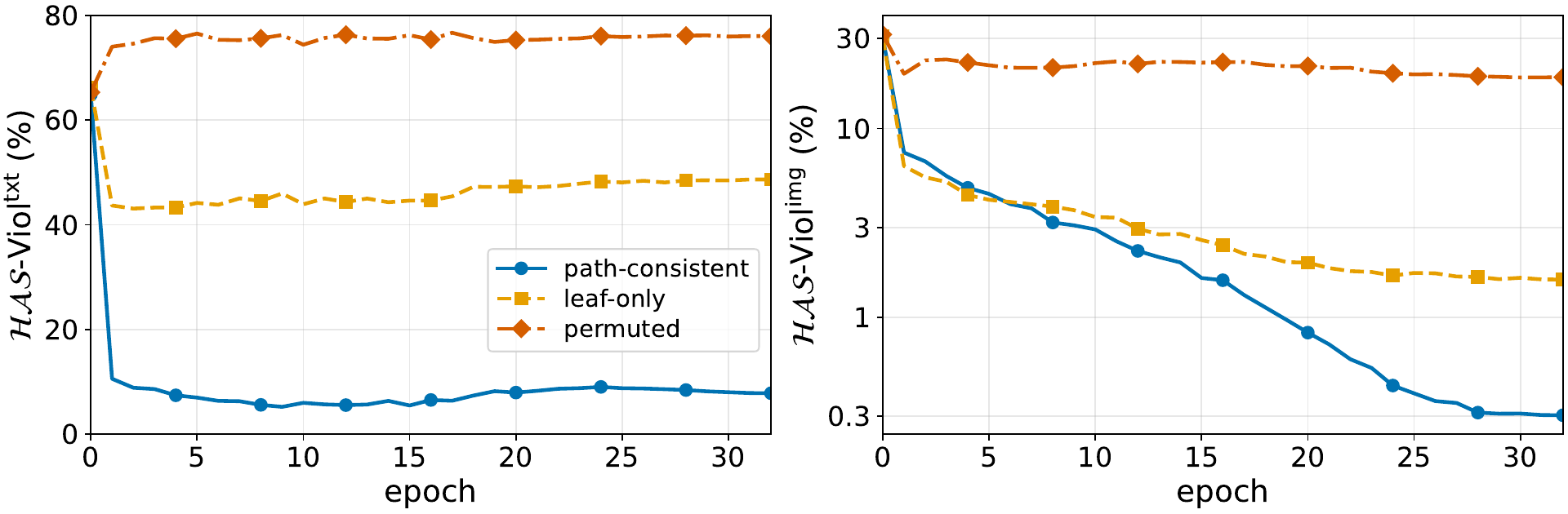}
\caption{Text-side \HASViol{} is lowest with path-consistent captions during joint CLIP training. We plot text (left) and image (right, log scale) \HAS violation percentages (lower is better). Path-consistent captions pair an image with a leaf or ancestor; leaf-only captions only pair with the leaf; in the permuted pairing, we shuffle hypernyms among images at the same depth.}
\label{fig:trained-clip-captions}
\end{figure}

After every epoch, we measure \HASViol{} using the $375$ anchors and $95$th-percentile descendant threshold of \Cref{sec:hierarchical-angular-separation-empirical}. We average the $80$ prompt templates for each text concept and aggregate $50$ fixed training images per leaf with descendant-leaf weighting for each image concept.
In the left panel of \Cref{fig:trained-clip-captions}, the path-consistent pairing has the lowest \HASViol{} from the first epoch onward. Text-side violations fall less under leaf-only captions and remain high under permutation.

\section{Alternating Optimization: The Image-Side Update}\label{app:alternating}

The fixed-image results explain how image-side approximate \HAS transfers to text, but not how an image-side update changes variation among images. We therefore analyze an exact image update after the text-side update. Images sharing a caption then receive the same embedding; their input variation enters through caption-group averages, the global image mean, and a common shift weighted by caption counts. In \Cref{prop:image-side-update}, we bound the stochastic deviation of these updated embeddings.

We analyze one exact text-side step followed by one exact image-side step under the linear contrastive loss, with ridge regularization on both sides. We use the linear loss because the InfoNCE image-side step has no closed form.
\begin{equation}\label{eq:objective-alternating}
  \cL^{\mathrm{alt}}(\bA,\bZ)
  :=
  -\mathrm{tr}\!\left(\bA^\top \bZ\mathbf{B}^\top\right)
  +
  \frac{\lambda}{2}\|\bA\|_F^2
  +
  \frac{\lambda_Z}{2}\|\bZ\|_F^2,
\end{equation}
Here $\lambda,\lambda_Z>0$ and $\mathbf{B}$ is the label-encoding matrix of \Cref{eq:linear-sampled-path-matrix}; by \Cref{eq:objective-linear-sampled-path-trace}, the first two terms are exactly the regularized linear contrastive loss of \Cref{eq:objective-linear-sampled-path}. The cross-modal term is bilinear in $(\bA,\bZ)$, and the objective is strongly convex in either argument when the other is fixed. We obtain the exact minimizer of each step from \Cref{prop:linear-sampled-path-minimizer}.

To obtain the image-side update, we first analyze the same label-encoding matrix $\mathbf{B}$ used in the $\bA$-step by columns, splitting each entry into a matched-caption contribution and an off-caption coefficient shared across image columns. Set $\kappa:=1+\frac{1}{2(M-1)}+\frac{1}{2(N-1)}$ and, for $k\in[M]$, $\gamma_k^-:=\frac{1}{2(M-1)}+\frac{N_k}{2(N-1)}$. Expanding \Cref{eq:linear-sampled-path-matrix} and collecting terms in $Y_{kj}$ gives
\begin{equation}\label{eq:alternating-label-encoding}
  \begin{aligned}
    B_{kj}
    &=Y_{kj}+\frac{Y_{kj}-1}{2(M-1)}+\frac{Y_{kj}-N_k}{2(N-1)}\\
    &=\kappa Y_{kj}-\gamma_k^-,
    \qquad (k,j)\in[M]\times[N].
  \end{aligned}
\end{equation}
The indicator $Y_{kj}$ selects the paired text embedding, while summing $-\gamma_k^-$ against the text embeddings gives the uniform and caption-frequency-weighted means in the common shift below.

Fix $\bA\in\R^{d\times M}$. Applying \Cref{prop:linear-sampled-path-minimizer} to \Cref{eq:objective-alternating} gives the image-side minimizer
\begin{equation}\label{eq:z-step-minimizer}
  \bZ^{\mathrm{lin},*}(\bA)
  =
  \frac{1}{\lambda_Z}\,\bA\,\mathbf{B}.
\end{equation}
To express its columns, define the uniform and caption-frequency-weighted text means
\begin{equation}\label{eq:z-step-means}
  \bar{\ba} := \frac{1}{M}\sum_{k=1}^{M}\ba_k,
  \qquad
  \bar{\ba}_{\mathrm{capt}}
  :=
  \frac{1}{N}\sum_{k=1}^{M}N_k\,\ba_k
  =
  \frac{1}{N}\sum_{j=1}^{N}\ba_{c_j}.
\end{equation}
Using \Cref{eq:alternating-label-encoding} and the means in \Cref{eq:z-step-means}, the $j$-th column of \Cref{eq:z-step-minimizer} is
\begin{align}
  \bz_j^{\mathrm{lin},*}(\bA)
  &=\frac{1}{\lambda_Z}\sum_{k=1}^{M}B_{kj}\ba_k\notag\\
  &=\frac{1}{\lambda_Z}\left(\kappa\ba_{c_j}-\sum_{k=1}^{M}\gamma_k^-\ba_k\right)\notag\\
  &=\frac{1}{\lambda_Z}\left[
    \kappa\ba_{c_j}
    -
    \frac{M}{2(M-1)}\,\bar{\ba}
    -
    \frac{N}{2(N-1)}\,\bar{\ba}_{\mathrm{capt}}
  \right]
  \qquad\text{for every }j\in[N].\label{eq:z-step-columns}
\end{align}
Thus images with the same caption receive the same embedding, even before imposing the text-side optimum.

In \Cref{eq:z-step-columns}, the positive term favors the text embedding paired with image $j$, while the two subtracted means play the role of the global image mean $\bar{\bz}$ in \Cref{cor:linear-sampled-path-centroid-form}. To simplify this image-side update after the text step, we first show that its uniform text mean vanishes.

\begin{lemma}[Global centering at the $\bA$-step optimum]\label{lem:a-step-centered}
Every column of $\mathbf{B}$ sums to zero: $\sum_{k=1}^{M}B_{kj}=0$ for every $j\in[N]$. Consequently, for every $\bZ$, the $\bA$-step optimum of \Cref{prop:linear-sampled-path-minimizer} satisfies
\begin{equation}\label{eq:a-step-centered}
  \sum_{k=1}^{M}\ba_k^{\mathrm{lin},*}=\bzero.
\end{equation}
\end{lemma}
\begin{proof}
Fix $j\in[N]$. Each image has exactly one caption, so $\sum_{k=1}^{M}Y_{kj}=1$; also $\sum_{k=1}^{M}N_k=N$. Therefore, by \Cref{eq:alternating-label-encoding},
\begin{equation}
  \sum_{k=1}^{M}B_{kj}
  =
  \kappa-\frac{M}{2(M-1)}-\frac{N}{2(N-1)}
  =
  0,
\end{equation}
Thus $\mathbf{B}^\top\bone=\bzero$ with $\bone\in\R^M$ the all-ones vector. Since $\sum_{k=1}^{M}\ba_k^{\mathrm{lin},*}=\bA^{\mathrm{lin},*}\bone$, we obtain \Cref{eq:a-step-centered} from \Cref{prop:linear-sampled-path-minimizer}.
\end{proof}

At the $\bA$-step optimum, we remove the uniform mean from \Cref{eq:z-step-columns} using \Cref{lem:a-step-centered}. The remaining terms give an affine copy of the text embeddings.

\begin{corollary}[Image-side update as a scaled, shifted text embedding]\label{cor:z-step-affine-copy}
Let $\bA=\bA^{\mathrm{lin},*}$ for arbitrary fixed $\bZ$. With $\kappa$ from \Cref{eq:alternating-label-encoding}, define the \emph{caption-mean shift}
\begin{equation}\label{eq:caption-shift}
  \be_{\mathrm{capt}}
  :=
  -\frac{N}{2(N-1)\,\kappa}\,\bar{\ba}_{\mathrm{capt}}.
\end{equation}
Then the $\bZ$-step optimum is a scaled, shifted copy of the caption's text embedding:
\begin{equation}\label{eq:z-step-affine-copy}
  \bz_j^{\mathrm{lin},*}
  =
  \frac{\kappa}{\lambda_Z}\left(\ba_{c_j}^{\mathrm{lin},*}+\be_{\mathrm{capt}}\right)
  \qquad\text{for every } j\in[N].
\end{equation}
\end{corollary}
\begin{proof}
By \Cref{lem:a-step-centered}, $\bar{\ba}=\bzero$ at $\bA=\bA^{\mathrm{lin},*}$. Substituting this and \Cref{eq:caption-shift} into \Cref{eq:z-step-columns} gives \Cref{eq:z-step-affine-copy}.
\end{proof}

This update makes images with the same caption identical. The remaining stochastic deviation comes from caption-group averages and a common caption-count-weighted shift; in \Cref{prop:image-side-update}, we bound it after removing the update's positive scale.

\begin{proposition}[Caption-group averaging in the image update]\label{prop:image-side-update}
Suppose \Cref{ass:noisy-prototype,ass:leaf-uniform-caption,ass:prototype-centroid-consistency} hold. Starting from $\bZ$, compute the exact text-side minimizer $\bA^{\mathrm{lin},*}$ and then the exact image-side minimizer $\bZ^{\mathrm{lin},*}$. Fix the observed leaf and caption labels, and let $\mathbb{E}_{\bm{\varepsilon}}$ average over the independent variation vectors of \Cref{ass:noisy-prototype}. For each non-root $k$ and any $j\in\cP_k$, define the normalized stochastic deviation
\begin{equation}\label{eq:image-update-centered-deviation}
  \bs_k
  :=
  \frac{\lambda\lambda_Z}{\kappa^2N_k}\left(\bz_j^{\mathrm{lin},*}-\mathbb{E}_{\bm{\varepsilon}}\bz_j^{\mathrm{lin},*}\right),
\end{equation}
where $\kappa$ is from \Cref{eq:alternating-label-encoding} and the choice of $j\in\cP_k$ does not matter by \Cref{eq:z-step-affine-copy}. The effect of the common caption-count-weighted shift is measured by
\begin{equation}\label{eq:caption-concentration}
  \omega
  :=
  \frac{\sqrt{\sum_{k=1}^{M}N_k^3}}{(N-1)\sqrt{N_{\min}}}.
\end{equation}
With $\xi_{\max}^{\mathrm{lin}}$ defined before \Cref{thm:linear-sampled-path-noisy-has}, there exists a universal constant $C>0$ such that, for every $\epsilon\in(0,1)$, with probability at least $1-\epsilon$,
\begin{equation}\label{eq:image-update-noise-contraction}
  \max_{k\neq\mathrm{root}}\|\bs_k\|
  \le
  C\sigma\left(
    \bigl(1+\omega\bigr)\sqrt{\frac{d+\log(M-1)+\log(8/\epsilon)}{N_{\min}}}
    +\xi_{\max}^{\mathrm{lin}}\sqrt{\frac{d+\log(8/\epsilon)}{N}}
  \right).
\end{equation}
\end{proposition}

\begin{proof}

\textbf{Step 1: Identify the centered image update.}
We first write the common shift $\be_{\mathrm{capt}}$ in terms of the input images. For each image $j\in[N]$, define its caption-count weight by
\begin{equation}\label{eq:caption-count-weights}
  \begin{aligned}
    h_j
    &:=N_{c_j}-\frac{1}{\kappa}\sum_{k=1}^{M}N_k\gamma_k^-\\
    &=N_{c_j}-\frac{1}{\kappa}\left(\frac{N}{2(M-1)}+\frac{\sum_{k=1}^{M}N_k^2}{2(N-1)}\right).
  \end{aligned}
\end{equation}
The common shift in \Cref{eq:caption-shift} uses the $N_k/N$-weighted text mean of \Cref{eq:z-step-means}. The text-side minimizer of \Cref{prop:linear-sampled-path-minimizer} and the label encoding in \Cref{eq:alternating-label-encoding} apply to every $k\in[M]$, including nodes with $N_k=0$. Substituting them into the weighted mean gives
\begin{align}
  \lambda\bar{\ba}_{\mathrm{capt}}
  &=\frac{1}{N}\sum_{k=1}^{M}N_k\lambda\ba_k^{\mathrm{lin},*}
  &&\text{(by \Cref{eq:z-step-means})}\notag\\
  &=\frac{\kappa}{N}\sum_{k=1}^{M}N_k\sum_{i\in\cP_k}\bz_i-\left(\sum_{k=1}^{M}N_k\gamma_k^-\right)\bar{\bz}
  &&\text{(by \Cref{eq:linear-sampled-path-minimizer,eq:alternating-label-encoding})}\notag\\
  &=\frac{\kappa}{N}\sum_{j=1}^{N}N_{c_j}\bz_j-\frac{1}{N}\sum_{j=1}^{N}\left(\sum_{k=1}^{M}N_k\gamma_k^-\right)\bz_j
  &&\text{(since $j\in\cP_{c_j}$ and $\bar{\bz}=\tfrac{1}{N}\textstyle\sum_{j}\bz_j$)}\notag\\
  &=\frac{\kappa}{N}\sum_{j=1}^{N}\left(N_{c_j}-\frac{1}{\kappa}\sum_{k=1}^{M}N_k\gamma_k^-\right)\bz_j
  &&\text{(collecting terms in $\bz_j$)}\notag\\
  &=\frac{\kappa}{N}\sum_{j=1}^{N}h_j\bz_j
  &&\text{(by \Cref{eq:caption-count-weights})}.
\end{align}
Substituting this expression into \Cref{eq:caption-shift} gives
\begin{align}
  \be_{\mathrm{capt}}
  &=-\frac{N}{2(N-1)\kappa}\bar{\ba}_{\mathrm{capt}}\notag\\
  &=-\frac{1}{2(N-1)\lambda}\sum_{j=1}^{N}h_j\bz_j.\label{eq:caption-shift-images}
\end{align}

Fix a non-root $k$ and $j\in\cP_k$, so $c_j=k$. Substituting this shift and the text-side empirical-mean expression of \Cref{cor:linear-sampled-path-centroid-form} into \Cref{eq:z-step-affine-copy} gives
\begin{align}
  \bz_j^{\mathrm{lin},*}
  &=\frac{\kappa}{\lambda_Z}\left(\ba_k^{\mathrm{lin},*}+\be_{\mathrm{capt}}\right)
  &&\text{(by \Cref{eq:z-step-affine-copy} with $c_j=k$)}\notag\\
  &=\frac{\kappa}{\lambda_Z}\left[\frac{\kappa N_k}{\lambda}\left(\bar{\bz}_k-\xi_k^{\mathrm{lin}}\bar{\bz}\right)-\frac{1}{2(N-1)\lambda}\sum_{i=1}^{N}h_i\bz_i\right]
  &&\text{(by \Cref{eq:linear-sampled-path-centroid-form,eq:caption-shift-images})}\notag\\
  &=\frac{\kappa^2N_k}{\lambda\lambda_Z}\left(
    \bar{\bz}_k-\xi_k^{\mathrm{lin}}\bar{\bz}
    -\frac{1}{2\kappa(N-1)N_k}\sum_{i=1}^{N}h_i\bz_i
  \right)
  &&\text{(factoring out $\kappa^2N_k/(\lambda\lambda_Z)$)}.\label{eq:image-update-empirical-form}
\end{align}

The path-consistent captions and equal image counts across leaves from \Cref{sec:has-setup}, together with \Cref{ass:leaf-uniform-caption,ass:prototype-centroid-consistency,ass:noisy-prototype}, allow us to apply \Cref{lem:caption-mean-decomposition} and \Cref{cor:zero-caption-bias}. With $\xi_k^{\mathrm{lin}}$ from \Cref{cor:linear-sampled-path-noisy-centroid-form}, substituting \Cref{eq:caption-bias-empirical-means} with $\bm{b}_k=\bm{b}=\bzero$ and $\bz_i=\bm{\mu}_{y_i}+\bm{\varepsilon}_i$ for every $i\in[N]$ gives
\begin{align}
  \frac{\lambda\lambda_Z}{\kappa^2N_k}\bz_j^{\mathrm{lin},*}
  &=\left(\bm{\mu}_k+\bar{\bm{\varepsilon}}_k\right)-\xi_k^{\mathrm{lin}}\left(\bm{\mu}_{\mathrm{root}}+\bar{\bm{\varepsilon}}\right)
  &&\text{(by \Cref{eq:image-update-empirical-form,eq:caption-bias-empirical-means})}\\
  &\quad-\frac{1}{2\kappa(N-1)N_k}\sum_{i=1}^{N}h_i\left(\bm{\mu}_{y_i}+\bm{\varepsilon}_i\right)\\
  &=\bm{\mu}_k-\xi_k^{\mathrm{lin}}\bm{\mu}_{\mathrm{root}}-\frac{1}{2\kappa(N-1)N_k}\sum_{i=1}^{N}h_i\bm{\mu}_{y_i}
  &&\text{(grouping prototype and noise terms)}\\
  &\quad+\left(\bar{\bm{\varepsilon}}_k-\xi_k^{\mathrm{lin}}\bar{\bm{\varepsilon}}-\frac{1}{2\kappa(N-1)N_k}\sum_{i=1}^{N}h_i\bm{\varepsilon}_i\right).
\end{align}
The parenthesized noise term has zero mean by \Cref{ass:noisy-prototype}, so it equals $\bs_k$ in \Cref{eq:image-update-centered-deviation}. Thus the exact prototype decomposition is
\begin{equation}\label{eq:image-update-prototype-form}
  \bz_j^{\mathrm{lin},*}
  =
  \frac{\kappa^2N_k}{\lambda\lambda_Z}\left(
    \bm{\mu}_k-\xi_k^{\mathrm{lin}}\bm{\mu}_{\mathrm{root}}
    -\frac{1}{2\kappa(N-1)N_k}\sum_{i=1}^{N}h_i\bm{\mu}_{y_i}
    +\bs_k
  \right).
\end{equation}

\textbf{Step 2: Bound the caption-group and global means.}
By \Cref{ass:leaf-uniform-caption}, $N_k\ge N_{\min}>0$ for every non-root $k$, and by \Cref{ass:noisy-prototype}, the variations are independent, centered, and sub-Gaussian. Apply \Cref{lem:subgaussian-empirical-means} with $\cK=V\setminus\{\mathrm{root}\}$, $I_k=\cP_k$ for $k\in\cK$, coefficients $\{\xi_k^{\mathrm{lin}}\}_{k\in\cK}$, and $u=\log(8/\epsilon)$. Since $|\cK|=M-1$ and $\min_{k\in\cK}|I_k|=N_{\min}$, with probability at least $1-\epsilon/2$,
\begin{equation}
  \max_{k\neq\mathrm{root}}\|\bar{\bm{\varepsilon}}_k-\xi_k^{\mathrm{lin}}\bar{\bm{\varepsilon}}\|
  \le
  C\sigma\left(
    \sqrt{\frac{d+\log(M-1)+\log(8/\epsilon)}{N_{\min}}}
    +\xi_{\max}^{\mathrm{lin}}\sqrt{\frac{d+\log(8/\epsilon)}{N}}
  \right).
\end{equation}

\textbf{Step 3: Bound the common weighted shift and combine the events.}
We first control the caption-count weights. By \Cref{eq:caption-count-weights}, $N_{c_j}-h_j$ is nonnegative and takes the same value for every $j\in[N]$. Since $\sum_{k=1}^{M}N_k=N$, Cauchy--Schwarz gives $N^2\le M\sum_{k=1}^{M}N_k^2$. Hence
\begin{align}
  &\sum_{j=1}^{N}\left(N_{c_j}+h_j\right)\notag\\
  &\quad=\sum_{j=1}^{N}\left[2N_{c_j}-\frac{1}{\kappa}\left(\frac{N}{2(M-1)}+\frac{\sum_{k=1}^{M}N_k^2}{2(N-1)}\right)\right]
  &&\text{(by \Cref{eq:caption-count-weights})}\notag\\
  &\quad=2\sum_{k=1}^{M}N_k^2-\frac{1}{\kappa}\left(\frac{N^2}{2(M-1)}+\frac{N\sum_{k=1}^{M}N_k^2}{2(N-1)}\right)
  &&\text{(since $N_k$ images have $c_j=k$)}\notag\\
  &\quad\ge\left(2-\frac{M}{2(M-1)}-\frac{N}{2(N-1)}\right)\sum_{k=1}^{M}N_k^2
  &&\text{(since $\kappa\ge1$ and $N^2\le M\textstyle\sum_{k}N_k^2$)}\notag\\
  &\quad\ge0
  &&\text{(since $M\ge2$ and $N\ge2$)}.\label{eq:caption-weight-sum-bound}
\end{align}
Therefore, with $\bw:=(h_1,\dots,h_N)$,
\begin{align}
  \|\bw\|^2
  &=\sum_{j=1}^{N}N_{c_j}^2-\sum_{j=1}^{N}\left(N_{c_j}-h_j\right)\left(N_{c_j}+h_j\right)
  &&\text{(difference of squares)}\notag\\
  &\le\sum_{j=1}^{N}N_{c_j}^2
  &&\text{(by \Cref{eq:caption-weight-sum-bound}, since $N_{c_j}-h_j\ge0$ is constant)}\notag\\
  &=\sum_{k=1}^{M}N_k^3
  &&\text{(since $N_k$ images have $c_j=k$)}.\label{eq:caption-weight-norm-bound}
\end{align}

The Hoeffding and net argument in the proof of \Cref{lem:subgaussian-empirical-means} applies to the fixed weights $\bw$ and gives $\Pr\bigl(\|\sum_{j=1}^{N}h_j\bm{\varepsilon}_j\|>C\sigma\|\bw\|\sqrt{d+t}\bigr)\le 2e^{-t}$ for every $t>0$. Taking $t=\log(4/\epsilon)$ makes the probability of exceeding this bound at most $\epsilon/2$. On the complementary event,
\begin{align}
  &\max_{k\neq\mathrm{root}}
  \left\|\frac{1}{2\kappa(N-1)N_k}\sum_{j=1}^{N}h_j\bm{\varepsilon}_j\right\|\notag\\
  &\quad\le\frac{1}{2(N-1)N_{\min}}\left\|\sum_{j=1}^{N}h_j\bm{\varepsilon}_j\right\|
  &&\text{(since $\kappa\ge1$ and $N_k\ge N_{\min}$)}\notag\\
  &\quad\le\frac{C\sigma\|\bw\|\sqrt{d+\log(4/\epsilon)}}{2(N-1)N_{\min}}
  &&\text{(by the Hoeffding bound with $t=\log(4/\epsilon)$)}\notag\\
  &\quad\le\frac{C\sigma\sqrt{\sum_{k=1}^{M}N_k^3}\sqrt{d+\log(4/\epsilon)}}{2(N-1)N_{\min}}
  &&\text{(by \Cref{eq:caption-weight-norm-bound})}\notag\\
  &\quad=\frac{C}{2}\,\sigma\,\omega\sqrt{\frac{d+\log(4/\epsilon)}{N_{\min}}}
  &&\text{(by \Cref{eq:caption-concentration})}.
\end{align}
Since $\log(4/\epsilon)\le\log(M-1)+\log(8/\epsilon)$, a union bound over the two failure events gives \Cref{eq:image-update-noise-contraction} with probability at least $1-\epsilon$.
\end{proof}

In \Cref{lem:subgaussian-feature-max-norm}, we bound per-instance input noise at scale $C\sigma\sqrt{d+\log(2N/\epsilon)}$. The caption-group term in \Cref{eq:image-update-noise-contraction} is smaller by order $\sqrt{N_{\min}}/(1+\omega)$, up to logarithmic factors. The separate $N^{-1/2}$ term comes from the global image mean.

The result concerns one exact alternating step under the linear loss, so it is an analytic surrogate for joint gradient descent rather than a model of CLIP training. The bound controls stochastic deviation around the caption-grouped prototype decomposition, not the deterministic prototype margin: reduced variation can protect a small positive margin, but the bound alone does not certify reversal of a negative margin. Whether the same contraction holds for the subtree-aggregated prototypes of \Cref{sec:hierarchical-angular-separation-empirical}, which mix caption groups along root-to-node paths, remains open.

\section{Relaxing the Caption-Symmetry Assumptions}\label{app:caption-relaxation}

In \Cref{lem:caption-bias-transfer}, \HAS transfer under observed captions depends on a bound on caption bias. In \Cref{app:caption-bias-bounds}, we bound two sources of that bias, caption imbalance and inexact prototype consistency, and give transfer conditions for both losses. We then examine partial ancestor omission, which creates random caption imbalance, in \Cref{app:partial-omission}. Finally, in \Cref{app:total-omission}, we analyze total omission, for which a caption mean is undefined. Throughout this section, we sample exactly $n$ images per leaf as in \Cref{sec:has-setup}. In \Cref{app:caption-permutation}, we support these bounds with empirical results: permuting which images match each caption increases the caption bias well outside the small-bias regime of \Cref{cor:caption-relaxation}, and approximate \HAS{} transfer fails.

\subsection{Caption Imbalance and Inexact Prototype Consistency}\label{app:caption-bias-bounds}

Caption imbalance changes how leaf prototypes contribute to a concept's caption mean. For example, \texttt{dog} captions drawn mostly from dalmatians overrepresent one descendant leaf; off-path captions add images from outside the dog's subtree. Inexact prototype consistency adds bias even when captions are leaf-uniform. To bound these effects, for every non-root node $k$, split its positive set into
\begin{equation}\label{eq:off-path-split}
  \cP_k^{\mathrm{on}} := \left\{j\in\cP_k : y_j\in\mathrm{Sub}(k)\right\},
  \qquad
  \cP_k^{\mathrm{off}} := \cP_k\setminus\cP_k^{\mathrm{on}},
\end{equation}
where $L_k:=\mathrm{Sub}(k)\cap V_{\mathrm{leaf}}$ denotes the leaf descendants of each node $k\in V$, and $n_{k\ell}^{\mathrm{on}}:=|\{j\in\cP_k^{\mathrm{on}}:y_j=\ell\}|$ for $\ell\in L_k$ counts its on-path captions from leaf $\ell$. A nonempty, leaf-uniform on-path set has prototype mean $\frac{1}{|L_k|}\sum_{\ell\in L_k}\bm{\mu}_\ell$, which equals $\bm{\mu}_k$ under \Cref{ass:prototype-centroid-consistency}. When every non-root node has $N_k>0$, define the caption imbalance by
\begin{equation}\label{eq:caption-relaxation-budget}
  \delta_{\mathrm{cap}}
  :=
  \max_{k\neq\mathrm{root}}\biggl[
    \underbrace{\frac{1}{N_k}\underbrace{\sum_{\ell\in L_k}}_{\makebox[0pt]{\scriptsize leaves of $k$}}\overbrace{\left|n_{k\ell}^{\mathrm{on}}-\frac{|\cP_k^{\mathrm{on}}|}{|L_k|}\right|}^{\substack{\text{extra or missing}\\\text{captions of leaf $\ell$}}}\overbrace{\|\bm{\mu}_\ell-\bm{\mu}_k\|}^{\substack{\text{prototype distance}\\\text{of leaf $\ell$}}}}_{\text{uneven leaf coverage}}
    +\underbrace{\frac{1}{N_k}\underbrace{\sum_{j\in\cP_k^{\mathrm{off}}}}_{\makebox[0pt]{$\substack{\text{images captioned $k$}\\\text{with leaf outside $\mathrm{Sub}(k)$}}$}}\overbrace{\|\bm{\mu}_{y_j}-\bm{\mu}_k\|}^{\substack{\text{prototype distance}\\\text{of image $j$}}}}_{\text{off-path captions}}
  \biggr],
\end{equation}
and the prototype mismatch by
\begin{equation}\label{eq:caption-consistency-budget}
  \delta_{\mathrm{cons}}:=\max_{k\in V}\left\|\frac{1}{|L_k|}\sum_{\ell\in L_k}\bm{\mu}_\ell-\bm{\mu}_k\right\|.
\end{equation}
The maximum in $\delta_{\mathrm{cons}}$ includes the root because its prototype mismatch contributes to the global caption bias.

\begin{lemma}[Caption bias under imbalanced captions and inexact prototypes]\label{lem:off-path-bias}
Suppose exactly $n$ images are sampled per leaf and the observed captions satisfy $N_k>0$ for every non-root node $k$. Write $\bm{b}_k:=\bar{\bm{\mu}}_k-\bm{\mu}_k$ for the caption bias of each non-root node $k$ and $\bm{b}:=\bar{\bm{\mu}}-\bm{\mu}_{\mathrm{root}}$ for the global caption bias, as in \Cref{def:caption-bias}. Then
\begin{equation}\label{eq:off-path-bias-bound}
  \|\bm{b}\|\le\delta_{\mathrm{cons}},
  \qquad
  \max_{k\neq\mathrm{root}}\|\bm{b}_k\|\le\delta_{\mathrm{cap}}+\delta_{\mathrm{cons}}.
\end{equation}
In particular, $\delta_{\mathrm{cap}}+(1+\xi_{\max})\delta_{\mathrm{cons}}$ is a caption bias bound for any coefficients $\{\xi_k\in\R\}_{k\in V\setminus\{\mathrm{root}\}}$, where $\xi_{\max}:=\max_{k\neq\mathrm{root}}|\xi_k|$. Under \Cref{ass:prototype-centroid-consistency}, $\bm{b}=\bzero$ and $\delta_{\mathrm{cap}}$ is a caption bias bound for any such coefficients.
\end{lemma}

\begin{proof}
Balanced image sampling gives
\begin{align*}
  \bm{b}
  &=\bar{\bm{\mu}}-\bm{\mu}_{\mathrm{root}}
  &&\text{(by \Cref{def:caption-bias})}\\
  &=\frac{1}{|V_{\mathrm{leaf}}|}\sum_{\ell\in V_{\mathrm{leaf}}}\bm{\mu}_\ell-\bm{\mu}_{\mathrm{root}}
  &&\text{(since each leaf contributes $n$ images)}.
\end{align*}
Since $L_{\mathrm{root}}=V_{\mathrm{leaf}}$, this is the root term in \Cref{eq:caption-consistency-budget}, so $\|\bm{b}\|\le\delta_{\mathrm{cons}}$.
For a non-root node $k$ with $\cP_k^{\mathrm{on}}\neq\varnothing$, grouping images by their leaf and adding and subtracting the centroid of the descendant leaf prototypes give
\begin{align}
  \frac{1}{|\cP_k^{\mathrm{on}}|}\sum_{j\in\cP_k^{\mathrm{on}}}\left(\bm{\mu}_{y_j}-\bm{\mu}_k\right)
  &=\sum_{\ell\in L_k}\frac{n_{k\ell}^{\mathrm{on}}}{|\cP_k^{\mathrm{on}}|}\left(\bm{\mu}_\ell-\bm{\mu}_k\right)\notag\\
  &=\sum_{\ell\in L_k}\left(\frac{n_{k\ell}^{\mathrm{on}}}{|\cP_k^{\mathrm{on}}|}-\frac{1}{|L_k|}\right)\left(\bm{\mu}_\ell-\bm{\mu}_k\right)+\left(\frac{1}{|L_k|}\sum_{\ell\in L_k}\bm{\mu}_\ell-\bm{\mu}_k\right).\label{eq:caption-imbalance-centered}
\end{align}
By \Cref{def:caption-bias} and the split in \Cref{eq:off-path-split},
\begin{equation}\label{eq:off-path-decomposition}
  \bm{b}_k
  =
  \frac{1}{N_k}\sum_{j\in\cP_k^{\mathrm{on}}}\left(\bm{\mu}_{y_j}-\bm{\mu}_k\right)
  +
  \frac{1}{N_k}\sum_{j\in\cP_k^{\mathrm{off}}}\left(\bm{\mu}_{y_j}-\bm{\mu}_k\right).
\end{equation}
For $\cP_k^{\mathrm{on}}=\varnothing$, the first sum in \Cref{eq:off-path-decomposition} is zero. Otherwise, substituting \Cref{eq:caption-imbalance-centered} into that sum and applying the triangle inequality give
\begin{align*}
  \|\bm{b}_k\|
  &\le\frac{1}{N_k}\sum_{\ell\in L_k}\left|n_{k\ell}^{\mathrm{on}}-\frac{|\cP_k^{\mathrm{on}}|}{|L_k|}\right|\|\bm{\mu}_\ell-\bm{\mu}_k\|
  +\frac{1}{N_k}\sum_{j\in\cP_k^{\mathrm{off}}}\|\bm{\mu}_{y_j}-\bm{\mu}_k\|\\
  &\qquad+\frac{|\cP_k^{\mathrm{on}}|}{N_k}\left\|\frac{1}{|L_k|}\sum_{\ell\in L_k}\bm{\mu}_\ell-\bm{\mu}_k\right\|\\
  &\le\delta_{\mathrm{cap}}+\delta_{\mathrm{cons}},
\end{align*}
since the first two terms are the bracket in \Cref{eq:caption-relaxation-budget} at node $k$ and $|\cP_k^{\mathrm{on}}|\le N_k$. The same bound holds when $\cP_k^{\mathrm{on}}=\varnothing$, because only the off-path sum then remains. Taking the maximum over non-root $k$ gives the second bound in \Cref{eq:off-path-bias-bound}. Since $|\xi_k|\le\xi_{\max}$ for every non-root $k$, the triangle inequality gives
\begin{equation*}
  \max_{k\neq\mathrm{root}}\|\bm{b}_k-\xi_k\bm{b}\|
  \le\max_{k\neq\mathrm{root}}\|\bm{b}_k\|+\xi_{\max}\|\bm{b}\|
  \le\delta_{\mathrm{cap}}+(1+\xi_{\max})\delta_{\mathrm{cons}},
\end{equation*}
so $\delta_{\mathrm{cap}}+(1+\xi_{\max})\delta_{\mathrm{cons}}$ is a caption bias bound by \Cref{eq:caption-bias-level}. Under \Cref{ass:prototype-centroid-consistency}, $\delta_{\mathrm{cons}}=0$, so $\bm{b}=\bzero$ and this value reduces to $\delta_{\mathrm{cap}}$ for any coefficients.
\end{proof}

By \Cref{lem:off-path-bias}, we can use $\delta_{\mathrm{cap}}+(1+\xi_{\max})\delta_{\mathrm{cons}}$ as the additive bias value in \Cref{lem:caption-bias-transfer}. The next corollary gives the resulting transfer conditions. Within this bound, the uneven-coverage term vanishes for leaf-uniform on-path captions, the off-path term vanishes for path-consistent captions, and $\delta_{\mathrm{cons}}$ vanishes under \Cref{ass:prototype-centroid-consistency}.

\begin{corollary}[\HAS transfer under imbalanced captions and inexact prototypes]\label{cor:caption-relaxation}
Suppose \Cref{ass:noisy-prototype} holds, exactly $n$ images are sampled per leaf, $\rho_{\min}>0$, and the observed captions are deterministic with $N_k>0$ for every non-root node $k$. \Cref{ass:prototype-centroid-consistency} need not hold. Fix $\epsilon\in(0,1)$ and $\lambda>0$.
For the loss under consideration, compute $\xi_{\max}$ from its coefficients, assume the corresponding case below, and let $\tau$ denote its threshold.
\begin{description}
  \item[\textbf{Linear loss.}] Assume $D_\epsilon^{\mathrm{lin}}+\delta_{\mathrm{cap}}+(1+\xi_{\max})\delta_{\mathrm{cons}}<\rho_{\min}/4$ and set
  \begin{equation}\label{eq:caption-relaxation-linear-threshold}
    \tau:=\frac{8\left(D_\epsilon^{\mathrm{lin}}+\delta_{\mathrm{cap}}+(1+\xi_{\max})\delta_{\mathrm{cons}}\right)}{\rho_{\min}}.
  \end{equation}
  \item[\textbf{InfoNCE.}] Assume $D_\epsilon^{\mathrm{nce}}>0$, $\lambda\ge\lambda_0(\epsilon)$, and $2D_\epsilon^{\mathrm{nce}}+\delta_{\mathrm{cap}}+(1+\xi_{\max})\delta_{\mathrm{cons}}<\rho_{\min}/4$, and set
  \begin{equation}\label{eq:caption-relaxation-infonce-threshold}
    \tau:=\frac{8\left(2D_\epsilon^{\mathrm{nce}}+\delta_{\mathrm{cap}}+(1+\xi_{\max})\delta_{\mathrm{cons}}\right)}{\rho_{\min}}.
  \end{equation}
\end{description}
Then, with probability at least $1-\epsilon$, all of the following hold simultaneously:
\CaptionBiasTransferConclusions{\tau}
\end{corollary}

\begin{proof}
By \Cref{lem:off-path-bias}, $\delta_{\mathrm{cap}}+(1+\xi_{\max})\delta_{\mathrm{cons}}$ is a caption bias bound, so we can take $\delta_{\mathrm{bias}}$ equal to it. By \Cref{eq:caption-total-deviation}, $D_\epsilon^{\mathrm{bias}}$ then equals $D_\epsilon^{\mathrm{lin}}+\delta_{\mathrm{bias}}$ for the linear loss and $2D_\epsilon^{\mathrm{nce}}+\delta_{\mathrm{bias}}$ for InfoNCE. In both cases, it is less than $\rho_{\min}/4$ by assumption, and $\tau_\epsilon^{\mathrm{bias}}$ equals the threshold $\tau$ in \Cref{eq:caption-relaxation-linear-threshold} or~\eqref{eq:caption-relaxation-infonce-threshold}. The remaining InfoNCE hypotheses, $D_\epsilon^{\mathrm{nce}}>0$ and $\lambda\ge\lambda_0(\epsilon)$, also hold by assumption. Applying \Cref{lem:caption-bias-transfer} with this $\delta_{\mathrm{bias}}$ therefore gives parts (i)--(iii) at $\tau$ with probability at least $1-\epsilon$.
\end{proof}

\subsection{Partial Ancestor Omission}\label{app:partial-omission}

The deterministic bound above applies to each realized caption assignment. To study the random imbalance caused by ancestor omission, we fix the image sample and independently retain each occurrence of a concept caption with the same probability across leaves. Conditional on the retained count, the positive set is a uniform sample of the original leaf-uniform captions. We can therefore bound its expected bias separately from the effect of a smaller $N_k$ on the stochastic transfer term.

\begin{lemma}[Caption bias under partial ancestor omission]\label{lem:partial-omission-bias}
Suppose \Cref{ass:prototype-centroid-consistency} holds, and fix a non-root node $k$ with subtree radius $R_k:=\max_{\ell\in L_k}\|\bm{\mu}_\ell-\bm{\mu}_k\|$. Suppose that, before omission, the positive set of $k$ contains equally many images from each leaf in $L_k$, and that each of these images independently keeps caption $k$ with the same probability. Then, for every $m\ge1$ with $\Pr(N_k=m)>0$, the caption bias of the retained positive set satisfies
\begin{equation}\label{eq:partial-omission-bias}
  \E\left[\bm{b}_k\mid N_k=m\right]=\bzero,
  \qquad
  \E\left[\|\bm{b}_k\|\mid N_k=m\right]\le\frac{R_k}{\sqrt{m}}.
\end{equation}
\end{lemma}

\begin{proof}
Condition on $N_k=m$, and let $\cP_k^{\mathrm{full}}$ be the positive set of $k$ before omission. Since every image keeps caption $k$ independently with the same probability, the retained positive set $\cP_k$ is a uniformly random $m$-subset of $\cP_k^{\mathrm{full}}$. Write $\bm{x}_j:=\bm{\mu}_{y_j}-\bm{\mu}_k$ for $j\in\cP_k^{\mathrm{full}}$, so that $\|\bm{x}_j\|\le R_k$ and
\begin{align*}
  \frac{1}{|\cP_k^{\mathrm{full}}|}\sum_{j\in\cP_k^{\mathrm{full}}}\bm{x}_j
  &=\frac{1}{|L_k|}\sum_{\ell\in L_k}\bm{\mu}_\ell-\bm{\mu}_k
  &&\text{(equal image counts across $L_k$)}\\
  &=\bzero
  &&\text{(by \Cref{ass:prototype-centroid-consistency})}.
\end{align*}
Hence
\begin{align*}
  \E[\bm{b}_k\mid N_k=m]
  &=\E\left[\frac{1}{m}\sum_{j\in\cP_k}\bm{x}_j\,\middle|\,N_k=m\right]
  &&\text{(by \Cref{def:caption-bias})}\\
  &=\frac{1}{|\cP_k^{\mathrm{full}}|}\sum_{j\in\cP_k^{\mathrm{full}}}\bm{x}_j
  &&\text{(each image is retained with probability $m/|\cP_k^{\mathrm{full}}|$)}\\
  &=\bzero
  &&\text{(by the zero average above)}.
\end{align*}
Sampling without replacement has variance at most that of sampling with replacement, so
\begin{align*}
  \E[\|\bm{b}_k\|\mid N_k=m]
  &\le\sqrt{\E[\|\bm{b}_k\|^2\mid N_k=m]}
  &&\text{(by Jensen's inequality)}\\
  &\le\sqrt{\frac{1}{m}\cdot\frac{1}{|\cP_k^{\mathrm{full}}|}\sum_{j\in\cP_k^{\mathrm{full}}}\|\bm{x}_j\|^2}
  &&\text{(variance with replacement, zero average)}\\
  &\le\frac{R_k}{\sqrt{m}}
  &&\text{(since $\|\bm{x}_j\|\le R_k$)}.
\end{align*}
\end{proof}

Omission can also lower $N_k$, which affects the stochastic transfer term through the realized $N_{\min}$ and $\xi_{\max}$. If the resulting labels remain path-consistent, are independent of the image variations, and keep $N_k>0$ for every non-root node $k$, we can condition on them and apply \Cref{cor:caption-relaxation}. The off-path term in $\delta_{\mathrm{cap}}$ is then zero, as is $\delta_{\mathrm{cons}}$ under \Cref{ass:prototype-centroid-consistency}.

With leaf-dependent retention, $\E[\bm{b}_k\mid N_k=m]$ need not vanish. For each realized path-consistent assignment, we can still use \Cref{lem:off-path-bias} to bound the caption bias by the $\delta_{\mathrm{cap}}$ of that assignment, whose off-path term is zero.

\subsection{Total Caption Omission: Collapse of Uncaptioned Concepts}\label{app:total-omission}

When $N_k=0$, the caption mean is undefined, so the transfer corollary does not apply. We analyze the optimizer directly to determine how uncaptioned concepts affect \HAS.
\begin{proposition}[Collapse of uncaptioned concepts]\label{prop:total-ancestor-omission}
Fix the image features and observed captions, let $\lambda>0$, and define the uncaptioned non-root concepts by
\begin{equation}\label{eq:total-omission-sets}
  U:=\{k\in V\setminus\{\mathrm{root}\}:N_k=0\}.
\end{equation}
Under either the linear contrastive loss of \Cref{eq:objective-linear-sampled-path} or the InfoNCE loss of \Cref{eq:objective-simplified-finite}, every pair of distinct concepts in $U$ has equal optimal text embeddings. For every non-root non-leaf node $u$, write $\mathcal{O}_u:=V\setminus\bigl(\mathrm{Sub}(u)\cup\mathrm{Anc}(u)\bigr)$ for its off-branch set. If the optimal text embeddings are nonzero at every non-root node, then every such $u$ with $U\cap\mathrm{Desc}(u)\neq\varnothing$ satisfies
\begin{equation}\label{eq:total-omission-violation-bound}
  \HASViol^{\mathrm{txt}}(u)
  \ge
  \frac{|U\cap\mathcal{O}_u|}{|\mathcal{O}_u|}\times100\%.
\end{equation}
In particular, \HAS cannot hold if some non-root non-leaf node $u$ has both $U\cap\mathrm{Desc}(u)\neq\varnothing$ and $U\cap\mathcal{O}_u\neq\varnothing$. Under the linear contrastive loss, every uncaptioned concept $k$ also has the exact embedding
\begin{equation}\label{eq:total-omission-collapse}
  \ba_k^{\mathrm{lin},*}=-\frac{N}{2(M-1)\lambda}\,\bar{\bz}.
\end{equation}
\end{proposition}

\begin{proof}
Fix either loss and write $\{\ba_m\}_{m\in[M]}$ for its optimal text embeddings. Let $k\neq k'$ belong to $U$. Exchanging $\ba_k$ and $\ba_{k'}$ leaves the objective unchanged. Neither index occurs in a positive term or a text-to-image term, the image-to-text terms sum symmetrically over concept indices, and the ridge penalty assigns the same coefficient to both embeddings. The linear loss has a unique minimizer by \Cref{prop:linear-sampled-path-minimizer}, and the InfoNCE loss has a unique minimizer by \Cref{lem:strong-convexity}. The swapped optimizer must therefore equal the original optimizer, so $\ba_k=\ba_{k'}$.

Suppose every non-root $\ba_m$ is nonzero, and fix a non-root non-leaf node $u$ with $U\cap\mathrm{Desc}(u)\neq\varnothing$. Choose $v\in U\cap\mathrm{Desc}(u)$. For every $w\in U\cap\mathcal{O}_u$, the equality above gives $\ba_w=\ba_v$, and therefore
\begin{equation*}
  s_u^{\mathrm{txt}}
  \le
  \cos(\ba_u,\ba_v)
  =
  \cos(\ba_u,\ba_w),
\end{equation*}
where $s_u^{\mathrm{txt}}$ is the descendant threshold in \Cref{def:slack-violation-percentage} at zero slack. Thus every $w\in U\cap\mathcal{O}_u$ is counted in $\HASViol^{\mathrm{txt}}(u)$, which proves \Cref{eq:total-omission-violation-bound}. If $U\cap\mathrm{Desc}(u)$ and $U\cap\mathcal{O}_u$ are both nonempty, the right-hand side of \Cref{eq:total-omission-violation-bound} is positive, so \HAS cannot hold. If any non-root optimal text embedding is zero, the required cosine is undefined, so \HAS cannot hold in that case either.

For the additional linear characterization, fix an uncaptioned concept $k$. Then $Y_{kj}=N_k=0$ for every $j\in[N]$, so \Cref{eq:linear-sampled-path-matrix} gives
\begin{align*}
  B_{kj}
  &=\frac{1}{2}\left(0-\frac{1}{M-1}\right)
    +\frac{1}{2}\left(0-\frac{0}{N-1}\right)\\
  &=-\frac{1}{2(M-1)}.
\end{align*}
By \Cref{prop:linear-sampled-path-minimizer}, the optimal embedding is therefore
\begin{align*}
  \ba_k^{\mathrm{lin},*}
  &=\frac{1}{\lambda}\sum_{j=1}^{N}B_{kj}\bz_j
  &&\text{(by \Cref{eq:linear-sampled-path-minimizer})}\\
  &=-\frac{1}{2(M-1)\lambda}\sum_{j=1}^{N}\bz_j
  &&\text{(substituting $B_{kj}$)}\\
  &=-\frac{N}{2(M-1)\lambda}\bar{\bz}
  &&\text{(by the definition of $\bar{\bz}$)},
\end{align*}
which is \Cref{eq:total-omission-collapse}.
\end{proof}

\end{document}